\pdfoutput=1

\documentclass{article}

\usepackage[eandd, final]{neurips_2026}

\usepackage[utf8]{inputenc}
\usepackage[T1]{fontenc}
\usepackage{microtype}

\usepackage{amsmath}
\usepackage{amssymb}
\usepackage{amsfonts}
\usepackage{mathtools}
\usepackage{amsthm}
\usepackage{nicefrac}

\DeclareMathOperator*{\argmin}{arg\,min}

\usepackage{graphicx}
\usepackage{subcaption}
\usepackage{wrapfig}    %
\usepackage{needspace}  %
\usepackage{placeins}   %
\usepackage{flafter}    %
\usepackage{booktabs}
\usepackage{makecell}
\usepackage{multirow}
\usepackage{longtable}
\usepackage[table]{xcolor}   %
\usepackage{array}

\newcolumntype{L}[1]{>{\raggedright\let\newline\\\arraybackslash\hspace{0pt}}m{#1}}
\newcolumntype{C}[1]{>{\centering\let\newline\\\arraybackslash\hspace{0pt}}m{#1}}
\newcolumntype{R}[1]{>{\raggedleft\let\newline\\\arraybackslash\hspace{0pt}}m{#1}}

\usepackage{url}
\usepackage{hyperref}
\definecolor{linkcolor}{RGB}{92, 92, 192}
\definecolor{oracleTest}{HTML}{1565C0}
\definecolor{oracleTrain}{HTML}{00897B}
\definecolor{oracleInternal}{HTML}{C62828}
\newcommand{\NEWon}{}

\hypersetup{colorlinks=true, linkcolor=linkcolor, citecolor=linkcolor}

\usepackage{tcolorbox}

\usepackage{tikz}
\usetikzlibrary{backgrounds}
\usetikzlibrary{tikzmark}
\usetikzlibrary{calc}
\usetikzlibrary{arrows,shapes,positioning,shadows,trees,mindmap}
\usetikzlibrary{arrows.meta}

\theoremstyle{plain}
\newtheorem{theorem}{Theorem}[section]

\theoremstyle{definition}
\newtheorem{definition}[theorem]{Definition}

\theoremstyle{remark}

\usepackage{snaptodo}
\snaptodoset{block rise=1em}
\snaptodoset{block rise=2em}
\snaptodoset{margin block/.style={font=\tiny}} 
\snaptodoset{margin block/.style={font=\tiny}}

\title{Crossing the Validation Crisis: Cross-Validation Reduces Benchmarking Variance Surprisingly Well}

\author{
  \textbf{Célestin Eve}$^{1,2}$ \quad
  \textbf{Gaël Varoquaux}$^{2,3}$ \quad
  \textbf{Thomas Moreau}$^{1}$ \\
  \vspace{0.5em} \\
  \small $^{1}$MIND Team, Université Paris-Saclay, Inria, CEA, Palaiseau, France \\
  \small $^{2}$SODA Team, Inria, Palaiseau, France \quad
  \small $^{3}$Probabl \\
  \texttt{celestin.eve@inria.fr}
}

\newcommand{\codeurl}{\url{https://github.com/soda-inria/crossing-the-validation-crisis}}

\begin{document}

\maketitle

\begin{abstract}
  Modern machine learning progresses through empirical work, benchmarking new methods to evaluate relative performance. However, the statistical variability inherent in evaluation, exacerbated by the stochastic nature of many algorithms, often makes performance estimation unreliable with the limited test samples available. {The result is} a validation crisis{,} in which genuine advances are difficult to discern. Here, we show that doing more cross-validation markedly improves confidence when evaluating and comparing learning algorithm performances. We introduce the concept of \emph{sample gain}, which quantifies the virtual data augmentation achieved by using multiple cross-validation splits to reduce benchmarking variance. Experiments on both synthetic and real-world datasets (histopathologic scans and NLP fine-tuning) demonstrate that multiple splits can substantially improve the reliability and stability of performance estimates, with diminishing returns often setting in later than expected. We also introduce a procedure to dynamically early-stop cross-validation by estimating from the first few {splits whether further splits} will bring large sample gains. Our findings highlight the value of pushing cross-validation on available samples to achieve robust and reliable {benchmarking, and lead to a simple recommendation: \emph{split repeatedly}.}
\end{abstract}

\section{Introduction: the challenge of evaluation noise}

\begin{wrapfigure}[11]{r}{0.44\textwidth}\vspace{-2em}
    \includegraphics[width=\linewidth]{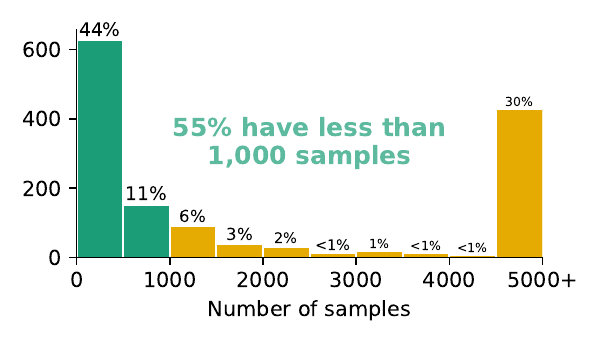}
    \caption{Size distribution of the 20\% most used {OpenML datasets, November 2026}.}
    \label{fig:openml_20_percent_datasets}
\end{wrapfigure}

Machine learning (ML) has evolved into an empirical science, where progress is driven by benchmarking of learning algorithms \citep{eriksson2025trustaibenchmarksinterdisciplinary, hardt2026emerging}. Investigators developing a new algorithm need to assess whether it advances the state of the art, providing evidence of improved performance.

\looseness=-1
But benchmarking algorithms is inherently variable {due to sample ordering, weight initialization, hyperparameter optimization and, above all, data splitting \citep{bouthillier_accounting_2021}}. This variability can lead to inconsistent algorithm rankings and unreliable inferences about relative effectiveness. For instance, in the Grand Challenge on \texttt{PatchCamelyon} (\texttt{PCam}) for histopathology image classification\footnote{\url{https://patchcamelyon.grand-challenge.org/evaluation/challenge/leaderboard/}}, top-performing algorithms {were} statistically indistinguishable, highlighting how {noise can obscure close margins}.

For major application fields, reference benchmarks {like} ImageNet for image classification \citep{deng_imagenet_2009} or Atari games for reinforcement learning \citep{bellemare_arcade_2013} provide reproducibility. {Yet in many ML domains} such as time series anomaly detection \citep{sarfraz_position_2024} or medical imaging \citep{christodoulou_confidence_2024, christodoulou_false_2025}, {reference benchmarks are not fixed, and} custom evaluation processes {worsen} inconsistencies.

\begin{wraptable}{r}{0.57\textwidth}
\vskip-0.9em
\centering
\caption{Datasets exemplifying the validation crisis.}
\label{table:datasets_poor_sample}
\scriptsize
\setlength{\tabcolsep}{1.6pt}
\renewcommand{\arraystretch}{0.92}
\rowcolors{4}{white}{gray!10}
\begin{tabular}{L{1.64cm} L{1.70cm} L{2.30cm} R{1.06cm} R{.70cm}}
\toprule
\textbf{Domain} & \textbf{Dataset} & \textbf{Reference} & \textbf{Size} & \textbf{Citations} \\
\midrule
\multicolumn{5}{c}{\textbf{NLP}}\\
\midrule
Medical NLP & ACI-Bench & \citealt{yim_aci_bench_2023} & 207 & {237} \\
Medical NLP & {MTS-Dialog} & \citealt{ben_abacha_empirical_2023} & {1{,}701} & {179} \\
{Medical NLP} & {PubMedQA} & \citealt{jin_pubmedqa_2019} & {1,000} & {2,542} \\
\midrule
{General NLU} & {SuperGLUE} & \citealt{wang_superglue_2019} & {306--2,767} & {3,773} \\
Legal & CUAD & \citealt{hendrycks_cuad_2021} & 510 & {573} \\
\midrule
{Fact-checking} & {SciFact} & \citealt{wadden_scifact_2020} & {1,409} & {1,254} \\
Scientific NLP & SciDTB & \citealt{yang-li-2018-scidtb} & {798} & {77} \\
\midrule
\multicolumn{5}{c}{\textbf{Medical imaging}}\\
\midrule
{Segmentation} & MSD ({10 tasks}) & \citealt{antonelli_medical_2022} & 30--750 & {2,260} \\
{Classification} & {MedMNIST v2} (18 datasets) & \citealt{yang_medmnist_2023} & {780--1,633} (Q1) & {2,275} \\
\midrule
\hiderowcolors\multicolumn{5}{c}{\textbf{Others}}\\
\showrowcolors\midrule
{Code generation} & {MBPP} & \citealt{austin_mbpp_2021} & {974} & {5,736} \\
{Math reasoning} & {SVAMP} & \citealt{patel_svamp_2021} & {1,000} & {1,514} \\
\midrule
{Drug discovery} & {TDC ADMET} & \citealt{huang_tdc_2021} & {475--906} & {792} \\
\midrule
Tabular ML & {TabZilla}\newline (36 datasets) & \citealt{mcelfresh_2023} & 148--{2,550} (median) & {585} \\
Tabular ML & 111 datasets & \citealt{shmuel_comprehensive_2024} & 43--674 (Q1) & {120} \\
\bottomrule
\end{tabular}
\par\vspace{2pt}
{\scriptsize\raggedright Size: samples in the whole dataset, or in the parts used: RTE, CB, COPA and WSC for SuperGLUE, PQA-L for PubMedQA, the 9 ADMET sets under 1,000 molecules for TDC. Citations: Google Scholar, November 2026.\par}
\end{wraptable}

Compounding this issue is the frequent scarcity of data,
particularly in domains requiring human labeling, such as natural language processing (text annotation) or medical imaging~\citep{litjens_medical_2017}, where limited sample sizes amplify evaluation variability \citep{VAROQUAUX2018}. Meta-research studies across multiple ML communities provide growing evidence of pervasive small-sample regimes \citep{roberts_common_2021, kokol_machine_2022, chen_sampling_2023, christodoulou_false_2025}{, a situation we call a} \textbf{validation crisis}{,} in which insufficient data hampers reliable progress. Much ML and AI research on application domains relies on open datasets with limited sample sizes, some of which are very widely used: more than half of the 20\% most used OpenML datasets have fewer than 1,000 samples (\autoref{fig:openml_20_percent_datasets}). \autoref{table:datasets_poor_sample} gives several representative examples, including genAI settings.

Cross-validation (CV) is a classic tool to {estimate generalization performance with less variance than a single train/test split: it partitions the dataset, trains and evaluates the algorithm repeatedly, and averages the results} \citep{kohavi_cv_1995}. By increasing confidence, CV can enhance the reliability of comparisons \citep{arlot_survey_2010}. However, CV appears underused in modern practice: among the 10,963 studies {covered by the 220 systematic reviews of ML in healthcare published between 2010 and 2023 and analysed by} \citet{kolasa_systematic_2024}, fewer than 14\% {are reported to use CV}. {Reviewing 62 studies of ML on COVID-19 chest imaging, \citet{roberts_common_2021} find that most have a high risk of bias in their analysis, and one of the three main causes is validation on a single hold-out set.}
{Yet} CV escalates computational costs, {requiring} multiple fits per algorithm.
This is especially burdensome for large language models or deep neural networks \citep{strubell-etal-2019-energy}.
CV thus introduces a trade-off between benchmarking variance and cost.

Recent lines of work have attempted to mitigate the instability of empirical evaluation induced by small-sample regimes. One direction focuses on improving resampling strategies: \citet{nagler_reshuffling_2024}'s asymptotic analysis shows that 5-split CV can substantially reduce variance compared to single hold-out evaluations{, and that reshuffling splits per configuration can improve tuning}. This motivates the finite-sample question: how much precision do additional splits buy in practice, and when does the extra compute stop paying off? {Another direction relies on} large-scale and continuously updated benchmarks {to} average evaluation noise across datasets and experimental runs, as exemplified by TabArena \citep{erickson_tabarena_2025}. Older work bears on a complementary question, comparing algorithms rather than estimating the performance of one: \emph{racing} \citep{maron_hoeffding_1994, moore_efficient_1994} stops cross-validating a candidate model once it is almost certainly worse than a rival, and \citet{bouckaert_evaluating_2004} study how the number of repetitions in a performance estimate affects the \emph{replicability} of significance tests in such comparisons.

In this paper, we show that doing many cross-validation splits substantially improves {the precision of a learning algorithm's performance estimate and, as a corollary, the comparison of several algorithms}.
It brings a large \textbf{variance-equivalent test sample gain}: the benchmark behaves as though its test set had been enlarged, by a factor {that depends on the learning algorithm, from about $5$ for ridge regression to $15$ for random features after $200$ splits in our simulations, and up to about $11$ after $20$ splits on real data}. Our results indicate that \textbf{increasing the number of splits exhibits diminishing returns more slowly than anticipated}.
We also show that \textbf{this gain can be partially anticipated from the first few CV-splits}{, by a statistic we call the \emph{redundancy}}, turning the choice of the number of splits into an explicit statistical and computational trade-off. {Our code is available at \codeurl.}

\section{{What does benchmarking estimate?}}

\looseness=-1
There are various benchmark settings and goals: evaluation may focus on performance within a single dataset or on robustness across multiple datasets, and may compare either fixed decision functions or learning algorithms, an important distinction that we detail below. In this work, we focus on single-dataset benchmarking, as across-dataset benchmarking calls for different statistical methodology \citep{demsar_2006}. This section summarizes the single-dataset benchmarking methodology and its formal framework in machine learning, and then {turns} to practical resampling estimators.

\subsection{Problem setting}
\label{sec:problem-setting}

A machine-learning {task is} defined on the space $\mathcal{X} \times \mathcal{Y}$, with data $(X,Y)$ drawn from a joint distribution $p_{X, Y}$. Here, $X$ represents the \textit{input} features, typically with $\mathcal X = \mathbb R^d$ and $Y$ the \textit{output}, which can be discrete \textit{labels} $\mathcal Y = \{0,\dots,C-1\}$ for a $C$-class classification, or continuous \textit{outcomes} $\mathcal{Y} = \mathbb{R}$ for regression.
Let $\ell : \mathcal{Y} \times \mathcal{Y} \to \mathbb{R}$ be a loss function measuring the quality of a prediction $\hat y\in\mathcal Y$ compared to a ground truth $y$. The most common choices are the squared loss $\ell_2$ for regression and the error rate for classification. Given a \emph{decision function} (a trained predictor) $g \in \mathcal{Y}^\mathcal{X}$, its quality is theoretically measured by the \textit{population risk}:
\begin{equation}
    \mathcal{R}^*(g) = \mathbb E_{(X, Y) \sim p_{X, Y}}\left[\ell(g(X), Y)\right].
\end{equation}

\paragraph{Benchmarking trained predictors: decision functions}
Benchmarking decision functions strives to rank functions in $\mathcal{Y}^\mathcal{X}$ by estimating their relative performance. The \textit{oracle ranking} is the order that respects the population risk $\mathcal{R}^*$, which we call the \textit{oracle {risk}}. If we evaluate three decision functions $g_i, g_j, g_k$ where $\mathcal{R}^*(g_k) < \mathcal{R}^*(g_j) < \mathcal{R}^*(g_i)$, the \emph{oracle ranking} is:
$ 1.\;{g_k} \succ 2.\;{g_j} \succ 3.\;{g_i} $.
Note that if the chosen metric is a utility (e.g., $R^2$ or accuracy), the ranking order is reversed.

In practice, $p_{X, Y}$ is unknown. {With a dataset $\mathcal{D}$ sampled from it, the empirical score is}:
\begin{equation}
    \label{global_opt}
    R_{\mathcal{D}}(g) = \frac{1}{|\mathcal{D}|}  \sum_{(x,y) \in \mathcal{D}} \ell(g(x),y).
\end{equation}
This empirical score is an unbiased estimator of the oracle {risk}: for any sample size $m$, $\mathbb{E}_{\mathcal{D} \sim p_{X,Y}^{\otimes m}}[R_{\mathcal{D}}(g)] = \mathcal{R}^*(g)$. A ranking is considered robust if the empirical ranking derived from $R_{\mathcal{D}}$ mimics the oracle ranking.
The discrepancy between these rankings, caused by evaluating on finite observations rather than the full population, is a primary challenge in benchmarking.

\paragraph{Accounting for training: benchmarking learning algorithms}

While the previous {paragraph} treats decision functions as static objects, in ML these functions are the output of \textit{learning algorithms}. Benchmarking decision functions may suffice for deployment-oriented applications, but the ML research literature studies learning algorithms. Here, good benchmarking is essential for scientific progress, as unstable rankings can propagate misleading conclusions through the literature \citep{dietterich_approximate_1998}. Learning algorithms can be stochastic, and their performance depends on the data they ingest during training \citep{bouthillier_accounting_2021, picard2021torch}.

Let $F_\lambda$ be a learning algorithm that, for a fixed training size $n_{\text{tr}}$, maps a training set $\mathcal{D}^{\text{tr}} \in (\mathcal{X} \times \mathcal{Y})^{n_{\text{tr}}}$ and other more arbitrary states (such as random inits) $\xi \sim \Xi$ to a decision function $g \in \mathcal{Y}^{\mathcal{X}}$:
\begin{equation}
    F_\lambda: (\mathcal{D}^{\text{tr}}, \xi) \longmapsto g \enspace.
\end{equation}
Here, $\lambda$ denotes fixed hyper-parameters. The goal of $F_\lambda$ is typically to find a function that approximates the risk minimizer $g^* \in \argmin_{g} \mathcal{R}^*(g)$.
To benchmark learning algorithms, we are interested in their \emph{expected} performance across training sets of size $n_{\text{tr}}$ sampled from $p_{X,Y}^{\otimes n_{\text{tr}}}$ and internal random states. The underlying question of benchmarking is: which learning algorithm produces better decision functions for a specific task \citep{dietterich_approximate_1998}. The corresponding \emph{estimand}, or quantity to estimate, is the expected {population risk} accounting for these sources of variation:

\begin{tcolorbox}[colback=gray!10, colframe=gray!50, arc=2pt, boxrule=1pt, left=4pt, right=4pt, top=4pt, bottom=4pt]
\textbf{Definition}: \textit{Oracle {risk}} of a learning algorithm $F_\lambda$ for training size $n_{\text{tr}}$:
\begin{align}
    \label{eq:oracle_score}
    \mathcal{R}^*_{n_{\text{tr}}}(&F_\lambda)
    = \mathbb{E}_{\mathcal{D}^{\text{tr}}\sim p_{X,Y}^{\otimes n_{\text{tr}}}} \left[ \mathbb{E}_{\xi \sim \Xi} \left[ \mathcal{R}^*(F_\lambda(\mathcal{D}^{\text{tr}}, \xi)) \right] \right] \\
    &= \mathbb{E}_{\tikzmarknode{test}{\mathcal{D}^{\text{te}}}\sim p_{X,Y}^{\otimes n_{\text{te}}}} \left[ \mathbb{E}_{\tikzmarknode{train}{\mathcal{D}^{\text{tr}}}\sim p_{X,Y}^{\otimes n_{\text{tr}}}} \left[ \mathbb{E}_{\tikzmarknode{internal}{\xi} \sim \Xi} \left[ R_{\mathcal{D}^{\text{te}}}(F_\lambda(\mathcal{D}^{\text{tr}}, \xi)) \right] \right] \right].
    \nonumber
\end{align}%
\begin{tikzpicture}[overlay,remember picture,>=stealth,
    lbl/.style={anchor=north,inner sep=0pt,text height=1.6ex,text depth=0.25ex,font=\footnotesize}]
    \path let \p1=(test.south), \p2=(train.south), \p3=(internal.south) in
        coordinate (tip) at (0,{min(\y1,\y2,\y3)-1.5pt});
    \coordinate (top) at ($(tip)+(0,-7.5pt)$);
    \node[lbl,color=oracleTest] (testtitle) at (test.center |- top) {Test data};
    \draw[->,color=oracleTest] ($(test.center |- top)+(0,1pt)$) -- (test.center |- tip);
    \node[lbl,color=oracleTrain] (traintitle) at (train.center |- top) {Train data};
    \draw[->,color=oracleTrain] ($(train.center |- top)+(0,1pt)$) -- (train.center |- tip);
    \node[lbl,color=oracleInternal] (internaltitle) at (internal.center |- top) {Internal states};
    \draw[->,color=oracleInternal] ($(internal.center |- top)+(0,1pt)$) -- (internal.center |- tip);
\end{tikzpicture}%
\end{tcolorbox}
{In equation~\eqref{eq:oracle_score}}, the inner expectation handles internal algorithmic variance ({random states $\xi$}), the middle expectation handles the variance due to the choice of training data of size $n_{\text{tr}}$, and the outer expectation (approximated by a \textit{test set} $\mathcal{D}^{\text{te}}$) evaluates the generalization to the population distribution. This {risk} measures the capacity of the algorithm to consistently produce high-performing predictors. {This paper studies how precisely one algorithm's score is estimated, and treats comparisons as a complement.}

\paragraph{Tuning and selection are part of the procedure}\looseness=-1
In this formulation, the learning procedure $F_\lambda$ encompasses the whole procedure, including any model tuning and parameter selection such as an internal hyper-parameter search, a validation split, early stopping or checkpoint selection. These procedures are valid $F_\lambda$ provided that everything they consume is drawn from $\mathcal{D}^{\text{tr}}$, and \eqref{eq:oracle_score} is then the oracle {risk} of the procedure as a whole. In particular, the optimism that arises from selecting on validation data is part of the procedure's performance rather than a bias to be removed: it is what one compares when comparing algorithms that tune. \autoref{appendix:hpo} reports sample gains for such procedures.

\subsection{Estimation via resampling}
\label{sec:resampling}

In practice, the oracle {risk} $\mathcal{R}^*_{n_{\text{tr}}}(F_\lambda)$ defined in \eqref{eq:oracle_score} cannot be computed, as the data-generating distribution $p_{X,Y}$ is unknown and only a single finite dataset
\(\mathcal{D} = \{(x_i, y_i)\}_{i=1}^{n}\), with $(x_i,y_i) \overset{\mathrm{i.i.d.}}{\sim} p_{X,Y}$
is available. To estimate it, we use resampling methods like \textit{Monte-Carlo Cross-Validation} (MCCV, \citealt{picard_cross-validation_1984}), also called \textit{Repeated Random Sub-sampling Validation}, or \textit{ShuffleSplit} in the \texttt{scikit-learn} library \citep{pedregosa_scikit-learn_2011}. {Unlike $K$-fold CV \citep{Geisser01061975}, MCCV draws its splits independently, so that their number $K$ is not tied to the test fraction and can grow, or stop, at will.}

A draw $\eta \sim H$ randomly partitions $\mathcal{D}$ into training and test sets $\bigl(\mathcal{D}^{\text{tr}}_\eta, \mathcal{D}^{\text{te}}_\eta\bigr)$, with fixed sizes $n_{\text{tr}}$ and $n_{\text{te}}$ \citep{kohavi_cv_1995}. {Drawing first a total dataset, then a random train/test split, rewrites \eqref{eq:oracle_score} as:}
\begin{equation}
\mathcal{R}^*_{n_{\text{tr}}}(F_\lambda) = \mathbb{E}_{\mathcal{D}\sim p_{X,Y}^{\otimes n}} \Bigl[
\mathbb{E}_{\eta\sim H} \bigl[ \mathbb{E}_{\xi\sim\Xi} \bigl[ R_{\mathcal{D}^{\text{te}}_\eta} \bigl(F_\lambda(\mathcal{D}^{\text{tr}}_\eta,\xi)\bigr) \bigr] \bigr] \Bigr].
\end{equation}

A single hold-out evaluation is defined as:
\begin{equation}
  \widehat{R}_{\text{HO}} = R_{\mathcal{D}^{\text{te}}_\eta}\bigl(F_\lambda(\mathcal{D}^{\text{tr}}_\eta, \xi)\bigr).
\end{equation}

{With $K$} i.i.d. splits $(\eta_k)_{k=1}^K$ and randomness $(\xi_k)_{k=1}^K$, the MCCV estimator {is}:
\begin{equation}
  \widehat{R}_{K} = \frac{1}{K} \sum_{k=1}^{K} \underbrace{R_{\mathcal{D}^{\text{te}}_{\eta_k}}\bigl(F_\lambda(\mathcal{D}^{\text{tr}}_{\eta_k}, \xi_k)\bigr)}_{E_k}
\end{equation}

Here, $E_k$ denotes the evaluation on a single split of the MCCV procedure.
{These splits} may overlap, allowing the same observation to appear in the test set multiple times across repetitions. Hence, the variables $(E_k)_{k=1}^K$ are identically distributed but not independent.

\paragraph{An unbiased estimator with respect to the estimand}
Because each split samples a training set of fixed size $n_{\text{tr}}$, MCCV is an \emph{unbiased estimator of the oracle {risk} {\eqref{eq:oracle_score}} for training size $n_{\text{tr}}$}{, whatever the number of splits \citep[Eq.~13]{arlot_survey_2010}:}
\begin{equation}
\mathbb{E}\bigl[\widehat{R}_{K}\bigr] = \mathcal{R}^*_{n_{\text{tr}}}(F_\lambda).
\end{equation}
\autoref{appendix:benchmarking-derivations} translates the source results used in this section to our notation.

\paragraph{Variance of the estimator}
Results from \citet[Eq.~8]{nadeau_inference_2003} show:
\begin{equation}
  \operatorname{Var}[\widehat{R}_{K}] = \frac{1}{K} \sigma^2_{\text{HO}} + \frac{K-1}{K} \tau,
  \label{eq:cvvar}
\end{equation}
where $\sigma^2_{\text{HO}} = \operatorname{Var}[\widehat{R}_{\text{HO}}]$ is the variance of a single hold-out evaluation, and $\tau = \operatorname{Cov}[E_1, E_2] \mathrel{\ge} 0$ quantifies the covariance between evaluations on different splits, \emph{i.e.}, the correlation induced by reusing observations across splits.
As we have $\tau \le \sigma^2_{\text{HO}}$ (by Cauchy-Schwarz inequality), the variance decreases with $K$.
When the number of repetitions $K \to \infty$, the first term vanishes, but the second converges to $\tau$, an irreducible variance component stemming from the finite nature of the dataset.

{\citet[Prop.~3]{nadeau_inference_2003}} show that {no estimator of $\operatorname{Var}[\widehat{R}_{K}]$ that is linear and/or quadratic in the individual test losses is unbiased in general}.
{In their Section~3.1, they stipulate}:
\begin{equation}
\tau \approx \frac{n_{\text{te}}}{n_{\text{te}} + n_{\text{tr}}} \sigma^2_{\text{HO}}\enspace.
    \label{eq:nadeau_approx}
\end{equation}
Under this approximation, CV appears to hit diminishing returns quickly. {With $t$ the test fraction of eq.\,\eqref{eq:nadeau_approx}, eq.\,\eqref{eq:cvvar} becomes $\operatorname{Var}[\widehat{R}_{K}] \approx \sigma^2_{\mathrm{HO}}\,(t + (1-t)/K)$, within a factor $2$ of its asymptotic value $t\,\sigma^2_{\mathrm{HO}}$ once $K \ge 1/t$: $10$ splits for the stratified 10-fold CV recommended by \citet{kohavi_cv_1995}, $5$ for a test fraction of 20\%. More splits then seem of little use. Yet in our experiments, with a test fraction of 20\%, the sample gain (\autoref{sec:sample_gain}) keeps growing beyond $K=5$ and can exceed $10$ at $K=20$.}

\section{A measure of cross-validation efficiency: the \emph{sample gain}}
\label{sec:sample_gain}

To {quantify the variance reduction of equation~\eqref{eq:cvvar}}, we introduce the \emph{sample gain}, to compare the variance of an estimator with $K$ repetitions to {that of} a single split (a single hold-out) if we had access to more test data.
This quantity can be estimated empirically{, given} a large enough dataset.

\begin{wrapfigure}[15]{r}{0.49\textwidth}
\vskip-0.6em
    \centering
    \includegraphics[width=0.862\linewidth,trim=7 7 3 3,clip]{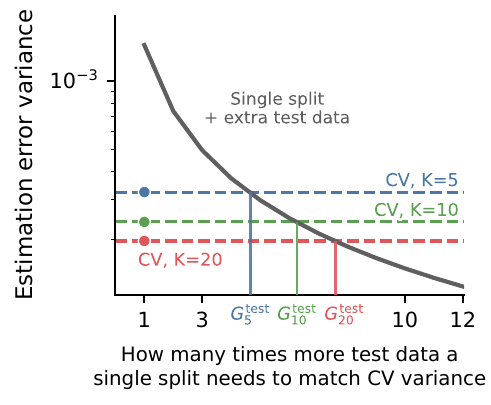}
    \setlength{\abovecaptionskip}{3pt}
    \caption{{Estimation-error variance and variance-equivalent test sample gains on one of our real-data experiments, $G^{\text{test}}_K$ being the sample gain of $K$ splits (equation~\eqref{eq:sample_gain}).}}
    \label{fig:test_variance_diagnostic_pedagogical}
\end{wrapfigure}

\autoref{fig:test_variance_diagnostic_pedagogical} illustrates the {estimation-error variance (defined below)} of a single-split estimator $\widehat{R}_{\text{HO}}$ evaluated with an increasingly large test set. It is compared with {that} of a CV estimator $\widehat{R}_K$ evaluated on a fixed test size $n_{\text{te}}$.
The \emph{sample gain} quantifies how much larger a single test set would need to be to have the same evaluation variability as the one observed with CV.

\paragraph{Measuring the evaluation noise}

Let $g = F_\lambda(\mathcal{D}^{\text{tr}}, \xi)$ be the predictor obtained using the data split $\mathcal{D}^{\text{tr}}$ and randomness $\xi$, with fixed training size $n_{\text{tr}}$.
We denote by $\widehat{R}^{\star}(g)$ an unbiased estimator of the oracle risk $\mathcal{R}^*(g)$, constructed independently of the test set used for evaluation. For a given test set $\mathcal{D}^{\text{te}}$, we can approximate the \emph{estimation error} on this split as:
\begin{equation}\postdisplaypenalty=10000 %
\delta_{\text{HO}} = R_{\mathcal{D}^{\text{te}}}(g) - \widehat{R}^{\star}(g).
\label{eq:delta_oracle_estimator}
\end{equation}

To understand what $\delta_{\text{HO}}$ measures, consider the variance of the empirical score $R_{\mathcal{D}^{\text{te}}}(g)$. By the Law of Total Variance, we can decompose it into evaluation noise and training instability:
\begin{equation}
    \operatorname{Var}[R_{\mathcal{D}^{\text{te}}} (g)] = \underbrace{\mathbb{E}_{g} \left[ \operatorname{Var}[R_{\mathcal{D}^{\text{te}}} (g) \mid g] \right]}_{\text{(A) Expected Evaluation Noise}} + \underbrace{\operatorname{Var}_{g} \left[ \mathbb{E}[R_{\mathcal{D}^{\text{te}}} (g) \mid g] \right]}_{\text{(B) Training Instability}}.
    \label{eq:total_variance_process}
\end{equation}
Term (B) simplifies to $\operatorname{Var}_{g}[\mathcal{R}^*(g)]${: how much the predictor's oracle risk} varies across training splits. Term (A) {is the score's variance} around the {oracle} risk due to finite test data.

\paragraph{Evaluation-noise approximation}

Suppose that the expected evaluation noise of the oracle-{risk} estimator $\widehat{R}^{\star}(g)$ is negligible compared to that of the test evaluation.
In this case,
\begin{equation}
\operatorname{Var}[\delta_{\text{HO}}] \approx \mathbb{E}_{g} \left[ \operatorname{Var}\left[R_{\mathcal{D}^{\text{te}}} (g) \mid g\right]\right].
\label{eq:delta_variance_approx}
\end{equation}
{The variance of $\delta_{\text{HO}}$ thus keeps the evaluation noise (A) and drops the training instability (B).} The calculations from {equation}~\eqref{eq:total_variance_process} to \eqref{eq:delta_variance_approx} are detailed in \autoref{appendix:sample_gain_justification}.

\paragraph{For the MCCV procedure} Let
$g_k = F_\lambda(\mathcal{D}^{\text{tr}}_{\eta_k}, \xi_k)$ and define the fold-level benchmark-adjusted error
$\delta_k = R_{\mathcal{D}^{\text{te}}_{\eta_k}}(g_k) - \widehat{R}^{\star}(g_k)$.
The corresponding $K$-split estimation error is:
\begin{equation}
    \Delta_K
    = \frac{1}{K}\sum_{k=1}^K \delta_k
    = \underbrace{ \widehat{R}_K }_{\text{CV Estimator}}
      - \underbrace{\frac{1}{K} \sum_{k=1}^K \widehat{R}^{\star}(g_k) }_{\text{Estimated Oracle {Risk}}}.
\end{equation}

{As with} $\widehat R_K$ in
{equation}~\eqref{eq:cvvar}, {the unconditional covariance expansion gives the variance of $\Delta_K$}. Since the fold-level errors
$\delta_1,\ldots,\delta_K$ are exchangeable under {MCCV}, as detailed in \autoref{appendix:sample_gain_justification}:
\begin{equation}
    \operatorname{Var}\left[\Delta_K\right]
    =
    \frac1K \operatorname{Var}\left[\delta_{\mathrm{HO}}\right]
    +
    \frac{K-1}{K}\tau^{\mathrm{te}},
    \label{eq:delta_cvvar}
\end{equation}
where $\tau^{\mathrm{te}} =\operatorname{Cov}(\delta_i,\delta_j)$ for $i\neq j$ is the covariance between benchmark-adjusted fold-level test errors{, estimated in \autoref{sec:inter-fold-covariance-estimation}}. It includes all dependence induced by reusing the finite {dataset $\mathcal{D}$} across MCCV splits, including the {\emph{train/test coupling} intrinsic to cross-validation:} a test observation in one split may belong to the training set of another split.

\begin{definition}[Variance-Equivalent Test Size]
{Let $V^\delta_K(m)$ be} the variance of $\Delta_K$ when each split uses a test set of size $m$, computed across independent {draws of $\mathcal{D}$, $\eta_k$ and $\xi_k$} (for $K=1$, $\Delta_1=\delta_{\text{HO}}$). The variance-equivalent test size $N^{\mathrm{equiv}}_K$ of a CV procedure with $K$ splits is defined as:
\begin{equation}
\begin{split}
N^{\mathrm{equiv}}_K = \min \Big\{ {m} \ge n_{\text{te}} \;\Big|\; &V^\delta_1({m}) \le V^\delta_K(n_{\text{te}}) \Big\}.
\end{split}
\end{equation}
\end{definition}

\begin{definition}[Variance-Equivalent Test Sample Gain]
The \textbf{variance-equivalent test sample gain}{, or \emph{sample gain},} of a CV procedure with $K$ splits is defined as:
\begin{equation}
G^{\text{test}}_K(F_\lambda, n_{\text{tr}}) = \frac{ N^{\mathrm{equiv}}_K}{n_{\text{te}}}.
\label{eq:sample_gain}
\end{equation}
\end{definition}

A value $G^{\text{test}}_K(F_\lambda, n_{\text{tr}}) = 5$ indicates that {$K$ repeated splits reduce the variance as much as a single hold-out with a test set five times larger}. {In the example of} \autoref{fig:test_variance_diagnostic_pedagogical}, the variance-equivalent test sample gains for 5, 10 and 20 splits ($G^{\text{test}}_K(F_\lambda, n_{\text{tr}})$ with $K=5,10,20$) can be {read off the $x$-values} at {which the single-split variance drops below each of} the multi-split CV variances.

This notion provides an interpretable measure of the efficiency of cross-validation: it quantifies variance reduction in units of additional evaluation samples, {a unit common to all learning algorithms}. The estimator used in our figures is detailed in {Appendix}~\ref{sec:sample-gain-estimation}, and its uncertainty in \autoref{appendix:gain-intervals}.

\paragraph{What repeated training runs cannot buy}
The construction of $\delta_k$ also fixes what the gain measures. Because $\delta_k$ subtracts the oracle {risk} of the \emph{same} predictor $g_k$, term (B) of \eqref{eq:total_variance_process} is removed by construction, and $G^{\text{test}}_K$ measures the reduction of the evaluation noise (A). Conditioning on the split, so that only the internal randomness $\xi$ of the algorithm remains, separates two parts of it:
\begin{equation}
    \operatorname{Var}[\delta_{\text{HO}}]
    = \underbrace{\operatorname{Var}\bigl[\mathbb{E}[\delta_{\text{HO}} \mid \mathcal{D}^{\text{tr}}, \mathcal{D}^{\text{te}}]\bigr]}_{(\mathrm{A}_1)\ \text{shared}}
    + \underbrace{\mathbb{E}\bigl[\operatorname{Var}[\delta_{\text{HO}} \mid \mathcal{D}^{\text{tr}}, \mathcal{D}^{\text{te}}]\bigr]}_{(\mathrm{A}_2)\ \text{fit-specific}}.
    \label{eq:evaluation-noise-split}
\end{equation}
The shared part $(\mathrm{A}_1)$ is the error that any fit trained on this split would make on these test observations: it comes from which observations happen to be tested. The fit-specific part $(\mathrm{A}_2)$ changes with the random state: it comes from which of these observations a particular fit gets wrong. It is common practice, especially for deep {networks, to retrain} several times on a fixed split and average the scores. This divides $(\mathrm{A}_2)$ by the number of runs (on raw scores, it also averages out the part of (B) due to the random state) but leaves $(\mathrm{A}_1)$ untouched, and for a deterministic algorithm, {where} $(\mathrm{A}_2) = 0$, repeated runs bring nothing. Only resampling the split acts on $(\mathrm{A}_1)$. \autoref{appendix:sample_gain_justification} derives the gain of a repeated fixed split, and \autoref{appendix:fixed-vs-reshuffled} {compares} the two schemes at matched computational cost.

\Needspace{4\baselineskip}
\section{{Measuring the sample gain: experimental protocol}}
\label{sec:experimental-evaluation}

\begin{wrapfigure}[7]{r}{0.48\linewidth}
\vskip-1.2em
    \centering
    \includegraphics[width=\linewidth]
    {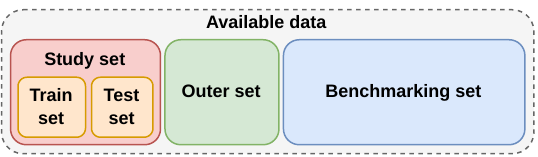}
    \caption{Breakdown and naming of the dataset.}
    \label{fig:dataset_terminology}
\end{wrapfigure}

\paragraph{Approximating oracle quantities}
Estimating the variance of $\delta_{\text{HO}}$ and $\Delta_K$ requires an estimator $\widehat{R}^{\star}(g)$ of $\mathcal{R}^*(g)$ {independent of the test set, with} low evaluation noise. We obtain it from a held-out set much larger than the test set, and decompose the available data into three parts {(\autoref{fig:dataset_terminology})}:

\begin{itemize}
    \item A \emph{study set} simulates realistic benchmarking conditions. It is repeatedly split into training and test subsets $\left(\mathcal{D}^{\text{tr}}_{k}, \mathcal{D}^{\text{te}}_{k}\right)$ with fixed training size $n_{\text{tr}}$ via MCCV. For procedures with internal model selection, such as the deep-learning runs of {Appendix}~\ref{appendix:real-training-loops}, $\mathcal{D}^{\text{tr}}_{k}$ is itself split into training and validation subsets ($n_{\text{tr}}$ denotes their union), and selection never uses $\mathcal{D}^{\text{te}}_{k}$.    \item An \emph{outer set}, used to independently evaluate $V^\delta_1(m)$ at growing test sizes $m$.
    \item A \emph{benchmarking set} $\mathcal{D}^{\text{bench}}$, large enough to serve as {an} oracle estimator $\widehat{R}^{\star}(g) = R_{\mathcal{D}^{\text{bench}}}(g)$ {whose evaluation noise is negligible, satisfying the approximation in \eqref{eq:delta_variance_approx}}.
\end{itemize}

\paragraph{The role of the outer and benchmarking sets}

These two sets are measurement apparatus: they let us observe $G^{\text{test}}_K$, and they require far more labelled data than a benchmark in a small-sample regime has to spare, which is why we carve them out of large public datasets. Nothing downstream depends on them: the recommendations that follow, repeated resampling and the early-stopping redundancy of \autoref{sec:empirical-results}, use the study set alone and therefore cost no additional labels.

\paragraph{Experimental settings}

{We} restrict study sets to sizes typical of ML benchmarks. {Each \emph{seed}, or run, draws its own study set, splits and random states, and variances are taken across seeds.} {We use} the Python libraries \texttt{scikit-learn} \citep{pedregosa_scikit-learn_2011}, \texttt{PyTorch} {\citep{paszke2019pytorch, ansel_pytorch2_2024}} and \texttt{Benchopt} \citep{moreau_benchopt_2022}. The experimental details are in \autoref{appendix:experimental_details}.

\paragraph{Synthetic data}

As a controlled setting, we use a regression task on synthetic data, with a $100{,}000$-sample benchmarking set and $n_{\text{te}}$ set to $20\%$ of the study set{, the test fraction of $5$-fold CV}.

The main runs use \texttt{Ridge} \citep{hoerl_ridge_1970}, \texttt{ExtraTrees} \citep{geurts_extremely_2006}, a one-hidden-layer \texttt{MLP} ($32$ units, \citealt{rumelhart_learning_1986}), and \texttt{RandomFeatures}, a ridge regression on $50$ random Fourier features of a Gaussian kernel \citep{rahimi_random_2007}, lightly penalised ($\alpha = 10^{-3}$). The last two draw their random weights or features afresh at every fit.{} {Further simulated designs, at several target-noise levels ({Appendix}~\ref{appendix:redundancy-runs}), add} \texttt{KNeighbors} \citep{CoverH67}, \texttt{DecisionTree} \citep{BreimanFOS84}, \texttt{HistGradientBoosting} {\citep{ke_lightgbm_2017}}, \texttt{SVR} {\citep{cortes_support-vector_1995, drucker_support_1996}}{, random forests \citep{breiman_random_2001} and gradient boosting \citep{gradientboosting}}{, which} lets us {examine which properties of a learning algorithm drive its sample gain}.

\paragraph{Histopathologic scans classification}

For medical image classification, we use \texttt{PatchCamelyon} (\texttt{PCam}, \citealt{veeling2018rotation}), from the Camelyon16 Grand Challenge on cancer metastasis detection {\citep{ehteshami_bejnordi_diagnostic_2017}}: {$327{,}680$} image patches of size $96 \times 96$ extracted from histopathology scans of lymph nodes, labeled by the presence or absence of metastatic tissue {in their central $32 \times 32$ region, of which we use the $294{,}912$ of the official training and validation splits ({Appendix}~\ref{appendix:real-training-loops})}.

We evaluate three convolutional neural {networks (CNNs)}: \texttt{WideResNet101\_2} \citep{zagoruyko_wide_2016}, \texttt{DenseNet121} \citep{huang_densely_2017}{{} and} \texttt{MobileNetV2} \citep{sandler_mobilenetv2_2018}. {We add} the transformer \texttt{ViT-B/16} {\citep{DosovitskiyB0WZ21}}{, trained with the CNNs' settings ({Appendix}~\ref{appendix:real-training-loops}), under which it learns little: its benchmarking accuracy stays between $0.51$ and $0.59$ at every training size ($0.60$ to $0.85$ for the CNNs). It thus stands for a learning algorithm that often fails to learn, not a competitive model. The pairwise-ranking analysis on \texttt{PCam} uses the three CNNs.}

\paragraph{NLP fine-tuning}

{We} use \texttt{Yelp Review Full} (\texttt{Yelp}, \citealt{zhang2015character}), a classification benchmark of {$700{,}000$} user reviews {rated} from 1 to 5{, of which we use the $650{,}000$ of the training split}. We fine-tune {three} transformer architectures: \texttt{BERT} {\citep{devlin_bert_2019}, \texttt{XLM-RoBERTa} \citep{ConneauKGCWGGOZ20}} and {\texttt{Qwen3-0.6B} \citep{qwen3}, the latter at the smallest training size only ({Appendix}~\ref{appendix:real-training-loops}),} keeping hyperparameters fixed to isolate the effect of the validation strategy. {Sample gains are reported for all three. The pairwise-ranking analysis of \autoref{sec:pairwise-ranking} uses \texttt{BERT} and \texttt{XLM-RoBERTa}.}

\Needspace{0.35\textheight}
\section{Empirical results}
\label{sec:empirical-results}
\subsection{Test sample gain {and when to stop splitting}}

\begin{wrapfigure}[17]{r}{0.44\textwidth}
    \vskip-5.7em
    \centering
    \includegraphics[width=\linewidth]{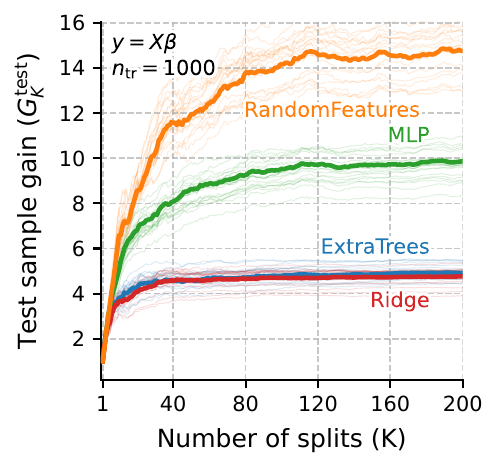}
    \caption{Long-run simulated test sample gains on the mean squared error. Thick lines: estimate on all seeds. Thin lines: $20$ of $1{,}000$ bootstrap resamples of the seeds ($95\%$ intervals in \autoref{appendix:gain-intervals}).}
    \label{fig:avg_test_sample_gain_simulated}
\end{wrapfigure}

\autoref{fig:avg_test_sample_gain_simulated} presents the variance-equivalent test sample gain results on the simulated linear dataset up to $K=200$ splits using 1,000 seeds, at training size $n_{\text{tr}}=1{,}000$. Gains are measured on the mean squared error, the metric on which the redundancy defined below is computed (\autoref{appendix:metrics} explains why $R^2$ is not used). \texttt{Ridge} and \texttt{ExtraTrees} settle early, at $G^{\text{test}}_{200} \approx 5$ ($4.8$ and $4.9$): cross-validating the data 200 times conceptually multiplies their test set by about 5 in this experimental setup.
\texttt{RandomFeatures} and the \texttt{MLP} go much further, up to $G^{\text{test}}_{200} = 15$ and {$10$}, with diminishing returns settling in after about 100 splits, suggesting that CV can yield very important gains for learning algorithms whose fit varies from one training set to the next, such as neural networks.

\begin{figure}[t]
    \centering
    \begin{subfigure}[t]{0.36770\linewidth}
        \centering
        \includegraphics[width=\linewidth]
        {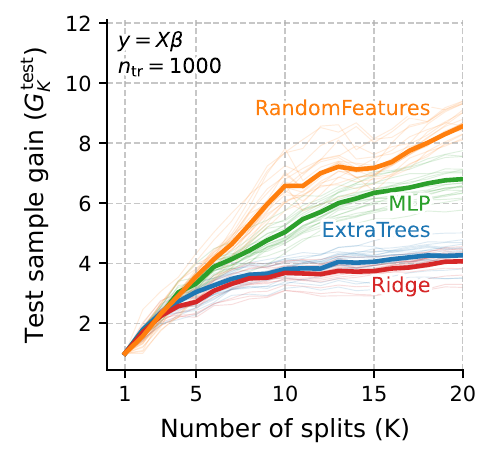}
    \end{subfigure}\hfill
    \begin{subfigure}[t]{0.30615\linewidth}
        \centering
        \includegraphics[width=\linewidth]
        {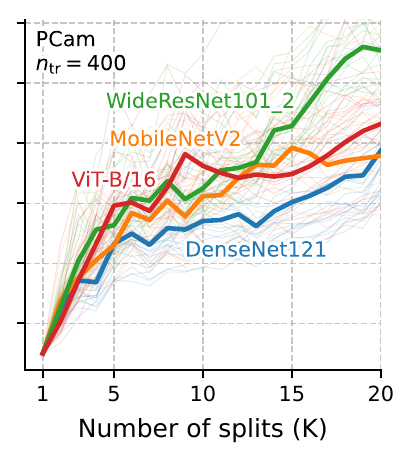}
    \end{subfigure}\hfill
    \begin{subfigure}[t]{0.30615\linewidth}
        \centering
        \includegraphics[width=\linewidth]
        {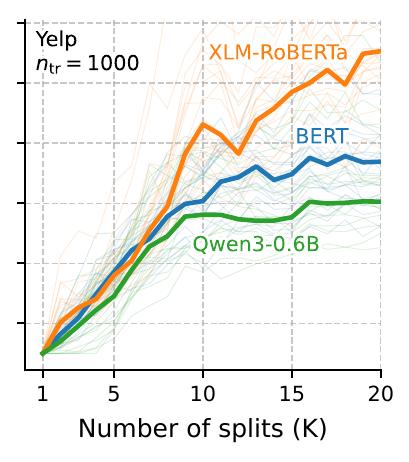}
    \end{subfigure}

    \caption{\textbf{Variance-equivalent test sample gains} {on a common vertical axis (lines as in \autoref{fig:avg_test_sample_gain_simulated})}.}
    \label{fig:all_data_test_sample_gains}
\end{figure}

\autoref{fig:all_data_test_sample_gains} presents the variance-equivalent test sample {gains on the three datasets} up to $K=20$ splits. {In general,} there is an \textbf{important variance-equivalent test sample gain for the vast majority of {our} setups}. Real-data curves are noisier than the simulated ones{, whose cheaper runs are averaged} over many more seeds (\autoref{appendix:experimental_details}), but {behave alike}.

\looseness=-1
These gains beyond $K=5$ splits are present in every experimental setting we work on: even for the lowest sample gains ($G^{\text{test}}_{20}$ of {$4.1$ and $4.3$} for \texttt{Ridge} and \texttt{ExtraTrees} on the simulated pipeline), the gain is still rising at $K=5$.
This is at odds with the common {heuristic of stopping at $K$ splits when the test fraction is $1/K$, since $K$-fold CV then holds out every observation once}. \autoref{appendix:repeated-kfold} shows that our results are not dependent on the CV procedure: repeated $K$-fold{, \emph{i.e.}, $K$-fold CV} {rerun on fresh partitions, yields} comparable results.
This leads to the statement that \textbf{CV reduces benchmarking variance surprisingly well}.

Two further checks reinforce it. Stratifying {the splits by class}, a standard variance-reduction technique \citep{kohavi_cv_1995}, leaves variances and sample gains essentially unchanged {for the three CNNs on \texttt{PCam}, whose classes are balanced, under every metric. The exception is \texttt{ViT-B/16}{. In our protocol it} is often close to a constant predictor, whose score follows the class share of its test fold{. Stratification} alone divides its single-split variance by $2$ under accuracy and by $35$ to $48$ under the Brier score and the negative log-likelihood, and repeated splitting {still pays on top of it}} (\autoref{appendix:stratification}). And at very small training budgets, from $5$ to $100$ labels {on the simulated process}, the gain is still rising at $K=20$ for every algorithm we tested, by a factor of $1.1$ to $2.0$ between $K=5$ and $K=20$, though {it is lower at the smallest budgets}: the train/test coupling grows as labels become scarcer (\autoref{appendix:label-scarce}).

\paragraph{{Anticipating} how sample gains settle in}

\begin{wrapfigure}[23]{r}{0.45\textwidth}
    \vskip-0.5em
    \centering
    \includegraphics[width=\linewidth]{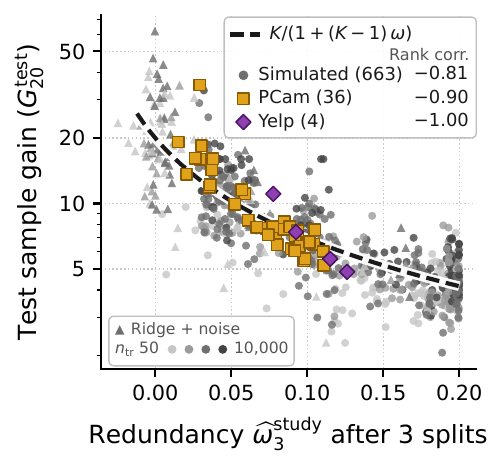}
    \captionsetup{justification=raggedright,singlelinecheck=false}
    \caption{Sample gain at $K=20$ against the redundancy after $3$ splits{, one point per simulated ({Appendix}~\ref{appendix:redundancy-runs}), \texttt{PCam} (\autoref{tab:pcam-redundancy}) or \texttt{Yelp} configuration}, each averaged over its runs{. Gains above $20$: anticorrelated fold errors or estimation noise ({Appendix}~\ref{sec:sample-gain-estimation})}.}
    \label{fig:redundancy-vs-gain}
\end{wrapfigure}

CV cost scales linearly with $K$. Knowing when additional splits are not worth computing is therefore a budgeting question, not just a statistical one.
{Equation~\eqref{eq:delta_cvvar} suggests that the growth of $G^{\text{test}}_K$ is linked with the covariance $\tau^{\mathrm{te}}$ between split-specific errors}.
To anticipate this behavior without relying on an outer or benchmarking set, {we define} the \emph{redundancy} $\widehat{\omega}^{\mathrm{study}}_k(F_\lambda,n_{\text{tr}})$, computed from the first $k$ splits of a single CV run {({Appendix}~\ref{sec:study-only-redundancy-statistics})}. For two splits $a<b$, let $I_{ab}$ denote the study observations held out by both. The redundancy is the correlation between the per-sample losses of the two split-specific predictors on $I_{ab}$, averaged over the pairs $a<b$ and multiplied by the test fraction $t = {n_{\text{te}}/(n_{\text{tr}}+n_{\text{te}})}$, the share of a test set that another split also tests. The loss is that of the metric whose gain is targeted. {Appendix}~\ref{sec:study-only-redundancy-statistics} derives the link $G^{\text{test}}_K \approx K/\bigl(1+(K-1)\,\omega\bigr)$ {for the population redundancy $\omega$}, up to a train/test coupling term. \autoref{fig:redundancy-vs-gain} shows it on $663$ simulated configurations: after three splits, the redundancy orders them by gain (rank correlation $-0.81$) and follows this curve, falling below it at small training sizes. It holds on real data as well: on $36$ \texttt{PCam} configurations ($4$ networks, $3$ sizes, $3$ metrics, each with its own per-sample loss), the rank correlation is $-0.90$ ($-0.92$ to $-0.95$ within each metric, \autoref{tab:pcam-redundancy}), and these points {and $4$ \texttt{Yelp} ones (accuracy)} follow the same curve.{}
High $\widehat{\omega}^{\mathrm{study}}_k$ values therefore indicate that {further splits} mostly average redundant information, making large {additional} gains unlikely. Conversely, low values {mean} that the splits err more independently: they do not guarantee a large gain, but {flag} settings where continuing CV {may} still pay off.

\Needspace{0.30\textheight}
\paragraph{Early stopping cross-validation} After three splits of a single CV run, or five for a reading about a third less noisy, one can compute the redundancy $\widehat{\omega}^{\mathrm{study}}_k$. If it is high, further repetitions are unlikely to bring large additional gains and CV should be early-stopped. If it is low, continuing CV remains worth considering. {The redundancy costs nothing beyond the fits already performed and, unlike racing (\autoref{appendix:significance-power}), needs only one candidate.}
The estimator, its derivation and a single-run validation on a pre-registered holdout of $243$ simulated configurations are in {Appendix}~\ref{sec:study-only-redundancy-statistics}: after three splits, the redundancy correlates with $\log G^{\text{test}}_{200}$ at $-0.65$ ($-0.73$ after five), and no run in its highest tertile reaches $G^{\text{test}}_{20} \ge 8$. Because the redundancy is estimated on the observations held out by two splits, those folds {need} a few observations: in our experiments it becomes a reliable per-study signal above a budget of roughly $20$ labels, that is, held-out folds of four to five observations (\autoref{appendix:label-scarce}). Below that budget no such signal is needed, since every algorithm still gains from {more} splits.
The sample gain itself is defined for any metric {computable} on a test set, AUC-ROC included. The redundancy is more demanding: {its} metric must decompose over observations{, excluding} AUC-ROC and $R^2$ (\autoref{appendix:metrics}).

\subsection{Pairwise ranking}
\label{sec:pairwise-ranking}

{Benchmarks often aim not only to estimate scores but to decide which learning algorithm to prefer, a complementary ranking problem.} {When} the oracle ranking {is} independent of the training size, {as for the families of algorithms of \autoref{appendix:ranking-sample-gain}, a ranking-equivalent sample gain can be defined and measured}. In our real-data experiments, however, {the ranking can depend on the training size, so we} evaluate rankings separately for each training size.

\begin{figure}[!tb]
    \centering
    \begin{subfigure}[t]{0.49\linewidth}
        \centering
        \includegraphics[width=\linewidth,height=.255\textheight,keepaspectratio]{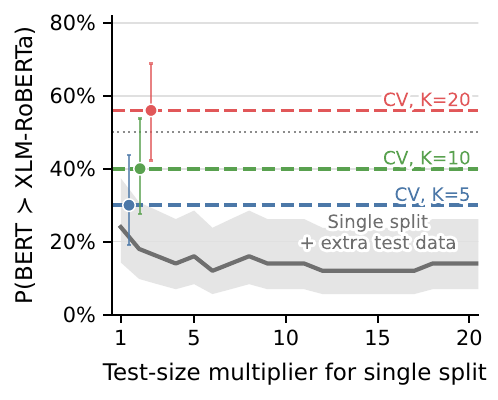}
        \caption{Proportion of retrieval of the quasi-oracle ranking.}
        \label{fig:nlp_ranking_retrieval}
    \end{subfigure}\hfill
    \begin{subfigure}[t]{0.49\linewidth}
        \centering
        \includegraphics[width=\linewidth,height=.255\textheight,keepaspectratio]{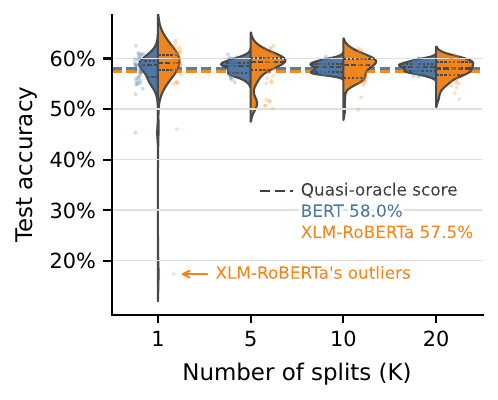}
        \caption{Per-seed test-score distributions.}
        \label{fig:nlp_ranking_violin}
    \end{subfigure}
    \caption{Pairwise ranking on \texttt{Yelp} at $n_{\text{tr}}=3{,}000$ {over $50$ seeds, with $95\%$ Wilson intervals in (a)}.}

    \label{fig:nlp_ranking}
\end{figure}

For each algorithm, we estimate its oracle {risk} at a given training size by averaging its benchmarking-set score over {seeds and splits}. This defines a \emph{quasi-oracle ranking}. We then test{, across seeds,} whether pairwise {quasi-oracle} score differences are statistically significant (\autoref{appendix:statistical_significance}). The results in \autoref{tab:stat_significance} show that \textbf{multi-split CV retrieves the quasi-oracle ranking more reliably than a single hold-out} when the benchmarking gap is statistically significant. Retrieving the order is not the same as certifying it: a significance test taken within a single CV run detects these differences far less often, and at our sample sizes more splits do not change this (\autoref{appendix:significance-power}).

\Needspace{5\baselineskip}
\paragraph{Ranking means is not the same as averaging rankings}

Our experiments on NLP also show {that CV changes what a ranking estimates}. 
Using \texttt{Yelp} with $n_{\text{tr}}=3{,}000$, the mean benchmarking score of \texttt{BERT} is {slightly} higher than that of \texttt{XLM-RoBERTa}. Yet single-split evaluations rank \texttt{XLM-RoBERTa} above \texttt{BERT} in roughly {70\%--85\%} of runs, even with {extra} test data (\autoref{fig:nlp_ranking_retrieval}).

The reason is visible in \autoref{fig:nlp_ranking_violin}: \texttt{XLM-RoBERTa} has severe low-score outliers. These learning failures lower its mean performance, although it often wins on individual splits. As $K$ increases, CV includes these failures and estimates \textbf{the ranking of mean scores rather than the mean split-wise ranking}. At $K=20$, {CV retrieves this ranking in $56\%$ of runs, against $24\%$ for a single split ($p=0.0025$, \autoref{appendix:statistical_significance}) and at most $18\%$ with extra test data: CV shows a ranking that a single split does not see}.

\section{Conclusion}
\label{sec:conclusion}

Our work shows that cross-validation can make machine-learning benchmarks much more reliable in small-data regimes. Through the variance-equivalent test sample gain, we quantify benefits of cross-validation in interpretable units: the amount of additional test data a single-split evaluation would need to reach comparable stability. Across synthetic regression, histopathology image classification, and NLP fine-tuning, repeated validation splits yield substantial gains, often far beyond the first few splits and the classic heuristic that each observation only needs to be held out once.

These gains are not uniform across algorithms. Some methods lead to benchmarks that settle quickly, whereas others keep benefiting from many more repetitions. We therefore introduce the redundancy, a study-only statistic computable after only a few splits of a single CV run.

{Our operational recommendation is simple: \textbf{split repeatedly}. When the task is classification, also stratify the splits: stratification is essentially free, but it does not substitute for repeated splitting.} Practitioners should use multi-split CV when computationally feasible, and rely on the redundancy, read after three to five splits,{} to decide whether larger $K$ is worth the cost. {As a complement, our} results {show} that multi-split CV improves algorithm ranking when benchmarking differences are statistically meaningful, and can expose learning failures hidden by single splits.

Limitations remain: because our protocol requires large benchmarking sets, our real-data experiments are constrained by dataset availability and computational cost, and cover only two domains. {A further limitation concerns label quality rather than quantity. Our simulated experiments span several target-noise levels, but the noise is {i.i.d.\ and part of the data distribution}, so the estimand is unchanged. Under weak supervision, where labels come from heuristics or non-expert annotators and are systematically rather than independently wrong, the estimand shifts from the oracle risk to agreement with the labelling function. We leave that setting to future work.} As automated systems have growing societal impact, rigorous algorithm selection is particularly critical in small-sample domains such as medical imaging. We contribute by making empirical {evaluations} more stable and uncertainty-aware.

\section*{Acknowledgements}
We thank Olivier Colliot for his valuable insights and discussions, especially for pointing us to the \texttt{PatchCamelyon} dataset.
We also thank Shan Raza, lead developer of \texttt{TIAToolbox}, for sharing with us the code used for training the pre-trained models of the library, which helped us in the experiments run on this same dataset.
This research was supported in part by the French National Research Agency (ANR) through the BenchArk project (ANR-24-IAS2-0003) as well as Hi! PARIS
and ANR/France 2030 program (ANR-23-IACL-0005). This work was performed using HPC resources from GENCI–IDRIS (Grant 2026-AD011015308R2).

\FloatBarrier

\bibliography{biblio}

\newpage
\FloatBarrier{}
\appendix

\counterwithin{figure}{section}
\def\thefigure{\thesection.\arabic{figure}}
\counterwithin{table}{section}
\def\thetable{\thesection.\arabic{table}}
\counterwithin{equation}{section}
\def\theequation{\thesection.\arabic{equation}}

\setcounter{tocdepth}{2}
\makeatletter
\let\appidx@lsec\l@section
\patchcmd{\appidx@lsec}{1.0em}{0.55em}{}{\errmessage{appidx: could not patch the section-entry gap}}
\definecolor{appidxname}{gray}{0.38}
\definecolor{appidxrule}{gray}{0.70}
\let\appidx@lsecfirst\l@section
\patchcmd{\appidx@lsecfirst}{\addpenalty\@secpenalty}{\nobreak}{}{\errmessage{appidx: could not patch the first-entry penalty}}
\patchcmd{\appidx@lsecfirst}{1.0em}{-1.5pt}{}{\errmessage{appidx: could not patch the first-entry gap}}
\newif\ifappidx@first
\newcommand{\appidxgroup}[1]{%
  \par\addpenalty{-\@highpenalty}\addvspace{1.0em \@plus\p@}%
  \hbox to\hsize{\small
    \if\relax\detokenize{#1}\relax\else
      {\color{appidxname}\textsc{\textls[60]{#1}}}\hskip0.6em\fi
    {\color{appidxrule}\leaders\hrule height \dimexpr 0.5ex+0.2pt\relax depth -\dimexpr 0.5ex-0.2pt\relax\hfill}}%
  \nobreak\global\appidx@firsttrue}
\newcommand*{\appidx@choose}{%
  \ifappidx@first\global\appidx@firstfalse\let\appidx@l\appidx@lsecfirst
  \else\let\appidx@l\appidx@lsec\fi}
\newcommand{\appidxsec}[2]{%
  \appidx@choose
  \appidx@l{\hyperref[#1]{\numberline{\ref*{#1}}#2}}{\hyperref[#1]{\pageref*{#1}}}}
\newcommand{\appidxsub}[2]{%
  \l@subsection{\hyperref[#1]{\numberline{\ref*{#1}}#2}}{\hyperref[#1]{\pageref*{#1}}}}
\newcommand{\appidxplain}[2]{%
  \appidx@choose
  \appidx@l{\hyperref[#1]{#2}}{\hyperref[#1]{\pageref*{#1}}}}
\makeatother

\section*{Contents of the appendices}
\begingroup\setlength{\parskip}{4.5pt}%
\appidxgroup{{Method details}}
\appidxsec{appendix:benchmarking-derivations}{Derivation of benchmarking statistics}
\appidxsub{appendix:cv-unbiasedness}{Unbiasedness of the CV estimator}
\appidxsub{appendix:cv-variance}{Variance of the CV estimator}
\appidxsub{appendix:cv-reshuffling-asymptotics}{{Asymptotic covariance of resampled estimates}}
\appidxsec{appendix:sample_gain_justification}{Measuring and approximating the evaluation noise: the justification of sample gain}
\appidxsec{sec:inter-fold-covariance-estimation}{Estimating inter-split covariance in cross-validation}
\appidxsub{sec:tau-estimation}{Raw fold scores: estimating $\tau$}
\appidxsub{sec:tau-te-estimation}{Benchmark-adjusted fold errors: estimating $\tau^{\mathrm{te}}$}
\appidxsec{sec:sample-gain-diagnostics}{Estimating and anticipating sample gains}
\appidxsub{sec:sample-gain-estimation}{Benchmark-aware estimation of sample gain $G^{\text{test}}_K$}
\appidxsub{sec:study-only-redundancy-statistics}{Study-only redundancy for early-stopping cross-validation}
\appidxsec{appendix:gain-intervals}{{Confidence intervals of the sample gains}}
\appidxgroup{{Experimental details}}
\appidxsec{appendix:experimental_details}{Experimental details}
\appidxsub{appendix:simulated-process}{Simulated data process}
\appidxsub{appendix:redundancy-runs}{Simulated runs used for the redundancy}
\appidxsub{appendix:real-training-loops}{Real data training loops}
\appidxsub{appendix:runs-and-time}{Runs{, time and assets}}
\appidxgroup{{Robustness of the gain}}
\appidxsec{appendix:hpo}{Sample gains under hyper-parameter optimisation}
\appidxsec{appendix:fixed-vs-reshuffled}{Repeated training runs versus resampled splits}
\appidxsec{appendix:repeated-kfold}{Comparison with repeated $5$-fold}
\appidxsec{appendix:stratification}{Stratified resampling}
\appidxsec{appendix:metrics}{Sample gains across evaluation metrics}
\appidxsec{appendix:label-scarce}{Behaviour at small label budgets}
\appidxgroup{{Comparing algorithms}}
\appidxsec{appendix:statistical_significance}{Statistical significance of pairwise rankings on real data}
\appidxsub{appendix:sig-formalization}{Formalization and methodology}
\appidxsub{appendix:sig-estimators}{{Test statistic}}
\appidxsub{appendix:sig-results}{Empirical results}
\appidxsec{appendix:significance-power}{{Significance tests and racing within a single cross-validation run}}
\appidxsec{appendix:ranking-sample-gain}{Varying training size: ranking-equivalent sample gain}
\appidxgroup{}
\appidxplain{appendix:checklist}{NeurIPS Paper Checklist}
\endgroup
\newpage

\section{Derivation of benchmarking statistics}
\label{appendix:benchmarking-derivations}

{This appendix restates in our notation the source results on the CV estimator: its unbiasedness ({Appendix}~\ref{appendix:cv-unbiasedness}) and its variance ({Appendix}~\ref{appendix:cv-variance}), on which \autoref{sec:resampling} relies, and the asymptotic covariance of \citet{nagler_reshuffling_2024} ({Appendix}~\ref{appendix:cv-reshuffling-asymptotics}), which gives the large-sample limit of the reference curve of the redundancy ({Appendix}~\ref{sec:study-only-redundancy-statistics}).}

\subsection{Unbiasedness of the CV estimator\texorpdfstring{ \citep{arlot_survey_2010}}{}}
\label{appendix:cv-unbiasedness}

\paragraph{Original formulation}
\citet{arlot_survey_2010} establish the unbiasedness of the cross-validation estimator in their Equation (13) as follows:
\begin{equation}
    \mathbb{E} \left[ \widehat{\mathcal{L}}^{\mathrm{CV}} \left( \mathcal{A}; D_n; \left(I^{(t)}_j\right)_{1 \leq j \leq B} \right) \right] = \mathbb{E} \left[ \mathcal{L}_P \left( \mathcal{A} \left( D_{n_t} \right) \right) \right].
    \tag{Arlot \& Celisse, Eq. 13}
\end{equation}
\textbf{Original notation:}
\begin{itemize}
    \item $\widehat{\mathcal{L}}^{\mathrm{CV}}$: the cross-validation estimator of the risk.
    \item $\mathcal{A}$: the learning algorithm.
    \item $D_n$: the full dataset of size $n$.
    \item $(I^{(t)}_j)_{1 \leq j \leq B}$: the sequence of $B$ training set indices (splits).
    \item $n_t$: the size of the training set (cardinality of $I^{(t)}_j$).
    \item $\mathcal{L}_P$: the population risk (loss).
    \item $D_{n_t}$: a training set of size $n_t$.
\end{itemize}

\paragraph{Translation to our notation}
Mapping the components $B \to K$, $n_t \to n_{\text{tr}}$, and $\mathcal{A} \to F_\lambda$, this confirms that the expectation of our MCCV estimator corresponds to the {oracle risk} for the specific training size $n_{\text{tr}}$:
\begin{equation}
    \mathbb{E}\bigl[\widehat{R}_{K}\bigr] = \mathcal{R}^*_{n_{\text{tr}}}(F_\lambda).
\end{equation}
{The identity only requires each split $\eta_k$ and each randomness $\xi_k$ to be drawn independently of the data, with $|\mathcal{D}^{\text{tr}}_{\eta_k}| = n_{\text{tr}}$. The training and test sets of a split are then independent samples of sizes $n_{\text{tr}}$ and $n_{\text{te}}$, so each $E_k$ has expectation $\mathcal{R}^*_{n_{\text{tr}}}(F_\lambda)$ by the second form of \eqref{eq:oracle_score}, and so does the average of any $K$ of them, even when the splits are not independent of one another. Splits stratified on the labels satisfy this only approximately.}

\subsection{Variance of the CV estimator\texorpdfstring{ \citep{nadeau_inference_2003}}{}}
\label{appendix:cv-variance}

\paragraph{Original formulation}
\citet{nadeau_inference_2003} derive the variance of the average estimator over $J$ splits in their Equation (8){, for $J$ training sets of size $n_1$ drawn independently of the data, each tested on its complement, of size $n_2 = n - n_1$}:
\begin{equation}
    \operatorname{Var}[\hat{\mu}_J] = \sigma_1 \left( \rho + \frac{1 - \rho}{J} \right).
    \tag{Nadeau \& Bengio, Eq. 8}
\end{equation}
\textbf{Original notation:}
\begin{itemize}
    \item $\hat{\mu}_J$: the average estimator over $J$ splits.
    \item $J$: the number of repetitions (splits).
    \item $\sigma_1$: the variance of the estimator on a single split.
    \item {$\sigma_2$: the covariance between the estimators of two different splits.}
    \item $\rho$: the correlation coefficient between the estimators of two different splits{, $\rho = \sigma_2/\sigma_1$}.
\end{itemize}

\paragraph{Translation to our notation}
We substitute $J \to K$, $\hat{\mu}_J \to \widehat{R}_{K}$, {$\sigma_2 \to \tau$} and $\sigma_1 \to \sigma^2_{\text{HO}}${, so that} $\tau = \rho \sigma^2_{\text{HO}}$.
Expanding the terms yields the decomposition into reducible error (scaled by $1/K$) and irreducible covariance:
\begin{equation}
    \operatorname{Var}[\widehat{R}_{K}]
    = \sigma^2_{\text{HO}} \left( \frac{\tau}{\sigma^2_{\text{HO}}} + \frac{1}{K} \left(1 - \frac{\tau}{\sigma^2_{\text{HO}}}\right) \right)
    = \frac{1}{K} \sigma^2_{\text{HO}} + \frac{K-1}{K} \tau.
\end{equation}
{The decomposition only needs fold scores with a common variance and a common pairwise covariance (their Lemma~1), so it holds for any learner and any loss, and $0 \le \tau \le \sigma^2_{\text{HO}}$ (their Proposition~1). Two other results of theirs are used in \autoref{sec:resampling}. No estimator of $\operatorname{Var}[\hat\mu_J]$ that is linear and/or quadratic in the individual test losses is unbiased in general (their Proposition~3). Their Section~3.1 approximates $\rho$ by $n_2/(n_1+n_2)$, {the test fraction $t$,} its value if the test losses did not depend on the training set, which gives \eqref{eq:nadeau_approx}.}

\subsection{{Asymptotic covariance of resampled estimates}\texorpdfstring{ \citep{nagler_reshuffling_2024}}{}}
\label{appendix:cv-reshuffling-asymptotics}

\paragraph{Original formulation}
\citet{nagler_reshuffling_2024} establish the limiting distribution of the estimation error in their Theorem 2.1:
\begin{equation}
    \sqrt{n} \left( \widehat{\mu}(\lambda_j) - \mu(\lambda_j) \right)_{j=1}^J \xrightarrow{d} \mathcal{N}(0, \Sigma),
    \tag{Nagler et al., Thm 2.1}
\end{equation}
where the covariance matrix $\Sigma$ is defined by:
\begin{equation}
    \Sigma_{i,j} = \tau_{i,j,M} K(\lambda_i, \lambda_j)
\end{equation}
\textbf{Original notation:}
\begin{itemize}
    \item $n$: the size of the observed dataset.
    \item {$\alpha$: the validation fraction of each of the $M$ splits, which are drawn independently of the data.}
    \item $\widehat{\mu}(\lambda_j)$: the $M$-split validation loss for configuration $\lambda_j$.
    \item $\mu(\lambda_j)$: {the expected loss of the model trained on the whole dataset, of size $n$.}
    \item $J$: the number of hyperparameter configurations evaluated.
    \item $\tau_{i,j,M}$: {a design factor that counts, suitably normalised, how often configurations $i$ and $j$ are validated on the same observations. It equals $1/\alpha$ for a single hold-out.}
    \item $K(\lambda_i, \lambda_j)$: {the asymptotic covariance between the losses of configurations $i$ and $j$ on one observation. We write it $C_{i,j}$, to avoid a clash with our number of splits $K$.}
\end{itemize}
{The theorem assumes a bounded and stable loss, and a risk that converges faster than $n^{-1/2}$ to that of a fixed rule (their Eqs.~(4) and~(5)).}

\paragraph{Translation to our notation}
While \citet{nagler_reshuffling_2024} frame their analysis around optimizing over a set of hyperparameter configurations $\{\lambda_1, \dots, \lambda_J\}$, this mathematical framework applies identically to benchmarking a discrete set of fixed learning algorithms $\{F_1, \dots, F_J\}${, provided that each satisfies these assumptions (for an algorithm with internal randomness, they must hold for the randomised rule)}.

By making the substitution $M \to K$ (number of splits), mapping $\lambda_j$ to the $j$-th learning algorithm $F_j$, replacing {$\mu(\lambda_j)$, the oracle risk at training size $n$,} with our oracle {risk} $\mathcal{R}^*_{n_{\text{tr}}}(F_j)$ {(under that condition, both are within $o(n^{-1/2})$ of the risk of the limiting rule)}, and replacing the $M$-split validation loss $\widehat{\mu}(\lambda_j)$ with our MCCV estimator $\widehat{R}_{K}(F_j)$, Theorem 2.1 can be rewritten for our benchmarking context:
\begin{equation}
    \sqrt{n} \left( \widehat{R}_{K}(F_j) - \mathcal{R}^*_{n_{\text{tr}}}(F_j) \right)_{j=1}^J \xrightarrow{d} \mathcal{N}(0, \Sigma) \enspace.
\end{equation}
{When all algorithms are evaluated on the same $K$ MCCV splits, as in our protocol, their Table~1 gives $\tau_{i,j,K} = 1 + (1-\alpha)/(K\alpha)$, so that
\begin{equation}
    \Sigma_{i,j} = \Bigl(1 + \frac{1-\alpha}{K\alpha}\Bigr) C_{i,j}
\end{equation}
(splits redrawn for each algorithm, their reshuffled variant, give $\tau_{i,j,K} = 1$ for $i \neq j$). For $K = 1$ the diagonal gives $\sigma^2_{\text{HO}} \approx C_{j,j}/n_{\text{te}}$, and matching it with \eqref{eq:cvvar} for $K > 1$ gives $\tau \approx C_{j,j}/n$, hence $\tau/\sigma^2_{\text{HO}} \to {n_{\text{te}}/n} = \alpha = t$. In this regime \eqref{eq:nadeau_approx} becomes exact and $G^{\text{test}}_K \to K/(1+(K-1)t)$, the reference curve \eqref{eq:gain-redundancy} with $\rho_e = 1$ and $\kappa = 0$. Departures from this curve come from the dependence of the losses on the training data, which Theorem~2.1 treats as higher order.}

\section{Measuring and approximating the evaluation noise: the justification of sample gain}
\label{appendix:sample_gain_justification}

{This appendix details the derivations of \autoref{sec:sample_gain}: the evaluation-noise approximation \eqref{eq:delta_variance_approx}, the exact variance expansion \eqref{eq:delta_cvvar} for MCCV, and the gain of a fixed split repeated $K$ times, which \autoref{appendix:fixed-vs-reshuffled} tests.}

\paragraph{Single-split benchmark-adjusted error}

{With $g$, $\widehat{R}^{\star}(g)$ and $\delta_{\mathrm{HO}}$ as in \eqref{eq:delta_oracle_estimator}, the law of total variance \eqref{eq:total_variance_process} splits the variance of $R_{\mathcal{D}^{\mathrm{te}}}(g)$ into (A) the expected evaluation noise, due to the finite test set, and (B) the training instability $\operatorname{Var}_{g}[\mathcal{R}^{*}(g)]$, which measures how much the {oracle} risk of $g$ varies across training sets and random {states}.}

By definition, $\delta_{\mathrm{HO}}$ subtracts an estimator of the {oracle} risk from the finite-test-set score. Applying the law of total variance to $\delta_{\mathrm{HO}}$ gives:
\begin{equation}
    \operatorname{Var}[\delta_{\mathrm{HO}}]
    =
    \mathbb{E}_{g}\!\left[
        \operatorname{Var}[\delta_{\mathrm{HO}} \mid g]
    \right]
    +
    \operatorname{Var}_{g}\!\left[
        \mathbb{E}[\delta_{\mathrm{HO}} \mid g]
    \right].
    \label{app:eq:delta_total_variance}
\end{equation}

Since $\widehat{R}^{\star}(g)$ is unbiased conditional on $g$,
\begin{equation}
\begin{split}
    \mathbb{E}[\delta_{\mathrm{HO}} \mid g]
    &=
    \mathbb{E}\!\left[
        R_{\mathcal{D}^{\mathrm{te}}}(g)
        -
        \widehat{R}^{\star}(g)
        \mid g
    \right] \\
    &=
    \mathcal{R}^{*}(g)
    -
    \mathcal{R}^{*}(g)
    =
    0.
\end{split}
\end{equation}

Thus the conditional-mean term vanishes in the single-split case. Moreover, because $\widehat{R}^{\star}(g)$ is constructed independently of $\mathcal{D}^{\mathrm{te}}$ conditional on $g$:
\begin{equation}
\begin{split}
    \operatorname{Var}[\delta_{\mathrm{HO}}]
    &=
    \mathbb{E}_{g}\!\left[
        \operatorname{Var}[\delta_{\mathrm{HO}} \mid g]
    \right]\\
    &=
    \mathbb{E}_{g}\!\left[
        \operatorname{Var}\!\left[
            R_{\mathcal{D}^{\mathrm{te}}}(g)
            \mid g
        \right]
    \right]
    +
    \mathbb{E}_{g}\!\left[
        \operatorname{Var}\!\left[
            \widehat{R}^{\star}(g)
            \mid g
        \right]
    \right].
\end{split}
\label{app:eq:delta_variance_exact_single}
\end{equation}

\paragraph{Evaluation-noise approximation}

Suppose that the expected evaluation noise of the oracle-{risk} estimator $\widehat{R}^{\star}(g)$ is negligible compared to that of the test set evaluation. In our experiments, $\widehat{R}^{\star}(g)$ is computed on a large benchmarking set. Under this approximation,
\begin{equation}
    \operatorname{Var}[\delta_{\mathrm{HO}}]
    \approx
    \mathbb{E}_{g}\!\left[
        \operatorname{Var}\!\left[
            R_{\mathcal{D}^{\mathrm{te}}}(g)
            \mid g
        \right]
    \right].
    \label{app:eq:delta_variance_approx}
\end{equation}
Thus $\operatorname{Var}[\delta_{\mathrm{HO}}]$ estimates the finite-test-sample evaluation noise, independently of the training-instability component.

\paragraph{Exact variance expansion for MCCV}

{For the MCCV procedure, let $g_k$, $\delta_k$ and $\Delta_K = \frac{1}{K}\sum_{k=1}^{K}\delta_k$ be as in \autoref{sec:sample_gain}.}

The variance decomposition used in the main text does not require conditioning on the learned predictors. It follows exactly from the unconditional covariance expansion. Under the MCCV split generator, the fold-level errors $\delta_1,\ldots,\delta_K$ are exchangeable. Hence, for any $k$,
\begin{equation}
    \operatorname{Var}(\delta_k)
    =
    \operatorname{Var}(\delta_{\mathrm{HO}}),
\end{equation}
and for any distinct $i\neq j$ we define the unconditional benchmark-adjusted inter-split covariance
\begin{equation}
    \tau^{\mathrm{te}}
    =
    \operatorname{Cov}(\delta_i,\delta_j).
    \label{app:eq:tau_te_unconditional}
\end{equation}
Then
\begin{equation}
\begin{split}
    \operatorname{Var}(\Delta_K)
    &=
    \operatorname{Var}\!\left[
        \frac{1}{K}\sum_{k=1}^{K}\delta_k
    \right] \\
    &=
    \frac{1}{K^2}
    \sum_{k=1}^{K}
    \operatorname{Var}(\delta_k)
    +
    \frac{1}{K^2}
    \sum_{\substack{i,j=1\\ i\neq j}}^{K}
    \operatorname{Cov}(\delta_i,\delta_j) \\
    &=
    \frac{1}{K}\operatorname{Var}(\delta_{\mathrm{HO}})
    +
    \frac{K-1}{K}\tau^{\mathrm{te}}.
\end{split}
\label{app:eq:DeltaK_variance_exact}
\end{equation}
This identity is exact. It includes all dependence induced by reusing the same finite study sample across MCCV splits: overlapping test sets, overlap between one fold's test set and another fold's training set, shared splitting randomness, and any residual dependence introduced by the benchmark correction.

\paragraph{Shared and fit-specific evaluation noise}
Conditioning on the split $(\mathcal{D}^{\mathrm{tr}}, \mathcal{D}^{\mathrm{te}})$, so that only the internal randomness $\xi$ remains, the law of total variance splits the single-split variance into the two parts of \eqref{eq:evaluation-noise-split}:
\begin{equation}
    \operatorname{Var}[\delta_{\mathrm{HO}}]
    = \underbrace{\operatorname{Var}\bigl[\mathbb{E}[\delta_{\mathrm{HO}} \mid \mathcal{D}^{\mathrm{tr}}, \mathcal{D}^{\mathrm{te}}]\bigr]}_{(\mathrm{A}_1)}
    + \underbrace{\mathbb{E}\bigl[\operatorname{Var}[\delta_{\mathrm{HO}} \mid \mathcal{D}^{\mathrm{tr}}, \mathcal{D}^{\mathrm{te}}]\bigr]}_{(\mathrm{A}_2)}.
\end{equation}
Now let $\delta_1, \dots, \delta_K$ be the errors of $K$ fits on the same split, with independent internal randomness $\xi_1, \dots, \xi_K$. Given the split, they are independent and identically distributed, with conditional mean $\mathbb{E}[\delta_{\mathrm{HO}} \mid \mathcal{D}^{\mathrm{tr}}, \mathcal{D}^{\mathrm{te}}]$ and conditional variance $\operatorname{Var}[\delta_{\mathrm{HO}} \mid \mathcal{D}^{\mathrm{tr}}, \mathcal{D}^{\mathrm{te}}]$, so that
\begin{equation}
    \operatorname{Var}\Bigl[\frac{1}{K}\sum_{k=1}^{K}\delta_k\Bigr] = (\mathrm{A}_1) + \frac{(\mathrm{A}_2)}{K}.
    \label{app:eq:frozen-split-variance}
\end{equation}
Two fits on the same split therefore have errors with correlation $\rho_{\mathrm{fix}} = (\mathrm{A}_1)/\bigl((\mathrm{A}_1)+(\mathrm{A}_2)\bigr)$. If the single-split variance decreases as the inverse of the test size, matching it with \eqref{app:eq:frozen-split-variance} gives the gain of a fixed split repeated $K$ times, $G^{\text{test}}_K \approx K/\bigl(1+(K-1)\,\rho_{\mathrm{fix}}\bigr)$: it equals $1$ for a deterministic algorithm, for which $(\mathrm{A}_2) = 0$, and it never exceeds $1/\rho_{\mathrm{fix}}$, however many fits are averaged.

\FloatBarrier{}
\section{Estimating inter-split covariance in cross-validation}
\label{sec:inter-fold-covariance-estimation}

This appendix derives {estimators of the two inter-split covariances of the main text, $\tau$ in \eqref{eq:cvvar} and $\tau^{\mathrm{te}}$ in \eqref{eq:delta_cvvar}, from independent replications of a CV run}. {Throughout, a fold is the test part of a split, and fold scores and errors are those of the split's predictor on it.} We first estimate the raw inter-split covariance $\tau$ from fold scores, then apply the same ANOVA logic to benchmark-adjusted fold errors to estimate $\tau^{\mathrm{te}}$, the covariance that governs the variance-ratio form of the test sample gain ({Appendix}~\ref{sec:sample-gain-estimation}).

\subsection{Raw fold scores: estimating \texorpdfstring{$\tau$}{tau}}
\label{sec:tau-estimation}

\paragraph{Model}
Fix an algorithm, dataset, and training-set size. For each independent seed $s = 1,\dots,S$, a $K$-split CV produces raw fold scores $E_{s,1}, \dots, E_{s,K}$, the seed-indexed version of the main-text fold score $E_k$. We model the scores as exchangeable within a seed, with
\begin{equation}\label{eq:tau-model}
  \operatorname{Var}(E_{s,k}) = \sigma^2_{\text{HO}},
  \qquad
  \operatorname{Cov}(E_{s,k}, E_{s,\ell}) = \tau
  \quad (k \neq \ell),
\end{equation}
where $\sigma^2_{\text{HO}}$ denotes the marginal variance of a single fold score (equivalently, under the matched test-set geometry used throughout this paper, the variance of a single held-out evaluation). Seeds are mutually independent. The parameter $\tau$ captures the irreducible covariance between fold scores that arises because {the splits reuse one study set: their training sets largely coincide, their test sets overlap, and an observation tested by one split may be trained on by another} \citep{nadeau_inference_2003}.

\paragraph{Sufficient statistics}
For each seed~$s$, define the fold mean and the within-seed sample
variance:
\begin{equation}\label{eq:seed-stats}
  \bar{E}_s = \frac{1}{K}\sum_{k=1}^{K} E_{s,k},
  \qquad
  V_s = \frac{1}{K-1}\sum_{k=1}^{K} (E_{s,k} - \bar{E}_s)^2.
\end{equation}

Under the model~\eqref{eq:tau-model} (with no further distributional
assumptions beyond finite fourth moments), their expectations are:
\begin{align}
  \mathbb{E}[V_s]
    &= \sigma^2_{\text{HO}} - \tau
     \;\eqqcolon\; \sigma^2_{\text{HO},\mathrm{within}},
      \label{eq:EV}\\[4pt]
  \mathbb{E}\!\bigl[\bar{E}_s\bigr]
    &= \mu
      \quad\text{(some common mean)}, \label{eq:EXbar}\\[2pt]
  \operatorname{Var}\!\bigl(\bar{E}_s\bigr)
    &= \frac{1}{K}\sigma^2_{\text{HO}} + \frac{K-1}{K}\,\tau.
     \label{eq:VarEbar}
\end{align}
The shorthand $\sigma^2_{\text{HO},\mathrm{within}}$ in~\eqref{eq:EV} denotes the within-seed residual component of the raw fold-score variance. Equation~\eqref{eq:EV} follows from expanding the within-seed variance and using the compound-symmetry covariance:
$\mathbb{E}[V_s]
  = \frac{1}{K-1}\bigl[K\sigma^2_{\text{HO}} - K\bigl(\frac{\sigma^2_{\text{HO}}}{K} +
    \frac{K-1}{K}\tau\bigr)\bigr]
  = \sigma^2_{\text{HO}} - \tau.$

\paragraph{ANOVA-type estimators}
Averaging over seeds gives the ``within'' and ``between'' estimators:
\begin{equation}\label{eq:WB}
  \hat{W} = \frac{1}{S}\sum_{s=1}^{S} V_s,
  \qquad
  \hat{B} = \frac{1}{S-1}\sum_{s=1}^{S}
             (\bar{E}_s - \bar{\bar{E}})^2,
  \qquad
  \bar{\bar{E}} = \frac{1}{S}\sum_{s=1}^{S}\bar{E}_s.
\end{equation}

By~\eqref{eq:EV} and~\eqref{eq:VarEbar},
\begin{equation}\label{eq:EWB}
  \mathbb{E}[\hat{W}] = \sigma^2_{\text{HO}} - \tau,
  \qquad
  \mathbb{E}[\hat{B}] = \frac{\sigma^2_{\text{HO}} + (K-1)\tau}{K},
\end{equation}
so the method-of-moments (ANOVA) estimator of~$\tau$ is
\begin{equation}\label{eq:tau-hat}
  \boxed{\;
  \hat\tau = \hat{B} - \frac{\hat{W}}{K}\;},
\end{equation}
with the companion estimator of $\sigma^2_{\text{HO}}$,
\begin{equation}\label{eq:sigma-HO-K-hat}
  \hat{\sigma}^2_{\text{HO}\mid K}
  \;\coloneqq\; \hat{W} + \hat{\tau},
\end{equation}
where the subscript ``$\mid K$'' signals that this estimator is built \emph{knowing} a {$K$-split} CV run has been observed ($K \ge 2$). A complementary estimator $\hat{\sigma}^2_{\text{HO}\mid 1}$ built from single-split replicates is introduced in {Appendix}~\ref{sec:sample-gain-estimation}. Unbiasedness of $\hat\tau$ is immediate:
$\mathbb{E}[\hat\tau]
  = \frac{\sigma^2_{\text{HO}} + (K-1)\tau}{K} - \frac{\sigma^2_{\text{HO}} - \tau}{K}
  = \tau.$

\medskip\noindent
\textbf{Remark.}
$\hat{B}$ alone is already an unbiased estimator of $\operatorname{Var}(\bar{E}_s)$, i.e.\ the variance of the $K$-split CV estimate. $\hat\tau$ refines this by subtracting the $\sigma^2_{\text{HO}}/K$ contribution, isolating the floor $\lim_{K\to\infty}\operatorname{Var}(\bar{E}_s) = \tau$.

\bigskip

\paragraph{Variance of $\hat\tau$}
Since $\hat\tau = \hat{B} - \hat{W}/K$ and the two statistics are computed from the \emph{same} fold scores, we need $\operatorname{Var}(\hat{B})$, $\operatorname{Var}(\hat{W})$, and $\operatorname{Cov}(\hat{B}, \hat{W})$. Write $\theta_s = (\bar{E}_s, V_s)$ for the per-seed summary. These are i.i.d.\ across seeds.

\medskip\noindent
\emph{Variance of $\hat{B}$.}\quad
$\hat{B}$ is the sample variance ($\mathrm{ddof}=1$) of the i.i.d.\ random variables $\bar{E}_1, \dots, \bar{E}_S$, each with variance $\gamma \coloneqq \operatorname{Var}(\bar{E}_s) = (\sigma^2_{\text{HO}} + (K-1)\tau)/K$ and fourth central moment $\mu_4 \coloneqq \mathbb{E}[(\bar{E}_s - \mu)^4]$.

Write $(S-1)\hat{B} = \sum_s (\bar{E}_s-\mu)^2 - S(\bar{\bar{E}}-\mu)^2$.  The three needed variances and covariance are:
\begin{align}
  \operatorname{Var}\!\Bigl(\textstyle\sum_s (\bar{E}_s-\mu)^2\Bigr)
    &= S(\mu_4 - \gamma^2), \label{eq:var-sum-sq}\\[3pt]
  \operatorname{Var}\!\bigl(S(\bar{\bar{E}}-\mu)^2\bigr)
    &= \frac{\mu_4}{S} + \frac{2S-3}{S}\,\gamma^2, \label{eq:var-grand-sq}\\[3pt]
  \operatorname{Cov}\!\Bigl(\textstyle\sum_s (\bar{E}_s-\mu)^2,\;
    S(\bar{\bar{E}}-\mu)^2\Bigr)
    &= \mu_4 - \gamma^2, \label{eq:cov-sq}
\end{align}
all obtained by expanding in terms of $\mathbb{E}[(\bar{E}_s-\mu)^4] = \mu_4$ and $\mathbb{E}[(\bar{E}_s-\mu)^2] = \gamma$ and using independence across seeds. In particular, $\mathbb{E}[(\sum_s (\bar{E}_s-\mu))^4] = S\mu_4 + 3S(S-1)\gamma^2$ (only the diagonal and pairwise-square terms survive).
Combining via
$\operatorname{Var}((S\!-\!1)\hat{B})
= \operatorname{Var}(\sum (\bar{E}_s-\mu)^2)
  + \operatorname{Var}(S(\bar{\bar{E}}\!-\!\mu)^2)
  - 2\operatorname{Cov}$
and simplifying:
\begin{equation}\label{eq:VarB}
  \operatorname{Var}(\hat{B})
  = \frac{1}{S}\!\left(\mu_4
    - \frac{S-3}{S-1}\,\gamma^2\right).
\end{equation}
Under Gaussian fold scores, $\bar{E}_s$ is Gaussian so $\mu_4 = 3\gamma^2$ (kurtosis {$3$}), and this reduces to $\operatorname{Var}(\hat{B}) = 2\gamma^2/(S-1)$.

\medskip\noindent
\emph{Variance of $\hat{W}$.}\quad
$\hat{W} = \frac{1}{S}\sum_s V_s$ is a sample mean of i.i.d.\ terms with mean $\sigma^2_{\text{HO}} - \tau$ and variance $\operatorname{Var}(V_s)$. Hence:
\begin{equation}\label{eq:VarW}
  \operatorname{Var}(\hat{W})
  = \frac{\operatorname{Var}(V_s)}{S}.
\end{equation}
Under joint normality of $(E_{s,1},\dots,E_{s,K})$ with the compound-symmetry covariance $\Sigma = (\sigma^2_{\text{HO}}-\tau)I_K + \tau\mathbf{1}\mathbf{1}^\top$, the quadratic form $(K-1)V_s = \mathbf{E}_s^\top \mathbf{H} \mathbf{E}_s$ (where $\mathbf{H} = I_K - \frac{1}{K}\mathbf{1}\mathbf{1}^\top$) satisfies $(K-1)V_s/(\sigma^2_{\text{HO}} - \tau) \sim \chi^2_{K-1}$, giving $\operatorname{Var}(V_s) = 2(\sigma^2_{\text{HO}} - \tau)^2/(K-1)$, so
\begin{equation}\label{eq:VarW-explicit}
  \operatorname{Var}(\hat{W})
  = \frac{2(\sigma^2_{\text{HO}} - \tau)^2}{S(K-1)}.
\end{equation}

\medskip\noindent
\emph{Covariance of $\hat{B}$ and $\hat{W}$.}\quad
Under normality, $\bar{E}_s$ and $V_s$ are independent within each seed (a consequence of Fisher--Cochran: the projection onto $\operatorname{span}(\mathbf{1})$ is orthogonal to the range of $\mathbf{H}$, and since $\mathbf{1}$ is an eigenvector of $\Sigma$, the two resulting Gaussian components are uncorrelated and hence independent). Since $\hat{B}$ depends only on $(\bar{E}_1,\dots,\bar{E}_S)$ and $\hat{W}$ only on $(V_1,\dots,V_S)$, they are independent, and $\operatorname{Cov}(\hat{B}, \hat{W}) = 0$.

\medskip\noindent
\emph{Combining.}\quad
Under normality,
\begin{equation}\label{eq:VarTau}
  \boxed{\;
  \operatorname{Var}(\hat\tau)
  = \operatorname{Var}(\hat{B})
    + \frac{1}{K^2}\operatorname{Var}(\hat{W})
  = \frac{2(\sigma^2_{\text{HO}} + (K-1)\tau)^2}{K^2(S-1)}
    + \frac{2(\sigma^2_{\text{HO}} - \tau)^2}{S\,K^2(K-1)}
  \;}.
\end{equation}

\paragraph{Interpretation}
As $K$ grows, the second term (from $\hat{W}$) vanishes as $O(K^{-3})$, while the first (from $\hat{B}$) dominates and converges to $2\tau^2/(S-1)$. The precision of $\hat\tau$ is then ultimately limited by how many independent seeds~$S$ we observe.

\paragraph{Plug-in and bootstrap}
In practice, we replace $(\sigma^2_{\text{HO}}, \tau, \mu_4)$ by their sample counterparts $(\hat{\sigma}^2_{\text{HO}\mid K}, \hat\tau, \hat\mu_4)$ to evaluate~\eqref{eq:VarTau}. Because the Gaussian kurtosis {of $3$} may not hold, our implementation supplements the plug-in formula with a non-parametric bootstrap over seeds: resample $(\bar{E}_s, V_s)_{s=1}^{S}$ with replacement $100$~times, recompute $\hat\tau$ for each replicate, and report percentile confidence intervals. This is valid because the seed-level summaries are i.i.d., making the ordinary bootstrap consistent for their functionals.

\paragraph{Relation to the Bengio--Grandvalet impossibility}
\citet{bengio_unbiased_2004} show that no {estimator of $\operatorname{Var}(\bar{E}_s)$ built from the errors of a single $K$-fold CV run is unbiased under all data distributions}. Our estimator does not contradict this result: it requires $S \geq 2$ \emph{independent replications} of the entire {$K$-split} procedure (each on a fresh random seed), making
$\bar{E}_1, \dots, \bar{E}_S$ an i.i.d.\ sample whose variance is trivially estimable. The Bengio--Grandvalet impossibility applies to the single-run regime ($S=1$), which is the common practical setting but not ours.

\subsection{Benchmark-adjusted fold errors: estimating \texorpdfstring{$\tau^{\mathrm{te}}$}{tau te}}
\label{sec:tau-te-estimation}

\paragraph{From raw fold scores to benchmark-adjusted deltas}
The raw-score inter-split covariance~$\tau$ of {Appendix}~\ref{sec:tau-estimation} {is dominated by the quality of the predictors, which varies with the study set: within one seed, the $K$ predictors are learned from overlapping study data, so their raw scores rise and fall together. Subtracting each predictor's benchmarking score removes this component. {Appendix}~\ref{sec:study-only-redundancy-statistics} splits the remaining covariance $\tau^{\mathrm{te}}$ into a test-overlap term, the redundancy, and a train/test coupling.}
In practice, we care about how much each fold score \emph{disagrees with a benchmark}, not about the raw fold similarity. Define the benchmark-adjusted fold score
\begin{equation}\label{eq:delta-def}
  \delta_{s,k}
  = E_{s,k} - \widehat{R}^{\star}_{s,k}(g_{s,k}),
\end{equation}
where $E_{s,k}=R_{\mathcal{D}^{\mathrm{te}}_{s,k}}(g_{s,k})$ is the raw test score of the fold-specific predictor $g_{s,k}=F_\lambda(\mathcal{D}^{\mathrm{tr}}_{s,k},\xi_{s,k})$, and $\widehat{R}^{\star}_{s}(g_{s,k})$ is its benchmarking-set estimate, computed independently of the fold test set. The inter-fold covariance on~$\delta$,
\begin{equation}\label{eq:tau-te-def}
  \tau^{\mathrm{te}} = \operatorname{Cov}(\delta_{s,k}, \delta_{s,\ell})
  \quad (k \neq \ell),
\end{equation}
is the fold covariance that remains \emph{after} the benchmark has cancelled the training-side drift. 

By the law of total covariance, this unconditional covariance can be written as:
\begin{equation}
\tau^{\mathrm{te}} = \operatorname{Cov}(\delta_{s,k},\delta_{s,\ell})
=
\mathbb{E}_{G}\!\left[
    \operatorname{Cov}(\delta_{s,k},\delta_{s,\ell}\mid G)
\right]
+
\operatorname{Cov}_{G}\!\left(
    \mathbb{E}[\delta_{s,k}\mid G],
    \mathbb{E}[\delta_{s,\ell}\mid G]
\right).
\end{equation}
The second term need not vanish in MCCV, because observations held out in one split may appear in
the training sets of other split predictors. Our estimator targets the unconditional covariance, which
is the quantity entering the exact variance formula for $\Delta_K$ (see \eqref{eq:delta_cvvar}).

\paragraph{Model}
Fix an algorithm, dataset, training-set size, and benchmarking set. For each independent seed~$s = 1, \dots, S$, the $K$-split CV and the benchmark evaluation together produce $\delta_{s,1}, \dots, \delta_{s,K}$. We model these as exchangeable within a seed, with
\begin{equation}\label{eq:tau-te-model}
  \operatorname{Var}(\delta_{s,k}) = \sigma^{2,\mathrm{te}},
  \qquad
  \operatorname{Cov}(\delta_{s,k}, \delta_{s,\ell}) = \tau^{\mathrm{te}}
  \quad (k \neq \ell),
\end{equation}
and seeds mutually independent. Here $\sigma^{2,\mathrm{te}}$ is the bench-adjusted analogue of $\sigma^2_{\text{HO}}$: the marginal variance of a single $\delta$ fold score, equivalently the variance of a single bench-adjusted held-out evaluation under the matched test-set geometry used throughout this paper. Exchangeability is therefore an explicit working model for these benchmark-adjusted fold scores. It is appropriate when the split-generation and benchmark-evaluation procedures are symmetric in the fold index $k$.

\paragraph{Sufficient statistics}
Mirroring~\eqref{eq:seed-stats}, define the per-seed delta mean and within-seed delta variance:
\begin{equation}\label{eq:seed-stats-delta}
  \bar{\delta}_s = \frac{1}{K}\sum_{k=1}^{K} \delta_{s,k},
  \qquad
  V^\delta_s
    = \frac{1}{K-1}\sum_{k=1}^{K} (\delta_{s,k} - \bar{\delta}_s)^2.
\end{equation}
Under~\eqref{eq:tau-te-model},
\begin{align}
  \mathbb{E}[V^\delta_s]
    &= \sigma^{2,\mathrm{te}} - \tau^{\mathrm{te}}
     \;\eqqcolon\; \sigma^{2,\mathrm{te}}_{\mathrm{within}},
      \label{eq:EV-delta}\\[3pt]
  \operatorname{Var}(\bar{\delta}_s)
    &= \frac{\sigma^{2,\mathrm{te}} + (K-1)\tau^{\mathrm{te}}}{K}.
    \label{eq:VarDeltaBar}
\end{align}

\paragraph{ANOVA-type estimators}
Averaging across seeds:
\begin{equation}\label{eq:WB-delta}
  \hat{W}_\delta
    = \frac{1}{S}\sum_{s=1}^{S} V^\delta_s,
  \qquad
  \hat{B}_\delta
    = \frac{1}{S-1}\sum_{s=1}^{S}
      (\bar{\delta}_s - \bar{\bar{\delta}})^2,
  \qquad
  \bar{\bar{\delta}}
    = \frac{1}{S}\sum_{s=1}^{S} \bar{\delta}_s,
\end{equation}
with expectations
\begin{equation}\label{eq:EWB-delta}
  \mathbb{E}[\hat{W}_\delta] = \sigma^{2,\mathrm{te}} - \tau^{\mathrm{te}},
  \qquad
  \mathbb{E}[\hat{B}_\delta]
    = \frac{\sigma^{2,\mathrm{te}} + (K-1)\tau^{\mathrm{te}}}{K},
\end{equation}
yielding the method-of-moments estimator
\begin{equation}\label{eq:tau-te-hat}
  \boxed{\;
  \hat{\tau}^{\mathrm{te}}
    = \hat{B}_\delta - \frac{\hat{W}_\delta}{K}
  \;},
\end{equation}
and companion estimator of $\sigma^{2,\mathrm{te}}$,
\begin{equation}\label{eq:sigma-te-K-hat}
  \hat{\sigma}^{2,\mathrm{te}}_{\mid K}
  \;\coloneqq\; \hat{W}_\delta + \hat{\tau}^{\mathrm{te}},
\end{equation}
the bench-adjusted analogue of~\eqref{eq:sigma-HO-K-hat}, again subscripted ``$\mid K$'' to indicate that it is built \emph{knowing} a {$K$-split} CV run has been observed ($K \ge 2$). Unbiasedness follows identically to
{Appendix}~\ref{sec:tau-estimation}:
\begin{equation}
    \mathbb{E}[\hat{\tau}^{\mathrm{te}}]
  = (\sigma^{2,\mathrm{te}} + (K-1)\tau^{\mathrm{te}})/K
    - (\sigma^{2,\mathrm{te}} - \tau^{\mathrm{te}})/K
  = \tau^{\mathrm{te}}.
\end{equation}
In practice, \eqref{eq:tau-te-hat} is computed by applying the raw estimator~\eqref{eq:tau-hat} to the delta column~$\delta_{s,k}$ rather than to the raw fold-score column~$E_{s,k}$. No new estimator machinery is needed.

\bigskip

\paragraph{Variance of $\hat{\tau}^{\mathrm{te}}$.}
The derivation in {Appendix}~\ref{sec:tau-estimation} relied only on compound symmetry within seeds and independence across seeds, both inherited by~\eqref{eq:tau-te-model}. All steps therefore transfer verbatim under the substitutions $(\sigma^2_{\text{HO}}, \tau) \mapsto (\sigma^{2,\mathrm{te}}, \tau^{\mathrm{te}})$, $(V_s, \bar{E}_s) \mapsto (V^\delta_s, \bar{\delta}_s)$. Writing $\gamma_\delta \coloneqq \operatorname{Var}(\bar{\delta}_s)
= (\sigma^{2,\mathrm{te}} + (K-1)\tau^{\mathrm{te}})/K$ and $\mu^\delta_4 \coloneqq \mathbb{E}[(\bar{\delta}_s - \mu_\delta)^4]$, the analogue of~\eqref{eq:VarB} is
\begin{equation}\label{eq:VarBDelta}
  \operatorname{Var}(\hat{B}_\delta)
  = \frac{1}{S}\!\left(
      \mu^\delta_4 - \frac{S-3}{S-1}\,\gamma_\delta^2
    \right),
\end{equation}
and, under joint Gaussianity of $(\delta_{s,1}, \dots, \delta_{s,K})$ with compound-symmetry covariance
$\Sigma_\delta = (\sigma^{2,\mathrm{te}} - \tau^{\mathrm{te}})I_K +
  \tau^{\mathrm{te}}\mathbf{1}\mathbf{1}^\top$,
the quadratic form $(K-1)V^\delta_s$ is
$(\sigma^{2,\mathrm{te}} - \tau^{\mathrm{te}}) \chi^2_{K-1}$, giving
\begin{equation}\label{eq:VarWDelta}
  \operatorname{Var}(\hat{W}_\delta)
  = \frac{2(\sigma^{2,\mathrm{te}} - \tau^{\mathrm{te}})^2}{S(K-1)}.
\end{equation}
Gaussianity together with compound symmetry implies $\bar{\delta}_s \perp V^\delta_s$ within each seed (Fisher--Cochran, exactly as in {Appendix}~\ref{sec:tau-estimation}), so $\operatorname{Cov}(\hat{B}_\delta, \hat{W}_\delta) = 0$, and combining:
\begin{equation}\label{eq:VarTauTe}
  \boxed{\;
  \operatorname{Var}(\hat{\tau}^{\mathrm{te}})
    = \operatorname{Var}(\hat{B}_\delta)
      + \frac{1}{K^2}\operatorname{Var}(\hat{W}_\delta)
    = \frac{2(\sigma^{2,\mathrm{te}} + (K-1)\tau^{\mathrm{te}})^2}{K^2(S-1)}
      + \frac{2(\sigma^{2,\mathrm{te}} - \tau^{\mathrm{te}})^2}{S\,K^2(K-1)}
  \;}.
\end{equation}

\paragraph{Interpretation}
The $K$-scaling parallels the raw case: the $\hat{W}_\delta$ contribution vanishes as $O(K^{-3})$ while $\operatorname{Var}(\hat{B}_\delta) \to 2(\tau^{\mathrm{te}})^2/(S-1)$, so precision is again ultimately seed-limited. Crucially, the asymptotic floor $2(\tau^{\mathrm{te}})^2/(S-1)$ is quadratic in $\tau^{\mathrm{te}}$, which is typically one to three orders of magnitude smaller than $\tau$ due to benchmark cancellation.
The same seed budget therefore yields confidence intervals on $\hat{\tau}^{\mathrm{te}}$ that are orders of magnitude tighter than on $\hat{\tau}$.

\paragraph{Plug-in and bootstrap}
We evaluate~\eqref{eq:VarTauTe} with the sample counterparts $(\hat{\sigma}^{2,\mathrm{te}}_{\mid K}, \hat{\tau}^{\mathrm{te}}, \hat{\mu}^\delta_4)$, and supplement it with a non-parametric bootstrap over seeds: resample
$(\bar{\delta}_s, V^\delta_s)_{s=1}^{S}$ with replacement $100$~times, recompute $\hat{\tau}^{\mathrm{te}}$ per replicate via~\eqref{eq:tau-te-hat}, and report percentile confidence intervals.

\FloatBarrier
\section{Estimating and anticipating sample gains}
\label{sec:sample-gain-diagnostics}

This section gathers the operational tools used to estimate or anticipate test sample gains. We first describe how the reported gains are estimated and how they relate to the inter-split covariance $\tau^{\mathrm{te}}$ of \autoref{sec:inter-fold-covariance-estimation}.{} We then define the redundancy, a study-only statistic for deciding, after only a few splits of a single CV run, whether further repetitions are likely to pay off.

\subsection{Benchmark-aware estimation of sample gain \texorpdfstring{$G^{\text{test}}_K$}{Gtest K}}
\label{sec:sample-gain-estimation}

\paragraph{Estimator}
The reported gains implement the definition of $G^{\text{test}}_K$ in \autoref{sec:sample_gain} directly. For each seed $s = 1, \dots, S$, the $K$-split CV run yields the benchmark-adjusted fold errors $\delta_{s,1}, \dots, \delta_{s,K}$ of~\eqref{eq:delta-def} and their average
\begin{equation}
  \Delta_K^{(s)}
  \coloneqq \frac{1}{K}\sum_{k=1}^{K}\delta_{s,k}.
\end{equation}
{The predictor of the first split is also scored on $n_{\text{ch}}$ chunks of the outer set, each the size $n_{\text{te}}$ of a test fold (up to $200$ of them). Let $\zeta_{s,0} = \delta_{\text{HO}}^{(s)}$ be its benchmark-adjusted error on the test fold, the single-split error, and $\zeta_{s,j}$ its benchmark-adjusted error on chunk $j$, for $1 \le j \le n_{\text{ch}}$. On the test fold and a set $A$ of $\alpha - 1$ chunks together, a test set of size $\alpha\,n_{\text{te}}$, its error is $\delta_{\text{HO}}^{(s)}(A) = \bigl(\zeta_{s,0} + \sum_{j \in A}\zeta_{s,j}\bigr)/\alpha$. The two sides of the definition are estimated by variances across seeds, the single-split one averaged over all $\binom{n_{\text{ch}}}{\alpha-1}$ sets $A$ of $\alpha - 1$ chunks,}
\begin{equation}
  \widehat{V}^\delta_1(\alpha\,n_{\text{te}})
  = {\binom{n_{\text{ch}}}{\alpha-1}^{-1} \sum_{|A| = \alpha - 1}} \widehat{\operatorname{Var}}_s\bigl[\delta_{\text{HO}}^{(s)}{(A)}\bigr],
  \qquad
  \widehat{V}^\delta_K(n_{\text{te}})
  = \widehat{\operatorname{Var}}_s\bigl[\Delta_K^{(s)}\bigr],
  \label{eq:sample-gain-curves}
\end{equation}
{so that the single-split curve does not depend on the order in which the chunks were scored. The average has a closed form: with $\widehat{c}_{jl}$ the covariance across seeds of $\zeta_{s,j}$ and $\zeta_{s,l}$ ($0 \le j, l \le n_{\text{ch}}$), and $\bar{c}_0$, $\bar{c}_{\mathrm{d}}$ and $\bar{c}_{\mathrm{o}}$ the means of $\widehat{c}_{0j}$, of $\widehat{c}_{jj}$ and of $\widehat{c}_{jl}$ ($j \neq l$) over the chunks $j, l \ge 1$,
\begin{equation}
  \widehat{V}^\delta_1(\alpha\,n_{\text{te}})
  = \frac{\widehat{c}_{00} + 2(\alpha-1)\,\bar{c}_0 + (\alpha-1)\,\bar{c}_{\mathrm{d}} + (\alpha-1)(\alpha-2)\,\bar{c}_{\mathrm{o}}}{\alpha^2},
\end{equation}
so every chunk contributes at every multiple $\alpha > 1$.} $\widehat{G}^{\text{test}}_K$ is the test-size multiple at which the first falls to the level of the second. With $\alpha_0$ the smallest whole multiple such that $\widehat{V}^\delta_1(\alpha_0\,n_{\text{te}}) \le \widehat{V}^\delta_K(n_{\text{te}})$, the crossing is read between $\alpha_0 - 1$ and $\alpha_0$, by linear interpolation of $\log \widehat{V}^\delta_1$ against $\log \alpha$. The estimate is exactly $1$ when $\alpha_0 = 1$, that is, when the $K$-split variance does not fall below the single-split one, and it is capped at the largest available multiple when the single-split curve never falls that low. The single-split errors are those of the first split of the same runs, so both variances are computed on the same seeds and $\widehat{G}^{\text{test}}_1 = 1$. {For \texttt{Qwen3-0.6B} on \texttt{Yelp}, they come from separate runs that share the seeds, the study sets and the first split ({Appendix}~\ref{appendix:real-training-loops}), so the two variances use the same seeds but not the same fits. Averaging over the sets of chunks, as the redundancy analyses of {Appendix}~\ref{sec:study-only-redundancy-statistics} also do, removes the noise that a fixed order of the chunks would share across the configurations scored on the same chunks.}

\paragraph{Variance-ratio form}
If the single-split variance decreases as the inverse of the test size, $V^\delta_1(\alpha\,n_{\text{te}}) = V^\delta_1(n_{\text{te}})/\alpha$, the crossing has a closed form, the ratio of the single-split and $K$-split variances at the test-fold size:
\begin{equation}\label{eq:sample-gain-def}
  G^{\text{test}}_K(F_\lambda,n_{\text{tr}})
  \approx
  \frac{V^\delta_1(n_{\text{te}})}{V^\delta_K(n_{\text{te}})}
  =
  \frac{\operatorname{Var}\!\left(\delta_{\text{HO}}^{(s)}\right)}
       {\operatorname{Var}\!\left(\Delta_K^{(s)}\right)}.
\end{equation}
This form links the gain to the inter-split covariance $\tau^{\mathrm{te}}$ of {Appendix}~\ref{sec:tau-te-estimation} and, through it, to the redundancy of {Appendix}~\ref{sec:study-only-redundancy-statistics}. The reported gains are the crossing estimates above. On the simulated runs of \autoref{fig:avg_test_sample_gain_simulated}, where the single-split variance follows {the inverse of the test size} within about $20\%$ up to $\alpha = 20$, the two agree within {$12\%$}: $G^{\text{test}}_{20}$ is $4.1$ against {$4.1$} for \texttt{Ridge}, $4.6$ against {$4.3$} for \texttt{ExtraTrees}, $7.1$ against {$6.8$} for the \texttt{MLP} and $7.6$ against {$8.6$} for \texttt{RandomFeatures}. {They part further when errors are heavy-tailed: for the random-search \texttt{MLP} of \autoref{appendix:hpo}, the single-split variance at $\alpha = 20$ is $0.84$ times {$V^\delta_1(n_{\text{te}})/20$, the value that the inverse of the test size predicts}, and the variance ratio gives $38.4$ at $K = 20$ against $31.5$ for the crossing.}

\paragraph{Decomposition in terms of $\tau^{\mathrm{te}}$ and $\sigma^{2,\mathrm{te}}$.}
By~\eqref{eq:VarDeltaBar}, the CV-estimation-error variance is
\begin{equation}\label{eq:cv-mean-var-identity}
  V^\delta_K(n_{\text{te}})
  = \operatorname{Var}(\Delta_K^{(s)})
  = \operatorname{Var}(\bar{\delta}_s)
  = \frac{\sigma^{2,\mathrm{te}} + (K-1)\,\tau^{\mathrm{te}}}{K},
\end{equation}
which is the population quantity estimated directly by $\hat{B}_\delta$. Under the matched test-set geometry used throughout this paper, the single-split variance equals the marginal $\delta$-score variance introduced in {Appendix}~\ref{sec:tau-te-estimation}:
\begin{equation}
  V^\delta_1(n_{\text{te}})
  = \operatorname{Var}\!\left(\delta_{\text{HO}}^{(s)}\right)
  = \operatorname{Var}(\delta_{s,k})
  = \sigma^{2,\mathrm{te}}.
\end{equation}
Substituting into~\eqref{eq:sample-gain-def} gives
\begin{equation}\label{eq:sample-gain-decomp}
  \boxed{\;
    G^{\text{test}}_K(F_\lambda,n_{\text{tr}})
    \approx
    \frac{K\,\sigma^{2,\mathrm{te}}}
         {\sigma^{2,\mathrm{te}} + (K-1)\,\tau^{\mathrm{te}}}
    = \frac{\sigma^{2,\mathrm{te}}}
           {\operatorname{Var}(\Delta_K^{(s)})}
  \;}.
\end{equation}

The first form expresses the test sample gain using the two ANOVA-aligned estimands of {Appendix}~\ref{sec:tau-te-estimation}. The second form makes clear that the gain is the ratio of the single-HO and CV-estimation-error variances of the benchmark-adjusted $\delta$ score.

\medskip\noindent
\textbf{Remark (idealised gain).}
If the benchmark is geometrically matched to training so that $\tau^{\mathrm{te}} = 0$, \eqref{eq:sample-gain-decomp} recovers the textbook $G^{\text{test}}_K(F_\lambda,n_{\text{tr}}) \approx K$. {An observed $\widehat{G}^{\text{test}}_K(F_\lambda,n_{\text{tr}})$ below $K$ therefore reflects a positive correlation between splits, $\tau^{\mathrm{te}} > 0$. One above $K$ reflects a negative correlation or the noise of the estimator (paragraph \emph{Estimates above $K$})}.

\paragraph{Intraclass-correlation form}
Under the matched test-set geometry of~\eqref{eq:sample-gain-decomp}, we can rewrite the gain approximation in intraclass-correlation (ICC) form:
\begin{equation}\label{eq:sample-gain-icc}
  G^{\text{test}}_K(F_\lambda,n_{\text{tr}})
  \approx
  \frac{K}{1 + (K-1)\,\rho_\delta},
  \qquad
  \rho_\delta \coloneqq \frac{\tau^{\mathrm{te}}}{\sigma^{2,\mathrm{te}}}.
\end{equation}

The ratio $\rho_\delta \le 1${} is the intraclass correlation of the fold $\delta$-scores, positive in our configurations outside the noise-injection grid (paragraph \emph{Estimates above $K$}). It is scale-free, requires no external single-split seeds, and is estimable from the ANOVA decomposition of the $K$-split runs ({Appendix}~\ref{sec:tau-te-estimation}). This gives a compact decision rule:
\begin{equation}\label{eq:decision-rule}
  \rho_\delta \ll \frac{1}{K-1}
    \;\Rightarrow\; G^{\text{test}}_K(F_\lambda,n_{\text{tr}}) \approx K;
  \qquad
  \rho_\delta \to 1
    \;\Rightarrow\; G^{\text{test}}_K(F_\lambda,n_{\text{tr}}) \approx 1
      \text{ (CV is redundant)}.
\end{equation}

\begin{figure}[t]
    \centering
    \includegraphics[width=0.54\linewidth]{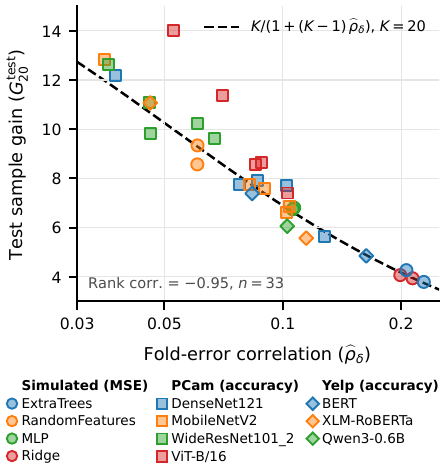}
    \caption{
    Estimated fold-error correlation
    $\widehat\rho_\delta$ versus observed
    $G^{\mathrm{test}}_{20}$ {in $33$ configurations}.
    The dashed curve is the ICC approximation
    $K/(1+(K-1)\widehat\rho_\delta)$.
    }
    \label{fig:rho-delta-vs-gain}
\end{figure}
\noindent
\autoref{fig:rho-delta-vs-gain} illustrates the empirical content of \eqref{eq:decision-rule}. When the estimated fold-error correlation is small, the repeated split evaluations retain complementary information and the observed sample gain can be large. Conversely, as $\widehat\rho_\delta$ increases, folds become more redundant and the
attainable gain decreases. The monotone trend follows the ICC curve closely enough to make $\widehat\rho_\delta$ a useful diagnostic for understanding why a given $K$-split CV run does or does not approach the ideal gain $K$. {The $33$ configurations are the four simulated learners of \autoref{fig:avg_test_sample_gain_simulated} at $n_{\text{tr}} = 1{,}000$ and, as in \autoref{fig:fixed-vs-resampled}, $100$ (mean squared error, $1{,}000$ seeds) and, on accuracy with $50$ seeds, the four \texttt{PCam} networks of \autoref{fig:all_data_test_sample_gains} at $n_{\text{tr}} = 24$, $80$, $240$, $400$ and $640$ and the three \texttt{Yelp} networks at $n_{\text{tr}} = 1{,}000$, \texttt{BERT} and \texttt{XLM-RoBERTa} also at $3{,}000$. The ratio $\widehat\rho_\delta = \hat{\tau}^{\mathrm{te}}/\hat{\sigma}^{2,\mathrm{te}}_{\mid K}$, from \eqref{eq:tau-te-hat} and \eqref{eq:sigma-te-K-hat}, is computed on the first $20$ splits of each seed. The rank correlation is {$-0.95$}, and the observed gains lie within a factor {$0.89$ to $1.40$} of the curve.}

\paragraph{Estimates above $K$}
When $\rho_\delta \ge 0$, the variance-ratio form gives $G^{\text{test}}_K \le K$: $K$ splits are worth at most $K$ independent hold-outs. {Gains above $K$ arise in two ways. In the noise-injection grid of {Appendix}~\ref{appendix:redundancy-runs}, the injected noise brings the redundancy close to zero and the coupling term of {Appendix}~\ref{sec:study-only-redundancy-statistics}, intrinsic to cross-validation, makes the fold errors slightly anticorrelated: in the $38$ configurations of \autoref{fig:redundancy-vs-gain} ($37$ of them in this grid) whose gain exceeds $K = 20$ even with the single-split variance estimated on the outer chunks, which is far less noisy, the implied $\rho_\delta$ lies between $-0.025$ and $0$ (median $-0.011$), and these gains are genuine. Elsewhere, the estimated $\widehat\rho_\delta$ is positive in every configuration we examined ($0.004$ to $0.32$, over \autoref{fig:rho-delta-vs-gain}, \autoref{tab:pcam-redundancy}, \autoref{tab:hpo_gains} and the pre-registered holdout of \autoref{fig:redundancy-vs-gain}), yet} a few reported gains exceed $K$, such as {$31.5$} for the random-search \texttt{MLP} of \autoref{tab:hpo_gains} and $35.1$ for \texttt{ViT-B/16} under the negative log-likelihood in \autoref{tab:pcam-redundancy}, both at $K = 20$. They reflect the noise of the single-split side of the estimator, which rests on the first split of each seed: its variance at the test-fold size comes from that split's test fold, and its decrease with the test size from outer chunks scored by the same predictor. With heavy-tailed losses, both are fragile. For the random-search \texttt{MLP}, $7$ of its $51$ worst fits fall on the first split, against $2.5$ expected, and inflate its variance. For \texttt{ViT-B/16}, the errors on the outer chunks vary about four times more across seeds than those on the test folds, so the single-split curve falls more slowly than {the inverse of the test size} and pushes the crossing up. Each of the $K$ splits is a single-split evaluation in its own right, so averaging the single-split variance over them gives a check that is much less sensitive to such fits: it is essentially the intraclass-correlation value $K/\bigl(1+(K-1)\,\widehat\rho_\delta\bigr)$, it brings these two gains to $16.1$ and $10.2$, and it changes the corresponding ratio for the main simulated learners by less than a quarter. The reported estimate keeps the first split, so that both sides are computed on the same splits and $\widehat{G}^{\text{test}}_1 = 1$. An estimate above $K$ should be read together with its interval (\autoref{appendix:gain-intervals}), which is then wide.

\medskip

\paragraph{Plug-in estimate of the variance-ratio form.}
Replacing each quantity in~\eqref{eq:sample-gain-decomp} by its sample counterpart gives a plug-in estimate of the variance-ratio form, which expresses the gain through the ANOVA estimates of {Appendix}~\ref{sec:tau-te-estimation} (the reported gains are the crossing estimates above):
\begin{equation}\label{eq:sample-gain-plugin}
  \boxed{\;
  \begin{split}
    \widehat{G}^{\text{test}}_K(F_\lambda,n_{\text{tr}})
    &= \frac{\widehat{V}^\delta_1(n_{\text{te}})}
           {\widehat{V}^\delta_K(n_{\text{te}})}
    = \frac{\hat{\sigma}^{2,\mathrm{te}}_{\mid 1}}{\hat{B}_\delta}
    \\
    &= \frac{K\,\hat{\sigma}^{2,\mathrm{te}}_{\mid 1}}
           {\hat{\sigma}^{2,\mathrm{te}}_{\mid K}
             + (K-1)\,\hat{\tau}^{\mathrm{te}}}
\end{split}
  \;}.
\end{equation}
Here $\widehat{V}^\delta_K(n_{\text{te}})=\hat{B}_\delta$ estimates $V^\delta_K(n_{\text{te}})$, while
$\widehat{V}^\delta_1(n_{\text{te}})=\hat{\sigma}^{2,\mathrm{te}}_{\mid 1}$ is the bench-adjusted across-seed sample variance of the single-split estimation errors $(\delta_{\text{HO}}^{(s)})_{s=1}^{S_{\text{HO}}}$:
\begin{equation}\label{eq:sigma-te-1-hat}
  \hat{\sigma}^{2,\mathrm{te}}_{\mid 1}
  \;\coloneqq\;
  \frac{1}{S_{\text{HO}} - 1}
    \sum_{s=1}^{S_{\text{HO}}}
      \bigl(
        \delta_{\text{HO}}^{(s)}
        - \bar{\delta}_{\text{HO}}
      \bigr)^2.
\end{equation}
The subscript ``$\mid 1$'' signals that this estimator is built \emph{knowing} a set of single-split runs has been observed (the degenerate $K=1$ case), and mirrors the ``$\mid K$'' convention introduced for $\hat{\sigma}^{2,\mathrm{te}}_{\mid K}$. The last equality in~\eqref{eq:sample-gain-plugin} follows from $\hat{\sigma}^{2,\mathrm{te}}_{\mid K} + (K-1)\hat{\tau}^{\mathrm{te}} = K\hat{B}_\delta$ (by~\eqref{eq:sigma-te-K-hat} and~\eqref{eq:tau-te-hat}).

\paragraph{Uncertainty}
The uncertainty of the reported $\widehat{G}^{\text{test}}_K$ is assessed by resampling the seeds and recomputing both variance curves and their crossing. \autoref{appendix:gain-intervals} gives the procedure, its calibration and the intervals of every gain figure.

\paragraph{Interpretation}
As $K$ grows, both the variance-ratio form \eqref{eq:sample-gain-decomp} and its intraclass-correlation form \eqref{eq:sample-gain-icc} approach the asymptotic bound\begin{equation}\label{eq:sample-gain-ceiling}
  G^{\text{test}}_K(F_\lambda,n_{\text{tr}})
  \;\xrightarrow[K\to\infty]{\text{local ratio}}\;
    \frac{\sigma^{2,\mathrm{te}}}{\tau^{\mathrm{te}}},
\end{equation}
so the irreducible ceiling on the test sample gain is dictated entirely by the ratio of the single-split variance to the residual fold covariance $\tau^{\mathrm{te}}$.

\subsection{Study-only redundancy for early-stopping cross-validation}
\label{sec:study-only-redundancy-statistics}

\paragraph{Operational goal}
The exact test sample gain in \eqref{eq:sample_gain} requires an outer set and a large benchmarking set. Practitioners usually do not have these quantities when deciding whether to spend more compute on additional splits. The goal of this section is therefore different: after only a few splits of a single CV run, can one decide whether continuing CV is likely to pay off?

The statistic below is computed from the study set only. It uses neither the outer set nor the benchmarking set, and it does not average over several experimental seeds. In practice, after observing the first $k$ splits of one CV run, with $k=3$ or $k=5$, one computes the statistic on these already-trained predictors. It should be interpreted as a triage rule rather than an unbiased estimator of $G^{\text{test}}_K(F_\lambda,n_{\text{tr}})$: high redundancy suggests stopping early, whereas low redundancy indicates that larger gains remain plausible.

\paragraph{The redundancy}
Consider one partial CV run, indexed by $s$, after $k$ splits. For split $a\in\{1,\dots,k\}$, keep the main-paper notation
\[
  g_{a,s} = F_\lambda(\mathcal{D}^{\mathrm{tr}}_{\eta_{a,s}},\xi_{a,s}),
  \qquad
  \mathcal{D}^{\mathrm{te}}_{a,s} = \mathcal{D}^{\mathrm{te}}_{\eta_{a,s}} .
\]
For two splits $a<b$, let
\[
  I_{ab,s}=\mathcal{D}^{\mathrm{te}}_{a,s}\cap\mathcal{D}^{\mathrm{te}}_{b,s}
\]
be the study observations held out in both splits. For $i\in I_{ab,s}$, define the out-of-fold loss
\[
  e_{a,s,i}=\ell(g_{a,s}(x_i),y_i),
\]
computed with the loss of the metric whose gain is targeted: the squared error for gains on the mean squared error, the $0$-$1$ error for accuracy, the negative log-likelihood for the log-loss. We compute the pair-averaged loss covariance and loss variance
\begin{align}
  \widehat C^{\mathrm{study}}_{e,k,s}
  &=
  {k \choose 2}^{-1}
  \sum_{a<b}
  \widehat{\operatorname{Cov}}_{i\in I_{ab,s}}(e_{a,s,i},e_{b,s,i}),
  \\
  \widehat V^{\mathrm{study}}_{e,k,s}
  &=
  {k \choose 2}^{-1}
  \sum_{a<b}
  \frac{1}{2}\left(
  \widehat{\operatorname{Var}}_{i\in I_{ab,s}}(e_{a,s,i})
  +
  \widehat{\operatorname{Var}}_{i\in I_{ab,s}}(e_{b,s,i})
  \right),
\end{align}
and their ratio, the loss correlation
\begin{equation}
  \widehat\rho^{\mathrm{study}}_{e,k,s}
  =
  \frac{\widehat C^{\mathrm{study}}_{e,k,s}}
       {\widehat V^{\mathrm{study}}_{e,k,s}} .
  \label{eq:study-loss-correlation}
\end{equation}
It is large when different split predictors tend to be wrong on the same held-out observations. Pairs that share fewer than two observations are skipped, and the statistic is deferred until a pair qualifies. With $t = {n_{\text{te}}/(n_{\text{tr}}+n_{\text{te}})}$ the test fraction, the redundancy is
\begin{equation}
  \boxed{
  \widehat\omega^{\mathrm{study}}_{k,s}(F_\lambda,n_{\text{tr}})
  =
  t\,\widehat\rho^{\mathrm{study}}_{e,k,s}
  }.
  \label{eq:study-redundancy-score}
\end{equation}
It is dimensionless and lies between $-t$ and $t$, so it can be compared across tasks, targets and study sizes.

\paragraph{Why the test fraction: relation to the sample gain}
Write $d_{a,i} = \ell(g_a(x_i),y_i) - \mathcal{R}^*(g_a)$ for the loss of an observation held out by split $a$, centred at the risk of the predictor that did not see it, and $\sigma^2_e = \mathbb{E}[d_{a,i}^2]$. Under the evaluation-noise approximation \eqref{eq:delta_variance_approx}, $\delta_a = n_{\text{te}}^{-1}\sum_{i\in\mathcal{D}^{\mathrm{te}}_a} d_{a,i}$, so that $\sigma^{2,\mathrm{te}} = \sigma^2_e/n_{\text{te}}$. The covariance of two splits' errors separates the observations that both splits test from the pairs of distinct observations:
\begin{equation}
  \tau^{\mathrm{te}}
  = \frac{1}{n_{\text{te}}^2}
    \Bigl(
      \mathbb{E}\Bigl[\sum_{i \in I_{ab}} d_{a,i}\, d_{b,i}\Bigr]
      +
      \mathbb{E}\Bigl[\sum_{i \in \mathcal{D}^{\mathrm{te}}_a}\;
                  \sum_{j \in \mathcal{D}^{\mathrm{te}}_b,\, j \neq i} d_{a,i}\, d_{b,j}\Bigr]
    \Bigr).
  \label{eq:tau-te-split}
\end{equation}
Two MCCV test sets share on average $\mathbb{E}|I_{ab}| = n_{\text{te}}^2/{(n_{\text{tr}}+n_{\text{te}})} = t\,n_{\text{te}}$ observations, on each of which the two losses covary by $\rho_e\,\sigma^2_e$, where $\rho_e$ is the population counterpart of \eqref{eq:study-loss-correlation}. The second term is a coupling between splits: observation $i$ may belong to the training set of $g_b$ and $j$ to that of $g_a$, and the two predictors share most of their training data. Writing {$\psi$} for the average covariance of such a pair, dividing by $\sigma^{2,\mathrm{te}}$ gives
\begin{equation}
  \rho_\delta = \frac{\tau^{\mathrm{te}}}{\sigma^{2,\mathrm{te}}} = t\,\rho_e + \kappa,
  \qquad
  \kappa = \frac{n_{\text{te}}\,\psi}{\sigma^2_e},
  \label{eq:rho-delta-decomp}
\end{equation}
and, with the intraclass-correlation form \eqref{eq:sample-gain-icc},
\begin{equation}
  G^{\text{test}}_K \approx \frac{K}{1 + (K-1)\,(\omega + \kappa)},
  \qquad
  \omega = t\,\rho_e .
  \label{eq:gain-redundancy}
\end{equation}
The redundancy $\omega$ is thus the part of the fold-error correlation carried by the observations that two splits both test. {Here, $t$ is the validation fraction of \citet{nagler_reshuffling_2024} {(their $\alpha$)}, and $\rho_\delta \to t$ is the large-sample limit that their Theorem~2.1 gives for MCCV when the split predictors converge to a common rule ({Appendix}~\ref{appendix:cv-reshuffling-asymptotics})}. With $\kappa=0$, \eqref{eq:gain-redundancy} is a reference curve: $G^{\text{test}}_K \to K$ when split predictors err independently ($\omega \to 0$), and when they err alike ($\rho_e \to 1$) the gain is capped, as $K\to\infty$, at $1/t = {(n_{\text{tr}}+n_{\text{te}})/n_{\text{te}}}$, cross-validation then averaging over the whole study set and no more. The coupling $\kappa$ is intrinsic to cross-validation, can take either sign, and is not estimable from the study set alone. This is one reason the redundancy is used as a triage signal rather than plugged into \eqref{eq:gain-redundancy} as an estimator of the gain. The overlap factor $t$ is specific to MCCV: in repeated $K$-fold, splits of the same repeat have disjoint test sets.

The two ingredients of $\omega$ can be checked directly with the benchmarking set. Across our simulated runs, replacing the study-only loss correlation \eqref{eq:study-loss-correlation} by the loss correlation of the same split predictors measured on the benchmarking set changes the gain given by the reference curve by less than $2\%$. Varying the test fraction $t\in\{0.1,0.2,0.3,0.5\}$ with everything else fixed (simulated linear regression with \texttt{Ridge}, and binary labels drawn from a logistic model with logistic regression; per-sample noise added to the predictions, on the logit scale for classification, drawn afresh at each fit, to vary $\rho_e$; training sizes $300$ and $1{,}000$, $K=20$, $50$ seeds), the factor $\rho_\delta/\rho_e$ fitted across configurations is $0.14$, $0.20$, $0.36$, $0.46$ in regression and $0.11$, $0.21$, $0.33$, $0.49$ in classification, with intercepts within $\pm 0.01$: it follows $t$.

\paragraph{Single-run validation on a pre-registered holdout}
We validate the early-stopping use of the redundancy on a simulated run whose design was fixed and pre-registered before launch: $3$ synthetic data families ({Appendix}~\ref{appendix:simulated-process}, {Appendix}~\ref{appendix:redundancy-runs}) with $d=20$ features, $3$ noise levels ($\sigma\in\{0.3,1,2.5\}$), $3$ training sizes ($n_{\text{tr}}\in\{75, 750, 5{,}000\}$), and $9$ learning algorithms{: \texttt{Ridge}, \texttt{ExtraTrees} and the \texttt{MLP} of the main runs (\autoref{sec:experimental-evaluation}), \texttt{DecisionTree} (depth $100$), \texttt{HistGradientBoosting} (depth $20$), \texttt{KNeighbors} ($20$ neighbours), \texttt{SVR}, \texttt{GradientBoosting} and a random forest ($200$ trees each)}, with $50$ seeds each: $243$ configurations and $12{,}150$ single partial CV runs. Each run is continued to $K=200$ splits, but the redundancy used for triage is computed only from the first $k$ splits of one seed, on the squared error, and compared to the exact configuration-level gains $G^{\text{test}}_{20}$ and $G^{\text{test}}_{200}$ on the mean squared error, computed only for analysis and not available to the early-stopping rule.

After three splits, the redundancy correlates with $\log G^{\text{test}}_{200}$ at $-0.65$ (rank correlation $-0.36$) across single runs{, and with $\log G^{\text{test}}_{20}$ at $-0.58$ ($-0.34$)}. After five splits these become $-0.73$ and $-0.38$ {($-0.65$ and $-0.36$ for $G^{\text{test}}_{20}$). The rank correlations are modest because $70\%$ of the holdout's configurations have $0.15 \le \widehat\omega \le 0.20$, where the reference curve \eqref{eq:gain-redundancy} is flat and differences in gain are small relative to their uncertainty.}

\emph{Fresh-seed confirmation.} We also repeated, with $50$ seeds never used before and an analysis plan fixed before any gain was computed, the design of an earlier simulated {run}: the $3$ families with $d=20$, $\sigma\in\{0.1,0.7,1.5\}$, $n_{\text{tr}}\in\{50,300,2{,}000\}$, and {the algorithms of the pre-registered holdout except \texttt{GradientBoosting}} ($216$ configurations, $10{,}800$ single runs). The results carry over {(lower block of \autoref{tab:study-redundancy-early-stop})}: after three splits, the redundancy correlates with $\log G^{\text{test}}_{200}$ at $-0.62$ ($-0.73$ after five{, with rank correlations of $-0.42$ and $-0.45$}).

\begin{table}[htbp]
    \centering
    \caption{Single-run early-stopping validation on the pre-registered simulated holdout {and on the fresh-seed confirmation: what stopping on a high redundancy forgoes}. Redundancy tertiles are computed from $\widehat\omega^{\mathrm{study}}_{k,s}$ using only the first $k$ splits of one seed{, over the runs of each design}. Gains are {exact configuration-level gains} on the mean squared error{, large when $G^{\text{test}}_{20}\ge7$ or $G^{\text{test}}_{200}\ge8$, thresholds set after the pre-registered analysis so that about one run in eight and one in seven reach them ($12.8\%$ and $15.2\%$ of the holdout's runs, $13.0\%$ of the fresh seeds', whose saved counts do not resolve $G^{\text{test}}_{200}$ below $20$). The results at the pre-registered thresholds, $G^{\text{test}}_{20}\ge10$ and $G^{\text{test}}_{200}\ge20$, are given in the text. Runs: runs with a large gain, in all and in the high tertile. Rate: share of a tertile's runs with a large gain. Share: share of the large-gain runs in a tertile. $^{\dagger}$Bounds set by the saved per-tertile counts, the per-run fresh-seed results being no longer available.}}
    \label{tab:study-redundancy-early-stop}
    \scriptsize
    \setlength{\tabcolsep}{3.0pt}
    \begin{tabular}{@{}ccccccccccc@{}}
        \toprule
        & & & \multicolumn{3}{c}{{\thead{Rate (\%)}}} & \multicolumn{3}{c}{{\thead{Share (\%)}}} & \multicolumn{2}{c}{{\thead{High tertile}}} \\
        \cmidrule(lr){4-6} \cmidrule(lr){7-9} \cmidrule(l){10-11}
        \thead{$k$} & {\thead{Large gain}} & {\thead{Runs}} & {\thead{Low}} & {\thead{Mid}} & {\thead{High}} & {\thead{Low}} & {\thead{Mid}} & {\thead{High}} & {\thead{Largest}} & {\thead{Runs}} \\
        \midrule
        \multicolumn{11}{@{}l}{{\emph{Pre-registered holdout: $243$ configurations, $12{,}150$ runs, tertiles of $4{,}050$}}} \\
        {3} & {$G^{\text{test}}_{20}\ge7$} & {1{,}550} & {33.93} & {2.02} & {2.32} & {88.6} & {5.3} & {6.1} & {7.9} & {94} \\
        {3} & {$G^{\text{test}}_{200}\ge8$} & {1{,}850} & {37.31} & {1.98} & {6.40} & {81.7} & {4.3} & {14.0} & {9.2} & {259} \\
        \addlinespace[2pt]
        {5} & {$G^{\text{test}}_{20}\ge7$} & {1{,}550} & {34.49} & {1.80} & {1.98} & {90.1} & {4.7} & {5.2} & {7.9} & {80} \\
        {5} & {$G^{\text{test}}_{200}\ge8$} & {1{,}850} & {37.98} & {1.46} & {6.25} & {83.1} & {3.2} & {13.7} & {9.2} & {253} \\
        \midrule
        \multicolumn{11}{@{}l}{{\emph{Fresh-seed confirmation: $216$ configurations, $10{,}786$ runs ($10{,}800$ at $k=5$), tertiles of $3{,}595$ to $3{,}600$}}} \\
        {3} & {$G^{\text{test}}_{20}\ge7$} & {1{,}399} & {36.71} & {0.67} & {1.53} & {94.4} & {1.7} & {3.9} & {$10.4$--$12.2^{\dagger}$} & {55} \\
        \addlinespace[2pt]
        {5} & {$G^{\text{test}}_{20}\ge7$} & {1{,}400} & {37.33} & {0.14} & {1.42} & {96.0} & {0.4} & {3.6} & {$10.4$--$12.2^{\dagger}$} & {51} \\
        \bottomrule
    \end{tabular}
\end{table}

This table should be read asymmetrically. A low redundancy does not guarantee a {large} gain{: after three splits, $33.9\%$ of the low-tertile runs of the holdout reach $G^{\text{test}}_{20} \ge 7$}. It only indicates that continuing is plausible. A high redundancy is more decisive{. After three or five splits, no high-tertile run of the holdout reaches the pre-registered thresholds, $G^{\text{test}}_{20} \ge 10$ and $G^{\text{test}}_{200} \ge 20$, and after three the lowest tertile holds $99.5\%$ and $100\%$ of those that do. On the fresh seeds, a stop on a high reading forgoes $3$ of the $599$ runs with $G^{\text{test}}_{20} \ge 10$ ($1$ of $600$ after five) and none of those with $G^{\text{test}}_{200} \ge 20$. At the post hoc thresholds of the table, the highest tertile holds $6.1\%$ ($G^{\text{test}}_{20} \ge 7$) and $14.0\%$ ($G^{\text{test}}_{200} \ge 8$) of the holdout's runs that reach them after three splits, and $3.9\%$ on the fresh seeds, against $88.6\%$, $81.7\%$ and $94.4\%$ for the lowest. On the holdout, these come from $3$ and $6$ configurations of \texttt{Ridge}, random forests, \texttt{ExtraTrees} and \texttt{KNeighbors}, with high readings (median $0.19$ to $0.20$) and gains of at most $7.9$ and $9.2$.} The separation lies between the lowest tertile and the other two: middle- and high-redundancy runs are {both} unlikely to reach large gains. {These shares rest on few configurations: the runs that reach the post hoc thresholds come from $31$ and $37$ configurations of five and seven algorithms on the holdout and from $28$ of three on the fresh seeds, mostly \texttt{DecisionTree}, and those that reach the pre-registered ones from $11$ \texttt{DecisionTree} configurations each on the holdout, partly different, and, for $G^{\text{test}}_{20} \ge 10$, from $12$ on the fresh seeds, ten \texttt{DecisionTree} and two \texttt{MLP}.}

For practitioners, this gives a simple early-stopping rule. Compute \eqref{eq:study-redundancy-score} after three splits, or after five for a less noisy reading (on this holdout the {median} $10$--$90\%$ range of single-run readings {within a configuration} narrows from $0.020$ to $0.013$). {After two splits, from a single pair, a single reading is noisier (median range $0.035$ on this holdout) and less informative: its correlation with $\log G^{\text{test}}_{20}$ is $-0.45$, against $-0.58$ after three. Averaged over seeds, it still orders the configurations almost as well on the simulated data (rank correlation with $G^{\text{test}}_{20}$ of $-0.78$ on the $663$ configurations of \autoref{fig:redundancy-vs-gain}, against $-0.81$ after three), less well on \texttt{PCam} ($-0.79$ to $-0.92$, against $-0.92$ to $-0.95$). It is undefined when the two test sets share fewer than two observations, in $17\%$ of the runs at $n_{\text{tr}} = 50$ and $6\%$ at $75$.} If redundancy is high, further CV repetitions are unlikely to pay off, even by $K=20$. If redundancy is low, continuing to $K=20$ is justified, and only those low-redundancy cases plausibly merit much heavier runs such as $K=200$.

\paragraph{Relation with configuration-level calibration}
The same relationship appears when the redundancy is averaged over seeds and compared at the configuration level. Pooling all simulated designs (\autoref{fig:redundancy-vs-gain}, $663$ configurations, {Appendix}~\ref{appendix:redundancy-runs}), the rank correlation between the redundancy and $G^{\text{test}}_{20}$ is $-0.81$ after three splits ($-0.82$ after five), its correlation with $\log G^{\text{test}}_{20}$ is $-0.83$, and configurations scatter around the reference curve \eqref{eq:gain-redundancy} with a median log-ratio of $-0.04$: high redundancy corresponds to low sample gain, and the upper tail of $G^{\text{test}}_K$ disappears as redundancy increases. The level of the curve depends on the training size: on {the pre-registered holdout and the fresh-seed confirmation}, gains fall below it at small training sizes (median log-ratio about $-0.25$ at $n_{\text{tr}}\le 75$) and come close to it at the largest ones ($-0.12$ at $2{,}000$, $+0.05$ at $5{,}000$), while \texttt{Ridge}, whose redundancy is close to its maximum $t$ in every configuration, scatters around it with no trend in $n_{\text{tr}}$ (median log-ratio $+0.05$ on the holdout and $-0.05$ on the fresh seeds). This is consistent with a train/test coupling $\kappa>0$ that weighs most when each training observation matters most. On the pre-registered holdout alone, the rank correlation is modest ($-0.34$){, as for its single runs, because most of its configurations lie where the curve is flat}.

\paragraph{Validation on real data}
The relationship carries over to \texttt{PCam}. We computed the redundancy on the four \texttt{PCam} networks (\texttt{DenseNet121}, \texttt{MobileNetV2}, \texttt{ViT-B/16}, \texttt{WideResNet101\_2}) at $n_{\text{tr}} = 100$, $300$ and $800$ (study sets of $125$, $375$ and $1{,}000$ observations, \autoref{tab:xp_splitting}), with $50$ further seeds, $K=20$ splits and test fraction $t=0.2$, separately for three metrics, each with its own per-sample loss: accuracy (0-1 loss), the negative log-likelihood, and the Brier score (squared error of the predicted probability). Gains are the variance-equivalent gains used for the simulated configurations, with the oracle {risks estimated} on a fixed subset of $60{,}000$ benchmarking images. The redundancy is defined in every run after three splits. Across the $12$ configurations of each metric, it orders the gains as on the simulated data (\autoref{tab:pcam-redundancy} and \autoref{fig:redundancy-vs-gain}): rank correlations with $G^{\text{test}}_{20}$ lie between {$-0.92$} and $-0.95$ {($-0.90$ to $-0.93$ after five splits)}. The curve \eqref{eq:gain-redundancy} with $\kappa=0$ predicts the gain with a median ratio of $1.04$ to the measured value. Gains on \texttt{PCam} are high ($G^{\text{test}}_{20}$ from $5.2$ to $35.1$) because the redundancy is low ($0.017$ to $0.107$ after five splits). The highest value, for \texttt{ViT-B/16} under the negative log-likelihood at $n_{\text{tr}} = 300$, exceeds $K$ through the noise of its single-split side ({Appendix}~\ref{sec:sample-gain-estimation}): it is $10.2$ with the single-split variance averaged over the splits, and its {$95\%$ bootstrap interval runs from $8.3$ to $96.9$}. {The \texttt{Yelp} runs of \texttt{BERT} and \texttt{XLM-RoBERTa} at $n_{\text{tr}} = 1{,}000$ and $3{,}000$ ($50$ seeds, $K=20$, {Appendix}~\ref{appendix:real-training-loops}) give the $4$ \texttt{Yelp} points of \autoref{fig:redundancy-vs-gain}, computed as for \texttt{PCam} with the gain on accuracy and the redundancy on the 0-1 loss: redundancies of $0.078$ to $0.126$ after three splits and gains of $4.9$ to $11.1$, in the reverse order (rank correlation $-1.00$ on $4$ points and $-0.90$ pooled with the $36$ \texttt{PCam} configurations), and a median ratio of $1.05$ ($0.72$ to $1.21$) to the curve after five splits.}

\begin{table}[htbp]
    \centering
    \caption{Redundancy against sample gain on \texttt{PCam} ($4$ networks $\times$ {$3$ training sizes, $n_{\text{tr}} = 100$, $300$, $800$}; $50$ seeds; $K=20$). The redundancy is computed after $k$ splits on the per-sample loss of the metric whose gain is targeted. Ratio: gain predicted by \eqref{eq:gain-redundancy} with $\kappa=0$ from the redundancy after five splits, divided by the measured $G^{\text{test}}_{20}$. The highest gain, $35.1$ for \texttt{ViT-B/16} under the negative log-likelihood at $n_{\text{tr}} = 300$, exceeds $20$ through the noise of the single-split side of the estimator: it is $10.2$ with the single-split variance averaged over the splits ({Appendix}~\ref{sec:sample-gain-estimation}).}
    \label{tab:pcam-redundancy}
    \small
    \begin{tabular}{@{} l c c c c c @{}}
        \toprule
        \thead{Metric} & \thead{Rank corr.\\$k=3$} & \thead{Rank corr.\\$k=5$} & \thead{$G^{\text{test}}_{20}$\\range} & \thead{Redundancy\\$k=5$} & \thead{Ratio\\median (range)} \\
        \midrule
        Accuracy & $-0.92$ & $-0.92$ & $5.5$--$16.1$ & $0.028$--$0.096$ & $1.13$ ($0.75$--$1.29$) \\
        Negative log-likelihood & $-0.95$ & $-0.93$ & $6.4$--$35.1$ & $0.017$--$0.104$ & $0.96$ ($0.38$--$1.17$) \\
        Brier score & $-0.93$ & $-0.90$ & $5.2$--$18.4$ & $0.024$--$0.107$ & $1.06$ ($0.72$--$1.27$) \\
        \midrule
        Pooled ($36$) & $-0.90$ & $-0.87$ & $5.2$--$35.1$ & $0.017$--$0.107$ & $1.04$ ($0.38$--$1.29$) \\
        \bottomrule
    \end{tabular}
\end{table}

\FloatBarrier
\section{Confidence intervals of the sample gains}
\label{appendix:gain-intervals}
The gain figures draw, for each curve, the sample gain estimated on all seeds (thick line) and a few bootstrap resamples {drawn at random} (thin lines){, with the vertical axis cut $8\%$ above the highest thick line, so that some thin lines leave the frame}. This appendix gives the values behind them, with their $95\%$ intervals (\autoref{tab:ci-fig4} to \autoref{tab:ci-figM1}{, with $^{\dagger}$ explained under Limits below}).

\begin{table}[htbp]
    \centering
    \caption{Sample gains $G^{\text{test}}_K$ of \autoref{fig:avg_test_sample_gain_simulated} and \autoref{fig:all_data_test_sample_gains} (left): simulated linear process, $n_{\text{tr}} = 1{,}000$, $1{,}000$ seeds, mean squared error.}
    \label{tab:ci-fig4}
    \small
    \begin{tabular}{@{} lcccc @{}}
        \toprule
        \thead{Algorithm} & $K=5$ & $K=10$ & $K=20$ & $K=200$ \\
        \midrule
        \texttt{Ridge} & {$2.7$ {\scriptsize $[2.4,\,3.1]$}} & {$3.7$ {\scriptsize $[3.2,\,4.3]$}} & {$4.1$ {\scriptsize $[3.5,\,4.7]$}} & {$4.8$ {\scriptsize $[4.2,\,5.5]$}} \\
        \texttt{ExtraTrees} & {$3.0$ {\scriptsize $[2.4,\,3.7]$}} & {$3.8$ {\scriptsize $[3.2,\,4.5]$}} & {$4.3$ {\scriptsize $[3.6,\,4.9]$}} & {$4.9$ {\scriptsize $[4.3,\,5.6]$}} \\
        \texttt{MLP} & {$3.3$ {\scriptsize $[2.9,\,3.9]$}} & {$5.0$ {\scriptsize $[4.4,\,5.8]$}} & {$6.8$ {\scriptsize $[6.0,\,7.7]$}} & {$9.9$ {\scriptsize $[8.8,\,11.2]$}} \\
        \texttt{RandomFeatures} & {$3.6$ {\scriptsize $[2.9,\,4.5]$}} & {$6.6$ {\scriptsize $[5.5,\,7.8]$}} & {$8.6$ {\scriptsize $[7.3,\,10.2]$}} & {$14.8$ {\scriptsize $[12.5,\,17.5]$}} \\
        \bottomrule
    \end{tabular}
\end{table}

\begin{table}[htbp]
    \centering
    \caption{Sample gains of \autoref{fig:all_data_test_sample_gains} (middle and right): accuracy{; \texttt{PCam}: $50$ seeds; \texttt{Yelp}: $50$ seeds; the single split is the first of the $20$ splits of each seed for \texttt{BERT} and \texttt{XLM-RoBERTa}, a separate run with the same seed for \texttt{Qwen3-0.6B}}.}
    \label{tab:ci-fig5}
    \small
    \begin{tabular}{@{} llccc @{}}
        \toprule
        \thead{Dataset} & \thead{Network} & $K=5$ & $K=10$ & $K=20$ \\
        \midrule
        {\texttt{PCam}, $n_{\text{tr}}=400$} & \texttt{DenseNet121} & {$4.7$ {\scriptsize $[3.4,\,7.6]$}} & {$5.4$ {\scriptsize $[4.1,\,7.9]$}} & {$7.8$ {\scriptsize $[5.7,\,11.9]$}} \\
         & \texttt{MobileNetV2} & {$4.6$ {\scriptsize $[3.4,\,7.3]$}} & {$6.2$ {\scriptsize $[4.3,\,10.7]$}} & {$7.6$ {\scriptsize $[5.4,\,12.2]$}} \\
         & \texttt{WideResNet101\_2} & {$5.3$ {\scriptsize $[3.6,\,9.4]$}} & {$6.5$ {\scriptsize $[4.9,\,9.4]$}} & {$11.1$ {\scriptsize $[7.9,\,16.7]$}} \\
         & \texttt{ViT-B/16} & {$5.9$ {\scriptsize $[4.3,\,8.5]$}} & {$7.2$ {\scriptsize $[4.8,\,11.7]$}} & {$8.6$ {\scriptsize $[5.7,\,13.9]$}} \\
        \midrule
        \texttt{Yelp}, $n_{\text{tr}}=1{,}000$ & \texttt{BERT} & {$3.7$ {\scriptsize $[2.7,\,5.5]$}} & {$6.1$ {\scriptsize $[4.2,\,9.1]$}} & {$7.4$ {\scriptsize $[5.2,\,11.1]$}} \\
         & \texttt{XLM-RoBERTa} & {$3.6$ {\scriptsize $[2.6,\,5.6]$}} & {$8.6$ {\scriptsize $[6.1,\,14.0]$}} & {$11.1$ {\scriptsize $[7.9,\,16.7]$}} \\
         & \texttt{Qwen3-0.6B} & {$2.9$ {\scriptsize $[1.5,\,5.0]$}} & {$5.6$ {\scriptsize $[3.9,\,9.0]$}} & {$6.1$ {\scriptsize $[4.3,\,9.8]$}} \\
        \bottomrule
    \end{tabular}
\end{table}

\begin{table}[htbp]
    \centering
    \caption{Sample gains of \autoref{fig:hpo-gains}: simulated linear process, $n_{\text{tr}}=1{,}000$, $1{,}000$ seeds, mean squared error.}
    \label{tab:ci-figI1}
    \small
    \begin{tabular}{@{} llccc @{}}
        \toprule
        \thead{Algorithm} & \thead{Tuning} & $K=5$ & $K=10$ & $K=20$ \\
        \midrule
        \texttt{MLP} & Fixed & {$3.3$ {\scriptsize $[2.9,\,3.9]$}} & {$5.0$ {\scriptsize $[4.4,\,5.8]$}} & {$6.8$ {\scriptsize $[6.0,\,7.7]$}} \\
         & Grid search & {$4.6$ {\scriptsize $[3.9,\,5.4]$}} & {$7.9$ {\scriptsize $[6.8,\,9.3]$}} & {$14.2$ {\scriptsize $[12.3,\,16.7]$}} \\
         & Random search & {$4.7$ {\scriptsize $[1.2,\,14.1]$}} & {$13.0$ {\scriptsize $[3.8,\,31.2]$}} & {$31.5$ {\scriptsize $[10.4,\,61.5]$}} \\
        \texttt{ExtraTrees} & Fixed & {$3.0$ {\scriptsize $[2.4,\,3.7]$}} & {$3.8$ {\scriptsize $[3.2,\,4.5]$}} & {$4.3$ {\scriptsize $[3.6,\,4.9]$}} \\
         & Grid search & {$2.4$ {\scriptsize $[2.0,\,2.8]$}} & {$3.1$ {\scriptsize $[2.5,\,3.8]$}} & {$3.8$ {\scriptsize $[3.1,\,4.6]$}} \\
         & Random search & {$3.1$ {\scriptsize $[2.6,\,3.8]$}} & {$4.2$ {\scriptsize $[3.6,\,5.0]$}} & {$5.2$ {\scriptsize $[4.4,\,6.1]$}} \\
        \bottomrule
    \end{tabular}
\end{table}

\begin{table}[htbp]
    \centering
    \caption{Sample gains of \autoref{fig:fixed-vs-resampled}: simulated linear process, $n_{\text{tr}}=100$, $1{,}000$ seeds, mean squared error; frozen: a single split repeated $K$ times.}
    \label{tab:ci-figJ1}
    \small
    \begin{tabular}{@{} llccc @{}}
        \toprule
        \thead{Algorithm} & \thead{Split} & $K=5$ & $K=10$ & $K=20$ \\
        \midrule
        \texttt{Ridge} & reshuffled & {$2.6$ {\scriptsize $[2.2,\,3.1]$}} & {$3.4$ {\scriptsize $[3.0,\,3.9]$}} & {$3.9$ {\scriptsize $[3.5,\,4.5]$}} \\
         & frozen & $1.0^{\dagger}$ {\scriptsize $[1.0,\,1.0]$} & $1.0^{\dagger}$ {\scriptsize $[1.0,\,1.0]$} & $1.0^{\dagger}$ {\scriptsize $[1.0,\,1.0]$} \\
        \texttt{ExtraTrees} & reshuffled & {$2.8$ {\scriptsize $[2.3,\,3.3]$}} & {$3.3$ {\scriptsize $[2.6,\,4.2]$}} & {$3.8$ {\scriptsize $[3.2,\,4.5]$}} \\
         & frozen & $1.0^{\dagger}$ {\scriptsize $[1.0,\,1.0]$} & $1.0^{\dagger}$ {\scriptsize $[1.0,\,1.1]$} & $1.0^{\dagger}$ {\scriptsize $[1.0,\,1.1]$} \\
        \texttt{MLP} & reshuffled & {$4.0$ {\scriptsize $[3.4,\,4.6]$}} & {$5.8$ {\scriptsize $[5.1,\,6.8]$}} & {$6.7$ {\scriptsize $[6.0,\,7.6]$}} \\
         & frozen & {$1.9$ {\scriptsize $[1.6,\,2.2]$}} & {$2.0$ {\scriptsize $[1.7,\,2.4]$}} & {$2.1$ {\scriptsize $[1.8,\,2.5]$}} \\
        \texttt{RandomFeatures} & reshuffled & {$4.5$ {\scriptsize $[3.6,\,5.7]$}} & {$7.0$ {\scriptsize $[5.7,\,8.5]$}} & {$9.3$ {\scriptsize $[7.6,\,11.4]$}} \\
         & frozen & {$2.1$ {\scriptsize $[1.6,\,2.7]$}} & {$2.1$ {\scriptsize $[1.3,\,3.3]$}} & {$2.3$ {\scriptsize $[1.4,\,3.7]$}} \\
        \bottomrule
    \end{tabular}
\end{table}

\begin{table}[htbp]
    \centering
    \caption{Sample gains of \autoref{fig:pcam_mccv_vs_kfold}: \texttt{PCam}, {$n_{\text{tr}}=240$}, accuracy, $50$ seeds.}
    \label{tab:ci-figF1}
    \small
    \begin{tabular}{@{} llccc @{}}
        \toprule
        \thead{Network} & \thead{Procedure} & $K=5$ & $K=10$ & $K=20$ \\
        \midrule
        \texttt{DenseNet121} & MCCV & {$3.5$ {\scriptsize $[2.7,\,5.2]$}} & {$6.0$ {\scriptsize $[4.5,\,8.6]$}} & {$7.7$ {\scriptsize $[5.6,\,12.4]$}} \\
         & {Repeated 5-fold} & {$6.3$ {\scriptsize $[4.6,\,9.3]$}} & {$5.4$ {\scriptsize $[4.0,\,7.9]$}} & {$6.6$ {\scriptsize $[5.0,\,9.8]$}} \\
        \texttt{MobileNetV2} & MCCV & {$3.2$ {\scriptsize $[2.4,\,4.8]$}} & {$5.2$ {\scriptsize $[3.6,\,7.9]$}} & {$6.8$ {\scriptsize $[4.9,\,10.7]$}} \\
         & {Repeated 5-fold} & {$3.9$ {\scriptsize $[2.7,\,6.3]$}} & {$5.4$ {\scriptsize $[4.0,\,8.2]$}} & {$7.1$ {\scriptsize $[5.4,\,10.5]$}} \\
        \texttt{WideResNet101\_2} & MCCV & {$4.4$ {\scriptsize $[3.0,\,7.0]$}} & {$7.0$ {\scriptsize $[4.9,\,11.3]$}} & {$10.2$ {\scriptsize $[7.0,\,17.7]$}} \\
         & {Repeated 5-fold} & {$5.0$ {\scriptsize $[3.5,\,7.4]$}} & {$6.1$ {\scriptsize $[4.2,\,9.9]$}} & {$8.5$ {\scriptsize $[6.3,\,12.7]$}} \\
        \bottomrule
    \end{tabular}
\end{table}

\begin{table}[htbp]
    \centering
    \caption{Sample gains of \autoref{fig:stratified-overlay}: \texttt{PCam}, {$n_{\text{tr}}=400$}, negative log-likelihood, $100$ seeds, against the stratified single-split baseline.}
    \label{tab:ci-figK1}
    \small
    \begin{tabular}{@{} llccc @{}}
        \toprule
        \thead{Network} & \thead{Resampling} & $K=5$ & $K=10$ & $K=20$ \\
        \midrule
        \texttt{DenseNet121} & unstratified & {$3.7$ {\scriptsize $[2.8,\,4.8]$}} & {$4.7$ {\scriptsize $[3.6,\,6.5]$}} & {$6.3$ {\scriptsize $[4.7,\,8.9]$}} \\
         & stratified & {$3.9$ {\scriptsize $[3.1,\,5.1]$}} & {$6.4$ {\scriptsize $[4.7,\,9.0]$}} & {$8.3$ {\scriptsize $[6.5,\,11.1]$}} \\
        \texttt{MobileNetV2} & unstratified & {$3.7$ {\scriptsize $[2.6,\,5.3]$}} & {$4.4$ {\scriptsize $[3.2,\,6.1]$}} & {$5.8$ {\scriptsize $[4.3,\,8.0]$}} \\
         & stratified & {$3.6$ {\scriptsize $[2.7,\,5.3]$}} & {$5.0$ {\scriptsize $[3.6,\,7.5]$}} & {$5.2$ {\scriptsize $[4.0,\,7.2]$}} \\
        \texttt{WideResNet101\_2} & unstratified & {$3.6$ {\scriptsize $[2.7,\,5.0]$}} & {$6.2$ {\scriptsize $[4.7,\,8.6]$}} & {$8.4$ {\scriptsize $[6.3,\,11.7]$}} \\
         & stratified & {$4.0$ {\scriptsize $[3.0,\,5.9]$}} & {$7.3$ {\scriptsize $[5.4,\,10.5]$}} & {$9.3$ {\scriptsize $[7.1,\,12.6]$}} \\
        \texttt{ViT-B/16} & unstratified & $1.0^{\dagger}$ {\scriptsize $[1.0,\,1.0]$} & $1.0^{\dagger}$ {\scriptsize $[1.0,\,1.0]$} & $1.0^{\dagger}$ {\scriptsize $[1.0,\,1.0]$} \\
         & stratified & {$4.9^{\dagger}$ {\scriptsize $[1.0,\,12.1]$}} & {$9.0$ {\scriptsize $[2.8,\,18.0]$}} & {$12.0$ {\scriptsize $[3.3,\,24.9]$}} \\
        \bottomrule
    \end{tabular}
\end{table}

\begin{table}[htbp]
    \centering
    \caption{Sample gains of \autoref{fig:metrics-gain}: \texttt{PCam}, \texttt{WideResNet101\_2}, {$n_{\text{tr}}=400$}, stratified resampling, $100$ seeds.}
    \label{tab:ci-figL1}
    \small
    \begin{tabular}{@{} lccc @{}}
        \toprule
        \thead{Metric} & $K=5$ & $K=10$ & $K=20$ \\
        \midrule
        accuracy & {$3.8$ {\scriptsize $[2.9,\,5.0]$}} & {$6.1$ {\scriptsize $[4.6,\,8.2]$}} & {$8.9$ {\scriptsize $[6.9,\,11.8]$}} \\
        Brier score & {$3.9$ {\scriptsize $[2.9,\,5.4]$}} & {$5.9$ {\scriptsize $[4.5,\,8.2]$}} & {$8.5$ {\scriptsize $[6.8,\,11.0]$}} \\
        negative log-likelihood & {$4.0$ {\scriptsize $[3.0,\,5.9]$}} & {$7.3$ {\scriptsize $[5.4,\,10.5]$}} & {$9.3$ {\scriptsize $[7.1,\,12.6]$}} \\
        \bottomrule
    \end{tabular}
\end{table}

\begin{table}[htbp]
    \centering
    \caption{Sample gains at the small budgets of \autoref{appendix:label-scarce} (\autoref{fig:small-label-gain} shows $n_{\text{tr}} = 30$ and $100$): simulated linear process, $1{,}000$ seeds, mean squared error.}
    \label{tab:ci-figM1}
    \small
    \begin{tabular}{@{} llccc @{}}
        \toprule
        \thead{$n_{\text{tr}}$} & \thead{Algorithm} & $K=5$ & $K=10$ & $K=20$ \\
        \midrule
        $5$ & \texttt{Ridge} & {$1.8$ {\scriptsize $[1.5,\,2.1]$}} & {$2.0$ {\scriptsize $[1.7,\,2.3]$}} & {$2.2$ {\scriptsize $[1.9,\,2.5]$}} \\
         & \texttt{ExtraTrees} & {$2.3$ {\scriptsize $[1.6,\,2.7]$}} & {$2.4$ {\scriptsize $[1.7,\,3.2]$}} & {$2.6$ {\scriptsize $[1.8,\,3.2]$}} \\
         & \texttt{MLP} & {$2.3$ {\scriptsize $[1.8,\,2.8]$}} & {$2.7$ {\scriptsize $[2.2,\,3.2]$}} & {$2.9$ {\scriptsize $[2.4,\,3.4]$}} \\
         & \texttt{RandomFeatures} & {$1.6$ {\scriptsize $[1.2,\,2.2]$}} & {$2.1$ {\scriptsize $[1.7,\,2.6]$}} & {$2.5$ {\scriptsize $[2.2,\,2.9]$}} \\
        \midrule
        $10$ & \texttt{Ridge} & {$2.3$ {\scriptsize $[2.0,\,2.7]$}} & {$2.7$ {\scriptsize $[2.2,\,3.2]$}} & {$2.8$ {\scriptsize $[2.4,\,3.4]$}} \\
         & \texttt{ExtraTrees} & {$2.4$ {\scriptsize $[2.0,\,2.8]$}} & {$2.9$ {\scriptsize $[2.6,\,3.3]$}} & {$3.2$ {\scriptsize $[2.7,\,3.7]$}} \\
         & \texttt{MLP} & {$2.5$ {\scriptsize $[2.0,\,3.1]$}} & {$2.9$ {\scriptsize $[2.3,\,3.6]$}} & {$3.7$ {\scriptsize $[3.2,\,4.4]$}} \\
         & \texttt{RandomFeatures} & {$2.7$ {\scriptsize $[2.2,\,3.3]$}} & {$3.4$ {\scriptsize $[2.7,\,4.2]$}} & {$4.0$ {\scriptsize $[3.3,\,4.9]$}} \\
        \midrule
        $15$ & \texttt{Ridge} & {$2.6$ {\scriptsize $[2.3,\,3.0]$}} & {$3.4$ {\scriptsize $[3.0,\,3.9]$}} & {$4.1$ {\scriptsize $[3.6,\,4.7]$}} \\
         & \texttt{ExtraTrees} & {$2.6$ {\scriptsize $[2.2,\,3.0]$}} & {$3.0$ {\scriptsize $[2.5,\,3.4]$}} & {$3.3$ {\scriptsize $[2.8,\,3.9]$}} \\
         & \texttt{MLP} & {$2.7$ {\scriptsize $[2.2,\,3.4]$}} & {$3.6$ {\scriptsize $[3.0,\,4.5]$}} & {$4.1$ {\scriptsize $[3.4,\,5.1]$}} \\
         & \texttt{RandomFeatures} & {$2.8$ {\scriptsize $[1.9,\,4.0]$}} & {$3.7$ {\scriptsize $[2.5,\,5.1]$}} & {$5.0$ {\scriptsize $[4.2,\,6.1]$}} \\
        \midrule
        $20$ & \texttt{Ridge} & {$2.8$ {\scriptsize $[2.4,\,3.4]$}} & {$3.4$ {\scriptsize $[2.9,\,4.1]$}} & {$4.1$ {\scriptsize $[3.6,\,4.7]$}} \\
         & \texttt{ExtraTrees} & {$2.6$ {\scriptsize $[2.2,\,3.2]$}} & {$2.9$ {\scriptsize $[2.1,\,3.9]$}} & {$3.1$ {\scriptsize $[2.3,\,4.1]$}} \\
         & \texttt{MLP} & {$3.2$ {\scriptsize $[2.6,\,4.0]$}} & {$4.4$ {\scriptsize $[3.7,\,5.2]$}} & {$5.3$ {\scriptsize $[4.6,\,6.2]$}} \\
         & \texttt{RandomFeatures} & {$3.0$ {\scriptsize $[2.3,\,4.0]$}} & {$4.5$ {\scriptsize $[3.7,\,5.9]$}} & {$5.7$ {\scriptsize $[4.7,\,7.3]$}} \\
        \midrule
        $25$ & \texttt{Ridge} & {$2.6$ {\scriptsize $[2.2,\,3.0]$}} & {$3.0$ {\scriptsize $[2.5,\,3.6]$}} & {$3.4$ {\scriptsize $[3.0,\,4.0]$}} \\
         & \texttt{ExtraTrees} & {$2.5$ {\scriptsize $[2.1,\,3.0]$}} & {$3.2$ {\scriptsize $[2.9,\,3.7]$}} & {$3.8$ {\scriptsize $[3.3,\,4.4]$}} \\
         & \texttt{MLP} & {$3.2$ {\scriptsize $[2.7,\,4.0]$}} & {$4.7$ {\scriptsize $[4.1,\,5.4]$}} & {$5.6$ {\scriptsize $[4.9,\,6.5]$}} \\
         & \texttt{RandomFeatures} & {$3.7$ {\scriptsize $[3.0,\,4.4]$}} & {$5.6$ {\scriptsize $[4.8,\,6.7]$}} & {$6.9$ {\scriptsize $[5.9,\,8.1]$}} \\
        \midrule
        $30$ & \texttt{Ridge} & {$2.9$ {\scriptsize $[2.6,\,3.2]$}} & {$3.6$ {\scriptsize $[3.3,\,4.2]$}} & {$4.1$ {\scriptsize $[3.6,\,4.7]$}} \\
         & \texttt{ExtraTrees} & {$2.6$ {\scriptsize $[2.3,\,3.2]$}} & {$3.5$ {\scriptsize $[3.1,\,4.2]$}} & {$4.2$ {\scriptsize $[3.6,\,4.8]$}} \\
         & \texttt{MLP} & {$3.4$ {\scriptsize $[2.9,\,4.1]$}} & {$4.2$ {\scriptsize $[3.5,\,5.3]$}} & {$5.3$ {\scriptsize $[4.5,\,6.3]$}} \\
         & \texttt{RandomFeatures} & {$3.6$ {\scriptsize $[2.8,\,4.6]$}} & {$4.7$ {\scriptsize $[3.6,\,5.9]$}} & {$6.5$ {\scriptsize $[5.3,\,8.2]$}} \\
        \midrule
        $50$ & \texttt{Ridge} & {$2.5$ {\scriptsize $[2.2,\,2.8]$}} & {$3.2$ {\scriptsize $[2.8,\,3.6]$}} & {$3.7$ {\scriptsize $[3.3,\,4.2]$}} \\
         & \texttt{ExtraTrees} & {$2.4$ {\scriptsize $[2.0,\,2.8]$}} & {$2.9$ {\scriptsize $[2.4,\,3.5]$}} & {$3.4$ {\scriptsize $[2.9,\,4.0]$}} \\
         & \texttt{MLP} & {$3.0$ {\scriptsize $[2.5,\,3.6]$}} & {$4.4$ {\scriptsize $[3.7,\,5.2]$}} & {$5.4$ {\scriptsize $[4.6,\,6.4]$}} \\
         & \texttt{RandomFeatures} & {$4.2$ {\scriptsize $[3.4,\,5.1]$}} & {$5.3$ {\scriptsize $[4.0,\,7.4]$}} & {$7.5$ {\scriptsize $[6.1,\,9.5]$}} \\
        \midrule
        $100$ & \texttt{Ridge} & {$2.6$ {\scriptsize $[2.3,\,3.1]$}} & {$3.7$ {\scriptsize $[3.3,\,4.2]$}} & {$4.3$ {\scriptsize $[3.7,\,4.9]$}} \\
         & \texttt{ExtraTrees} & {$2.6$ {\scriptsize $[2.1,\,3.3]$}} & {$3.5$ {\scriptsize $[3.0,\,4.1]$}} & {$4.0$ {\scriptsize $[3.5,\,4.6]$}} \\
         & \texttt{MLP} & {$3.5$ {\scriptsize $[3.0,\,4.2]$}} & {$5.0$ {\scriptsize $[4.2,\,5.9]$}} & {$6.7$ {\scriptsize $[5.8,\,7.9]$}} \\
         & \texttt{RandomFeatures} & {$5.1$ {\scriptsize $[3.8,\,6.9]$}} & {$7.7$ {\scriptsize $[6.3,\,9.7]$}} & {$10.5$ {\scriptsize $[8.6,\,12.8]$}} \\
        \bottomrule
    \end{tabular}
\end{table}

\paragraph{Construction} Each seed is one complete {replication}: its own study set, its $K$ splits, the first of which serves as the single hold-out, and its outer test sets of growing size. $G^{\text{test}}_K$ compares the variance across seeds of the $K$-split error with the variance across seeds of the single-split error as a function of the test size, and the estimate computes both on the observed seeds. The interval resamples the seeds with replacement ($1{,}000$ draws, fixed random seed), recomputes both variances and their crossing on each resampled set, and takes the $2.5\%$ and $97.5\%$ quantiles of the resulting gains, separately at each $K$. A draw uses the same resampled seeds for both variances, so it reproduces their dependence: the first split counts both in the single-split baseline and in the $K$-split average, and a set of seeds with a large single-split variance also tends to have a large $K$-split variance. {For \texttt{Qwen3-0.6B}, whose single-split baseline comes from separate runs with the same seeds, study sets and first split, the draws also resample the same seeds on both sides.} The intervals hold at each $K$ separately. They are not simultaneous bands. The thin lines of the figures are draws of the whole estimator, not gains of individual seeds. Their mean is not used as the estimate: for a skewed ratio of variances it adds the resampling bias to that of the estimate, and at $50$ seeds it lies {$12$ to $19\%$} above the $1{,}000$-seed value, against {$3$ to $14\%$} for the estimate.

\paragraph{Calibration} We checked the coverage of these intervals on the simulated runs of \autoref{fig:avg_test_sample_gain_simulated}. Splitting their $1{,}000$ seeds into $20$ disjoint groups of $50$, the size of our real-data experiments, the $95\%$ intervals of the four algorithms at $K = 5$, $10$ and $20$ contain the $1{,}000$-seed estimate in {$216$ of $240$ cases ($90.0\%$), and with groups of $200$ seeds in $56$ of $60$ ($93.3\%$). The shortfall is not chance: replaying this check on $1{,}000$ experiments resampled from the same seeds gives $89.6\%$ on average, against $95.5\%$ for intervals of exact coverage. The $K$-split errors are heavy-tailed across seeds, and $50$ seeds rarely contain the few that inflate the $K$-split variance. The estimate therefore leans high (by $3$ to $14\%$ in median across algorithms), and $20$ of the $24$ misses lie above the reference. Taking the $1{,}000$ seeds as the population, the actual coverage is $89.0\%$ at $50$ seeds and $91.7\%$ at $200$, and bias-corrected (BC, BCa), basic and log-scale bootstrap intervals reach $86$ to $90\%$ at $50$ seeds. Our $95\%$ intervals are thus about $90\%$ intervals at the real-data size.} Each group shares its seeds with the reference ($5\%$ or $20\%$ of them), which makes these coverages slightly optimistic.

\paragraph{Limits} The estimator returns exactly $1$ when the $K$-split variance does not fall below the single-split one, and it is capped at the largest test size available in the outer set. Draws piling up at these limits make a percentile interval unreliable. In the tables, $^{\dagger}$ marks the cells where at least $2.5\%$ of the draws equal $1$. This happens by construction for the frozen \texttt{Ridge} and \texttt{ExtraTrees} of \autoref{tab:ci-figJ1} and for the unstratified \texttt{ViT-B/16} of \autoref{tab:ci-figK1}, whose gains are $1$. No tabulated cell has $2.5\%$ of its draws at the cap. The real-data intervals, from $50$ to $100$ seeds, are wide and right-skewed: their upper bounds reflect draws in which the single-split variance is large relative to the $K$-split one. Some exceed $K$ ({most real-data upper bounds at $K = 5$ and $10$, and at $K = 20$ those of the stratified \texttt{ViT-B/16} of \autoref{tab:ci-figK1} and of the random-search \texttt{MLP}}), a sign of the noise of the single-split side discussed in {Appendix}~\ref{sec:sample-gain-estimation} rather than of more than $K$ hold-outs' worth of information.

\FloatBarrier
\section{Experimental details}
\label{appendix:experimental_details}

\subsection{Simulated data process}
\label{appendix:simulated-process}
All simulated datasets have normally distributed features and contain $150{,}000$ observations.
A held-out benchmarking set of $100{,}000$ observations is
used to estimate oracle performance, while the remaining observations form the
pool from which the study and outer sets are sampled.  Every random seed regenerates the covariates, the coefficients and the noise.

\begin{table}[t]
    \centering
    \caption{Dataset splittings in our experiments.}
    \label{tab:xp_splitting}
    \begin{tabular}{@{} l c c @{}}
        \toprule
        \thead{Dataset} & \thead{Training\\sizes} & \thead{Benchmarking\\size} \\
        \midrule
        Simulated & 50; 75; 100; 300; 750; 1,000; 2,000; 3,000; 5,000; 10,000& 100,000\\
        PatchCamelyon & 24; 80; 240; 400; 640& 244,912\\
        Yelp Review Full & {1,000; 3,000} & 600,000\\
        \bottomrule
    \end{tabular}

    \vspace{0.6em}
    \begin{minipage}{0.86\textwidth}
    \footnotesize
    Additional configurations are used by the robustness appendices. The small-budget sweep of \autoref{appendix:label-scarce} extends the simulated training sizes down to $5$ samples ($5$, $10$, $15$, $20$, $25$, $30$, $50$ and $100$) and adds a second set of \texttt{PCam} runs at $n_{\text{tr}}=24$ (four networks, $50$ seeds), which records the negative log-likelihood and the Brier score. The ranking analysis of \autoref{appendix:ranking-sample-gain} uses its own grid of $25$ simulated training sizes from $10$ to $10{,}000$ ($K=200$, {with} $500$ seeds for the \texttt{ExtraTrees} family {and} $100$ for the \texttt{Ridge} family). The hyper-parameter-search runs of \autoref{appendix:hpo} and the fixed-versus-reshuffled comparison of \autoref{appendix:fixed-vs-reshuffled} use the simulated process at the training sizes listed above. The stratified re-runs of \autoref{appendix:stratification} repeat the \texttt{PCam} configuration at $n_{\text{tr}}=400$ for the four networks, with $100$ seeds, unchanged apart from the fold-level stratification. The redundancy runs of \autoref{tab:pcam-redundancy} train the four \texttt{PCam} networks at $n_{\text{tr}} = 100$, $300$ and $800$, with $50$ further seeds, and {estimate} the oracle {risks} on a fixed, class-stratified subset of $60{,}000$ benchmarking images. For \texttt{PCam} and \texttt{Yelp}, $n_{\text{tr}}$ counts the training and validation subsets together, and the study set has $1.25\,n_{\text{tr}}$ observations.
    \end{minipage}
\end{table}

The baseline simulated dataset is a linear Gaussian regression.  We draw
$X_i\sim{\mathcal{N}_d(0,I_d)}$, $\beta\sim{\mathcal{N}_d(0,I_d)}$, and
$\varepsilon_i\sim\mathcal{N}(0,1)$ independently, and set
\[
    Y_i = X_i^\top\beta + \sigma\varepsilon_i .
\]
The main runs (\autoref{sec:experimental-evaluation} and Appendices~\ref{appendix:hpo}, \ref{appendix:fixed-vs-reshuffled}, \ref{appendix:metrics}, \ref{appendix:label-scarce} and \ref{appendix:ranking-sample-gain}) use this process without noise ($\sigma=0$) in dimension $d=5$. The redundancy analyses use $d=20$ and noise levels from $\sigma=0.1$ to $2.5$, listed per design in {Appendix}~\ref{appendix:redundancy-runs}.

To avoid calibrating the study-only redundancy rule on a single linear data
process, the redundancy designs also include{} two harder synthetic families.  The first one adds all
pairwise feature interactions:
\[
    Y_i
    =
    X_i^\top\beta
    +
    \sum_{1\leq a<b\leq d}\gamma_{ab}X_{ia}X_{ib}
    +
    \sigma\varepsilon_i ,
\]
where the interaction coefficients are Gaussian and scaled by the inverse
square root of the number of feature pairs.  The second one is an additive
nonlinear regression,
\[
    Y_i
    =
    \sum_{j=1}^d \beta_j \sin(4X_{ij})
    +
    \sigma\varepsilon_i .
\]
They are run only in those designs, with the same dimension and noise levels.

The test-fraction check of {Appendix}~\ref{sec:study-only-redundancy-statistics} also uses binary labels: with $X_i$ and $\beta$ as above and $d=20$, $Y_i\sim\mathrm{Bernoulli}\bigl(\operatorname{sigmoid}(s\,X_i^\top\beta/\sqrt{d})\bigr)$ with $s=1$. Its regression arm uses the linear process with $d=20$ and $\sigma=1$.

\subsection{Simulated runs used for the redundancy}
\label{appendix:redundancy-runs}
The redundancy analyses of \autoref{sec:empirical-results} and {Appendix}~\ref{sec:study-only-redundancy-statistics} pool four simulated designs{, $663$ configurations in all. The fresh-seed confirmation of {Appendix}~\ref{sec:study-only-redundancy-statistics}, whose design is given there, is analysed separately}. All use the processes above with $d=20$ features and $150{,}000$ observations per dataset, MCCV with test fraction $0.2$ and $K=200$ splits, and gains on the mean squared error. Every seed regenerates the data.
\begin{itemize}
    \item \emph{Noise-injection grid} ($150$ configurations, $30$ seeds): linear process with $\sigma\in\{0.1,0.5,1.5\}$, $n_{\text{tr}}\in\{50,100,300,1{,}000,3{,}000\}$. {The learning algorithms are} \texttt{Ridge} ($\alpha=1$), \texttt{DecisionTree} and \texttt{ExtraTrees} ($200$ trees), and two perturbed versions of \texttt{Ridge} that vary the redundancy in a controlled way. The first adds zero-mean Gaussian noise to the predictions, with standard deviation $c$ times that of the training targets ($c\in\{0.1,0.3,1,3\}$). The noise is drawn afresh at each fit, so that different splits perturb a given observation independently, while a given fit perturbs an observation identically wherever it is predicted. The second perturbs the fitted coefficients once per fit by $c$ times their root-mean-square ($c\in\{0.1,0.3,1\}$).
    \item \emph{Pre-registered holdout} ($243$ configurations, $50$ seeds): the three processes with $\sigma\in\{0.3,1,2.5\}$, $n_{\text{tr}}\in\{75,750,5{,}000\}$, and the $9$ learning algorithms of {Appendix}~\ref{sec:study-only-redundancy-statistics}.
    \item \emph{Two unstable-learner designs} ($162$ and $108$ configurations, $50$ seeds each): the three processes with $\sigma\in\{0.3,1\}$. {The first uses} $n_{\text{tr}}\in\{300,1{,}000,3{,}000\}$ with \texttt{Ridge} {($\alpha=1$)}, a depth-$20$ decision tree, \texttt{ExtraTrees} and random forests of $2$ or $5$ trees, $1$- and $5$-nearest neighbours and a one-hidden-layer \texttt{MLP} ($32$ units){, and the second uses} $n_{\text{tr}}\in\{1{,}000,3{,}000,10{,}000\}$ with \texttt{Ridge} and the same trees and forests.
\end{itemize}

\needspace{5\baselineskip}
\subsection{Real data training loops}
\label{appendix:real-training-loops}

{For \texttt{PCam}, a global stratified split, the same for every seed, holds out $244{,}912$ images as the benchmarking set. The remaining $50{,}000$ images form the pool from which each seed draws its study set, the rest of the pool being the outer set. For \texttt{Yelp}, the \texttt{BERT} and \texttt{XLM-RoBERTa} runs follow the same protocol, with $600{,}000$ reviews in the benchmarking set and a pool of $50{,}000$. The \texttt{Qwen3-0.6B} runs predate it: each of their seeds shuffles the $650{,}000$ training reviews, the first $600{,}000$ forming the benchmarking set, the next ones the study set and the remaining ones the outer set, so that their benchmarking set is not stratified and changes with the seed. The outer set is used in chunks of the size of a test fold, up to $200$ of them.}

To assess model performance across different data scales, we design a cross-validation protocol that operates on subsets, referred to as study sets, sampled from the main study pool{\ (with stratification, except for the \texttt{Qwen3-0.6B} runs). The procedure is repeated over $50$ or $100$ random seeds, depending on the run set. \autoref{tab:xp_time_long} and the captions give the number for each}.

\paragraph{Histopathologic scans classification} The \texttt{PCam} dataset is sourced via the Hugging Face \texttt{datasets} library \citep{lhoest_datasets_2021}, utilizing the \texttt{1aurent/PatchCamelyon} repository. The official training and validation splits are combined into a single unified dataset containing 294,912 histopathological images. The only preprocessing step applied is the conversion of images into \texttt{PyTorch} tensors{, followed for \texttt{ViT-B/16} by a resizing from $96 \times 96$ to $224 \times 224$ pixels}.

The models used in this study are {three CNNs, \texttt{DenseNet121}, \texttt{MobileNetV2} and \texttt{WideResNet101\_2}, and the transformer \texttt{ViT-B/16}, all} available in the \texttt{torchvision} library {\citep{torchvision2016}} {with weights} pretrained on \texttt{ImageNet} {\citep{deng_imagenet_2009, russakovsky_imagenet_2015}}. For each selected architecture, the final fully-connected classification layer was removed{, and a new linear layer maps the backbone's features to the two output classes of the \texttt{PCam} dataset: the convolutional features after adaptive average pooling for the CNNs, the class token of the encoder for \texttt{ViT-B/16}. No layer is frozen: the backbone is fine-tuned together with the new head}.

For each experimental run, defined by a specific model architecture, study set size, and random seed, the following steps are executed:
\begin{enumerate}
    \item Study set sampling: A study set of a specific size ({from $30$ to $1{,}000$ images}) is drawn using stratified sampling to maintain the original class distribution.
    \item Fold-Level Splitting: A CV is initiated on this study set. As the dataset is balanced, stratification is not performed by default. \autoref{appendix:stratification} reports stratified re-runs. Within each fold, the data is partitioned into three distinct subsets. In the terminology of \autoref{fig:dataset_terminology}, the portion labelled \emph{train} is the union of the training and validation subsets listed below. The test subset is held out from selection entirely, so that the benchmarked object is the whole procedure, checkpoint selection included (\autoref{sec:problem-setting}):
    \begin{itemize}
        \item A test set, comprising $20\%$ of the study set's data. This set is held out until the final evaluation for the fold.
        \item A validation set, also comprising $20\%$ of the study set's data.
        \item A training set, comprising the remaining $60\%$ of the study set's data.
    \end{itemize}
    \item Model training: The {network} is trained exclusively on the training set for {$30$ epochs, with batches of $64$ images,} using the Adam optimizer \citep{kingma_adam_2015} {with its default settings (learning rate $10^{-3}$)} and Cross-Entropy loss function.
    \item Model selection: After each training epoch, the model's performance is evaluated on the validation set. The model weights that yield the highest validation accuracy across all epochs are saved as the best model for that fold.
\end{enumerate}

Upon completion of training for a fold, the best-performing model is loaded and evaluated on four separate datasets:
\begin{enumerate}
    \item The validation set for that fold.
    \item The held-out test set for that fold, providing an unbiased estimate of generalization performance on the study set.
    \item The outer set, also providing unbiased estimates of generalization, but on samples that are not part of the CV in that seed{. It is scored after the first split of each run only, in chunks of the test-fold size}.
    \item The global benchmarking set, providing an estimate of oracle performance on unseen data.
\end{enumerate}
For each evaluation, we record the accuracy, loss, and inference runtime{. The later runs (\autoref{appendix:stratification}, \autoref{appendix:metrics} and \autoref{tab:pcam-redundancy}) also record the Brier score and the per-sample losses}. All experimental parameters (model name, study size, seed, fold) and the resulting performance metrics are systematically logged{\ and saved as Parquet files} for subsequent analysis.

\paragraph{NLP fine-tuning} The experiments on the \texttt{Yelp Review Full} dataset build on the HuggingFace fine-tuning tutorial\footnote{\url{https://huggingface.co/docs/transformers/main/en/training}}, adapted to our cross-validation protocol and, for the decoder-only model, to sequence classification. We fine-tune three checkpoints through \texttt{AutoModelForSequenceClassification} with a $5$-class head: \texttt{BERT} (\texttt{google-bert/bert-base-cased}), \texttt{XLM-RoBERTa} (\texttt{FacebookAI/xlm-roberta-base}) and \texttt{Qwen3-0.6B} (\texttt{Qwen/Qwen3-0.6B}). All three are \emph{fully} fine-tuned (every parameter is updated, with no adapter or LoRA) under identical optimisation settings (below), a per-device training batch size of $32$ with no gradient accumulation or checkpointing, mixed precision (\texttt{bf16} where supported, otherwise \texttt{fp16}), and inputs tokenised with truncation to $512$ tokens {and padded to the longest input of each batch (of each block of $1{,}000$ reviews for \texttt{Qwen3-0.6B})}. \texttt{Qwen3-0.6B} is decoder-only and about five times larger than the two encoders ($0.6$B versus $\sim0.11$B parameters): for it, \texttt{AutoModelForSequenceClassification} reads the classification head off the representation of the final non-padding token, and we set the tokenizer pad token to its end-of-sequence token and propagate the resulting \texttt{pad\_token\_id} to the model config. Because of its cost, \texttt{Qwen3-0.6B} was run only at the smallest training size ($n_{\text{tr}}=1{,}000$), whereas the two encoders were run at {both sizes}.

Within each experiment, the CV procedure is the same as the one used for \texttt{PCam}, detailed above.

For each CV fold, a pre-trained model is fine-tuned on the training fold. The training process is managed by the HuggingFace \texttt{Trainer} with the following hyperparameters:
\begin{itemize}
    \item Epochs: 3
    \item Optimizer: AdamW \citep{loshchilov_decoupled_2019} (default{, weight decay left at its default of $0$})
    \item Learning rate: 5e-5 (default)
    \item Training batch size: 32
    \item Evaluation strategy: The model's performance on the validation fold is evaluated at the end of each epoch.
    \item Model selection: The model checkpoint that achieves the highest accuracy on the validation fold is selected as the best model for that fold.
\end{itemize}

After the training process for a single fold is complete, the best-performing model is evaluated on the validation {fold, the test fold} and the benchmarking set{. The model of the first split also scores the outer set, in chunks of the test-fold size, and serves as the single split, as for \texttt{PCam}. For \texttt{Qwen3-0.6B}, the outer set is scored in separate single-split runs, which share the seeds, the study sets and the first split of the $K$-split runs but train their own model}. The primary metric recorded for all evaluations is accuracy. We also record training and evaluation runtimes, losses, and the epoch at which the best model was found.

\subsection{Runs{, time and assets}}
\label{appendix:runs-and-time}
{\paragraph{Assets and licenses} \autoref{tab:assets} lists the datasets, pre-trained weights and main libraries that the experiments use, with the version used, the license or terms of use, and the work that introduced each. Hugging Face identifiers resolve at \url{https://huggingface.co/datasets/<id>} for the datasets and at \url{https://huggingface.co/<id>} for the models. \texttt{PatchCamelyon} is released under CC0, following the license of the Camelyon16 data it is built from. \texttt{Yelp Review Full} was built by \citet{zhang2015character} from the Yelp Dataset Challenge. Its Hugging Face card refers to the Yelp Dataset Terms of Use,\footnote{\url{https://s3-media3.fl.yelpcdn.com/assets/srv0/engineering_pages/bea5c1e92bf3/assets/vendor/yelp-dataset-agreement.pdf}} which allow non-commercial use only. The four \texttt{PCam} networks start from \texttt{torchvision} weights trained on \texttt{ImageNet} {\citep{deng_imagenet_2009, russakovsky_imagenet_2015}}, whose terms of access restrict its use to non-commercial research and education. {The} \texttt{torchvision} {library} leaves it to the user to check the terms of its pre-trained weights. All our uses are non-commercial research. The \texttt{PCam} training script adapts one shared by the authors of \texttt{TIAToolbox} \citep{pocock_tiatoolbox_2022}, a library under the BSD-3-Clause license that our code does not call.}

\begin{table}[htbp]
    \centering
    \caption{Details concerning the learning algorithms in our experiments. Displayed time values are in hours. {GB: GradientBoosting.} $^{\dagger}$\texttt{Qwen3-0.6B} was run only at the smallest training size ($n_{\text{tr}}=1{,}000$), so its times cover that single size, whereas the other \texttt{Yelp} rows aggregate {both}; being decoder-only and roughly five times larger, its benchmarking-set evaluation dominates its cost. $^{\ddagger}$Run on $40$-core Intel CPU nodes; times are core-hours from the job scheduler, evaluation included; rows marked exploratory, on either machine, informed design choices but are not reported in the paper, and the ranking runs of \autoref{appendix:ranking-sample-gain} are not included. The mean-squared-error gains of \autoref{fig:avg_test_sample_gain_simulated} and \autoref{fig:all_data_test_sample_gains} come from the re-run of the $1{,}000$-seed, $200$-split design listed there, which also records the mean squared error. $^{\S}$Time estimated by re-running $5\%$ of the seeds on the Intel CPU nodes with the current code, which records fitting and evaluation separately, and scaling to all seeds; core-hours as for $^{\ddagger}$ (fitting on these nodes takes $0.63$ to $0.87$ times as long as on the AMD machine). $^{\P}$GPU-hours from the job scheduler, study and benchmarking evaluation together.}
    \label{tab:xp_time_long}
    \scriptsize
    \setlength{\tabcolsep}{2.2pt}
    \renewcommand{\arraystretch}{0.92}
    \makebox[\textwidth][c]{%
    \resizebox{\textwidth}{!}{%
    \begin{tabular}{@{} l l c c c c c c @{}}
        \toprule
        \thead{Dataset} & \thead{Learning\\algorithm} & \thead{Hardware\\used} & \thead{No. of\\seeds} & \thead{No. of\\splits} & \thead{Study\\time} & \thead{Bench.\\time} & \thead{Total\\time} \\
        \midrule
        \multirow{40}{*}{Simulated}

        & ExtraTrees$[n_{\text{estimators}} = 200]$
        & \multirow{24}{*}{\makecell{AMD CPU\\(128 cores)}} & 1,000 & 200 & -- & -- & 231.62\\
        & GradientBoosting$[n_{\text{estimators}} = 200]$
        & & 1,000 & 200 & -- & -- & 206.24\\
        & MLP$[\text{hidden layers} = (32)]$
        & & 1,000 & 200 & -- & -- & 192.76\\
        & Ridge$[\alpha = 1613]$
        & & 1,000 & 200 & -- & -- & 0.88\\
        & & & & & & & \\[-4pt]

        & GradientBoosting$[n_{\text{estimators}} = 20]${, exploratory}
        & & 500 & 20 & -- & -- & 3.26\\
        & & & & & & & \\[-4pt]

        & DecisionTree$[\text{max depth} = 5]${, exploratory}
        & & 100 & 20 & -- & -- & 0.04\\
        & DecisionTree$[\text{max depth} = 20]${, exploratory}
        & & 100 & 20 & -- & -- & 0.07\\
        & DecisionTree$[\text{max depth} = 100]${, exploratory}
        & & 100 & 20 & -- & -- & 0.07\\
        & HistGradientBoosting$[\text{max depth} = 3]${, exploratory}
        & & 100 & 20 & -- & -- & 0.82\\
        & HistGradientBoosting$[\text{max depth} = 20]${, exploratory}
        & & 100 & 20 & -- & -- & 2.03\\
        & KNeighbors$[n_{\text{neighbors}} = 5]${, exploratory}
        & & 100 & 20 & -- & -- & 0.02\\
        & KNeighbors$[n_{\text{neighbors}} = 20]${, exploratory}
        & & 100 & 20 & -- & -- & 0.02\\
        & SVR$[C = 10, \epsilon = 0.1]${, exploratory}
        & & 100 & 20 & -- & -- & 0.70\\
        & & & & & & & \\[-4pt]

        & Redundancy: earlier design repeated by the fresh-seed confirmation ($216$ configurations)
        & & 50 & 200 & -- & -- & 298.64\\
        & & & & & & & \\[-4pt]
        & Ridge, GB, ExtraTrees, MLP$(32)$, $n_{\text{tr}}=100$, frozen split$^{\S}$
        & & 1,000 & 20 & -- & -- & 17.48\\
        & Ridge, GB, ExtraTrees, MLP$(32)$, $n_{\text{tr}}=100$, resampled$^{\S}$
        & & 1,000 & 20 & -- & -- & 17.25\\
        & Ridge, GB, ExtraTrees, MLP$(32)$, $n_{\text{tr}}=5$--$100$$^{\S}$
        & & 1,000 & 20 & -- & -- & 112.95\\
        & ExtraTrees grid search$^{\S}$
        & & 1,000 & 20 & -- & -- & 69.58\\
        & ExtraTrees random search$^{\S}$
        & & 1,000 & 20 & -- & -- & 50.62\\
        & & & & & & & \\[-4pt]
        & Ridge, GB, ExtraTrees, MLP$(32)$, mean squared error re-run$^{\ddagger}$
        & \multirow{16}{*}{\makecell{Intel CPU\\(40 cores)}} & 1,000 & 200 & -- & -- & 454.64\\
        & MLP$[\text{hidden layers} = (64, 64)]$, exploratory$^{\ddagger}$
        & & 1,000 & 200 & -- & -- & 48.37\\
        & MLP$[\text{hidden layers} = (64, 64)]$, $n_{\text{tr}}=100$, frozen and resampled, exploratory$^{\ddagger}$
        & & 1,000 & 20 & -- & -- & 15.76\\
        & RandomFeatures$[n_{\text{features}} = 50]^{\ddagger}$
        & & 1,000 & 200 & -- & -- & 41.58\\
        & RandomFeatures$[n_{\text{features}} = 50]$, $n_{\text{tr}}=100$, frozen and resampled$^{\ddagger}$
        & & 1,000 & 20 & -- & -- & 31.78\\
        & RandomFeatures$[n_{\text{features}} = 50]$, $n_{\text{tr}}=100$, frozen, data redrawn$^{\ddagger}$
        & & 1,000 & 20 & -- & -- & 8.96\\
        & RandomFeatures$[n_{\text{features}} = 50]$ and MLP$(64, 64)$, $n_{\text{tr}}=5$--$100$$^{\ddagger}$
        & & 1,000 & 20 & -- & -- & 56.89\\
        & MLP grid and random search$^{\ddagger}$
        & & 1,000 & 20 & -- & -- & 176.56\\
        & Ridge, GB, ExtraTrees, MLP$(32)$, per-sample errors (\autoref{appendix:metrics})$^{\ddagger}$
        & & 200 & 20 & -- & -- & 10.32\\
        & Ridge and RandomFeatures grid and random search, MLP$(64)$, exploratory$^{\ddagger}$
        & & 1,000 & 20 & -- & -- & 34.91\\
        & & & & & & & \\[-4pt]
        & Redundancy: noise-injection grid ($150$ configurations)$^{\ddagger}$
        & & 30 & 200 & -- & -- & 608.02\\
        & Redundancy: pre-registered holdout ($243$ configurations)$^{\ddagger}$
        & & 50 & 200 & -- & -- & 4,197.68\\
        & Redundancy: unstable-learner designs ($270$ configurations)$^{\ddagger}$
        & & 50 & 200 & -- & -- & 830.36\\
        & Redundancy: fresh-seed confirmation ($216$ configurations)$^{\ddagger}$
        & & 50 & 200 & -- & -- & 1,832.27\\
        & Redundancy: test-fraction check ({Appendix}~\ref{sec:study-only-redundancy-statistics})$^{\ddagger}$
        & & 50 & 20 & -- & -- & 18.40\\

        \midrule
        \multirow{12}{*}{PatchCamelyon}
        & DenseNet121 & & 50 & 20 & 33.72 & 90.60 & 124.32\\
        & MobileNetV2 & \multirow{11}{*}{\makecell{NVIDIA\\V100}} & 50 & 20 & 25.44 & 74.79 & 100.23\\
        & ViT-B/16 & & 50 & 20 & 121.71 & 1,101.76 & 1,223.46\\
        & WideResNet101\_2 & & 50 & 20 & 51.27 & 231.68 & 282.95\\
        & & & & & & & \\[-4pt]
        & All four, {$n_{\text{tr}} = 100$/$300$/$800$} (\autoref{tab:pcam-redundancy})$^{\P}$ & & 50 & 20 & -- & -- & 418.5\\
        & All four, stratified, {$n_{\text{tr}} = 400$} (\autoref{appendix:stratification})$^{\P}$ & & 100 & 20 & -- & -- & 724.1\\
        & All four, $n_{\text{tr}} = 24$ (\autoref{appendix:label-scarce})$^{\P}$ & & 50 & 20 & -- & -- & 313.7\\

        \midrule
        \multirow{3}{*}{Yelp Review Full}
        & BERT & \multirow{3}{*}{\makecell{NVIDIA\\H100}} & {50} & 20 & {6.23} & {93.25} & {99.48}\\
        & XLM-RoBERTa & & {50} & 20 & {7.74} & {90.71} & {98.44}\\
        & Qwen3-0.6B$^{\dagger}$ & & 50 & 20 & 9.02 & 502.87 & 511.89\\
        \bottomrule

    \end{tabular}%
    }%
    }
\end{table}

\FloatBarrier
\begin{table}[htbp]
    \centering
        \caption{External assets used by the experiments. Rev.: commit of the Hugging Face Hub repository (first $7$ characters). Library versions are those of the Jean-Zay environments: the module \texttt{pytorch-gpu/py3/2.8.0} for the \texttt{PCam}, \texttt{BERT} and \texttt{XLM-RoBERTa} runs, the module \texttt{pytorch-gpu/py3/2.11.0} for the \texttt{Qwen3-0.6B} runs, and a Python~$3.11$ environment for the simulated runs on the Intel CPU nodes; the figure and table scripts run in the first or the last.}
    \label{tab:assets}
    \footnotesize
    \begin{tabular}{@{} >{\raggedright\arraybackslash}p{0.30\textwidth} >{\raggedright\arraybackslash}p{0.38\textwidth} >{\raggedright\arraybackslash}p{0.23\textwidth} @{}}
        \toprule
        Asset & Version used & License or terms of use \\
        \midrule
        \multicolumn{3}{@{}l}{\emph{Datasets}} \\
        \texttt{PatchCamelyon} \citep{veeling2018rotation, ehteshami_bejnordi_diagnostic_2017} & \texttt{1aurent/PatchCamelyon}, rev.\ \texttt{e4bd149} & CC0 1.0 \\
        \texttt{Yelp Review Full} \citep{zhang2015character} & \texttt{yelp\_review\_full}, rev.\ \texttt{c1f9ee9} & Yelp Dataset Terms of Use (non-commercial use) \\
        \midrule
        \multicolumn{3}{@{}l}{\emph{Pre-trained weights}} \\
        \texttt{BERT} \citep{devlin_bert_2019} & \texttt{google-bert/bert-base-cased}, rev.\ \texttt{cd5ef92} & Apache-2.0 \\
        \texttt{XLM-RoBERTa} \citep{ConneauKGCWGGOZ20} & \texttt{FacebookAI/xlm-roberta-base}, rev.\ \texttt{e73636d} & MIT \\
        \texttt{Qwen3-0.6B} \citep{qwen3} & \texttt{Qwen/Qwen3-0.6B}, rev.\ \texttt{c1899de} & Apache-2.0 \\
        \texttt{DenseNet121} \citep{huang_densely_2017}, \texttt{MobileNetV2} \citep{sandler_mobilenetv2_2018}, \texttt{WideResNet101\_2} \citep{zagoruyko_wide_2016}, \texttt{ViT-B/16} \citep{DosovitskiyB0WZ21} & \texttt{torchvision} weights \texttt{IMAGENET1K\_V1} & BSD-3-Clause (\texttt{torchvision}), \texttt{ImageNet} terms of access (non-commercial use) \\
        \midrule
        \multicolumn{3}{@{}l}{\emph{Libraries}} \\
        \texttt{PyTorch} {\citep{paszke2019pytorch, ansel_pytorch2_2024}}, \texttt{torchvision} {\citep{torchvision2016}} & $2.8.0$ and $0.23.0$ ($2.11.0$ and $0.26.0$ for \texttt{Qwen3-0.6B}) & BSD-3-Clause \\
        \texttt{Transformers} \citep{wolf_transformers_2020} & $4.55.4$ ($5.7.0$ for \texttt{Qwen3-0.6B}) & Apache-2.0 \\
        \texttt{Datasets} \citep{lhoest_datasets_2021} & $4.0.0$ ($4.8.5$ for \texttt{Qwen3-0.6B}) & Apache-2.0 \\
        \texttt{scikit-learn} \citep{pedregosa_scikit-learn_2011} & $1.9.1$ (simulated runs on the Intel CPU nodes), $1.7.1$ and $1.8.0$ (GPU runs) & BSD-3-Clause \\
        \texttt{Benchopt} \citep{moreau_benchopt_2022} & $1.9.0$ (Intel CPU nodes) & BSD-3-Clause \\
        \texttt{NumPy} \citep{harris_array_2020}, \texttt{SciPy} \citep{virtanen_scipy_2020}, \texttt{pandas} {\citep{mckinney_data_2010, pandas_zenodo_2020}}, \texttt{seaborn} \citep{waskom_seaborn_2021} & those of the environments above & BSD-3-Clause \\
        \texttt{Matplotlib} \citep{hunter_matplotlib_2007} & that of the environments above & Matplotlib license (based on the PSF license) \\
        \bottomrule
    \end{tabular}
    \end{table}

\FloatBarrier
\section{Sample gains under hyper-parameter optimisation}
\label{appendix:hpo}

The estimand of \eqref{eq:oracle_score} is the oracle {risk} of an entire learning procedure, which may include an internal hyper-parameter search. This appendix reports sample gains for such procedures.

\paragraph{Protocol} On the simulated regression process, we compare three variants of two algorithms: fixed hyper-parameters, grid search, and random search. For \texttt{ExtraTrees} the search ranges over \texttt{max\_features}, \texttt{min\_samples\_leaf} and \texttt{max\_depth}, with the number of trees held fixed. For the \texttt{MLP} it ranges over \texttt{hidden\_layer\_sizes}, \texttt{alpha} and \texttt{learning\_rate\_init}. Grid search evaluates a $12$-point grid and random search draws a matched budget of $12$ candidates from broader ranges, so that the two differ in search strategy and not in compute. Each candidate is scored by $3$-fold CV on the training portion of the split, and the selected configuration is refit on the full training portion. No test observation is used at any point of the selection. The fixed-hyper-parameter variant uses a single reference configuration held constant across splits: the one-hidden-layer \texttt{MLP} ($32$ units) and the $200$-tree \texttt{ExtraTrees} of the main experiments. The searches over \texttt{ExtraTrees} also keep $200$ trees. Candidates are scored by \texttt{scikit-learn}'s default score ($R^2$) within the inner CV, while the gains, as in the main text, are measured on the mean squared error.

\begin{figure}[htbp]
    \centering
    \includegraphics[width=0.55\linewidth]{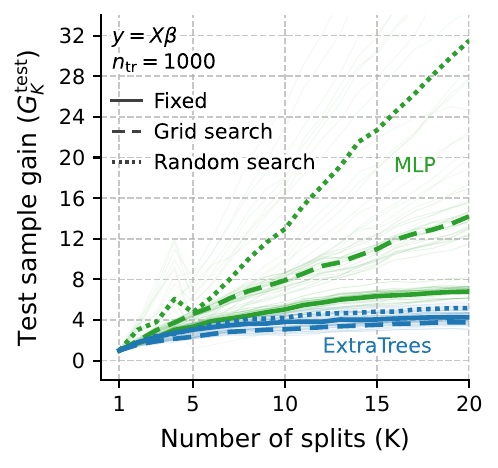}
    \caption{Test sample gain $G^{\text{test}}_K$ on the mean squared error on the simulated linear process at $n_{\text{tr}}=1{,}000$, for \texttt{MLP} and \texttt{ExtraTrees} with fixed hyper-parameters, grid search and random search. Colour: algorithm. Line style: tuning procedure. Thick lines: estimate on all seeds. Thin lines: $20$ bootstrap resamples of the seeds ($95\%$ intervals in \autoref{tab:ci-figI1}). The random-search \texttt{MLP} curve is erratic at small $K$ because of a few poorly fitting configurations. Random search gives the largest point estimate of the gain for both algorithms, for \texttt{ExtraTrees} above the $95\%$ intervals of the other procedures.}
    \label{fig:hpo-gains}
\end{figure}

\paragraph{Result} The sample-gain curves of the tuned procedures follow the same increasing trend as their fixed-hyper-parameter counterparts. \autoref{tab:hpo_gains} reports the values at $K=20$ for the setting of \autoref{fig:all_data_test_sample_gains} (left), that is simulated linear data at $n_{\text{tr}} = 1{,}000$. \autoref{fig:hpo-gains} shows the full curves. Random search occasionally selects a poorly fitting configuration: $51$ of the $20{,}000$ \texttt{MLP} fits, spread over $45$ seeds, have a test error more than ten times the median. These fits make the single-split errors heavy-tailed, and $7$ of them fall on the first split, against $2.5$ expected: they inflate the single-split variance, so that the random-search \texttt{MLP} gain, {$31.5$}, exceeds $K = 20$ and is the least precise of \autoref{tab:hpo_gains} ($95\%$ of the bootstrap resamples between {$10.4$ and $61.5$}). With the single-split variance averaged over the $20$ splits ({Appendix}~\ref{sec:sample-gain-estimation}), it is $16.1$, {close to the $14.2$ of grid search}.

\begin{table}[htbp]
    \centering
    \caption{Variance-equivalent test sample gains $G^{\text{test}}_{20}$ on the mean squared error on simulated linear data at $n_{\text{tr}} = 1{,}000$, for procedures with and without internal hyper-parameter search{} (estimates on all seeds; intervals in \autoref{tab:ci-figI1}).}
    \label{tab:hpo_gains}
    \begin{tabular}{@{} l c c c @{}}
        \toprule
        \thead{Algorithm} & \thead{Fixed\\parameters} & \thead{Grid\\search} & \thead{Random\\search} \\
        \midrule
        \texttt{MLP}         & {6.8} & {14.2} & {31.5} \\
        \texttt{ExtraTrees}  & {4.3} & {3.8} & {5.2}  \\
        \bottomrule
    \end{tabular}
\end{table}

\paragraph{Interpretation through the redundancy mechanism} For \texttt{ExtraTrees}, grid search selects nearly the same configuration on every split, so the split-specific predictors resemble one another and their fold errors remain correlated: the inter-fold covariance stays close to the fixed-hyper-parameter case. For the \texttt{MLP}, grid search also moves away from the fixed architecture and varies it across splits. Random search draws a different configuration on each split, yielding more diverse predictors whose errors on shared observations are less correlated, and therefore a larger gain. Computing the redundancy $\widehat{\omega}^{\mathrm{study}}_{k,s}$ on the squared error (after three splits, averaged over seeds) gives $0.065$ for the fixed \texttt{MLP}, $0.028$ for grid search and $0.023$ for random search, and $0.18$, $0.18$ and $0.15$ for \texttt{ExtraTrees}, with the same ordering after five splits:\[
  \text{FixedParameters} \;\geq \; \text{GridSearch} \;>\; \text{RandomSearch},
\]
the reverse of the ordering of the observed gains, which is what the triage rule of {Appendix}~\ref{sec:study-only-redundancy-statistics} predicts: high redundancy indicates an early-settling gain. For \texttt{ExtraTrees}, fixed parameters and grid search share the same redundancy and reach similar gains ({$4.3$ and $3.8$}).{} The redundancy has no parameter fitted to tuning procedures, so {that it orders them as their gains is consistent with} it capturing the intended mechanism.

Tuning therefore does not weaken the case for additional splits. If anything it strengthens it, since a stochastic search increases the diversity of the split-specific predictors and hence the value of averaging over more of them.

\FloatBarrier
\section{Repeated training runs versus resampled splits}
\label{appendix:fixed-vs-reshuffled}

This appendix measures {the two schemes of \autoref{sec:sample_gain}, repeated training runs on a fixed split and resampled splits,} against each other directly, at matched computational cost.

\paragraph{Protocol} On the simulated regression process of \autoref{appendix:experimental_details}, we redraw the data at each of $1{,}000$ seeds. Within each seed the same number of fits, $20$, is performed under two schemes:
\begin{itemize}
    \item \emph{reshuffled}: $20$ MCCV splits, each with its own train/test partition.
    \item \emph{frozen}: a single train/test partition, on which $20$ training runs are performed with different internal random states.
\end{itemize}
The frozen scheme is the common practice of averaging several training runs to absorb algorithmic randomness. Because both arms perform $20$ fits, the comparison isolates the effect of resampling the split rather than the effect of spending more compute.

\begin{figure}[htbp]
    \centering
    \includegraphics[scale=0.82]{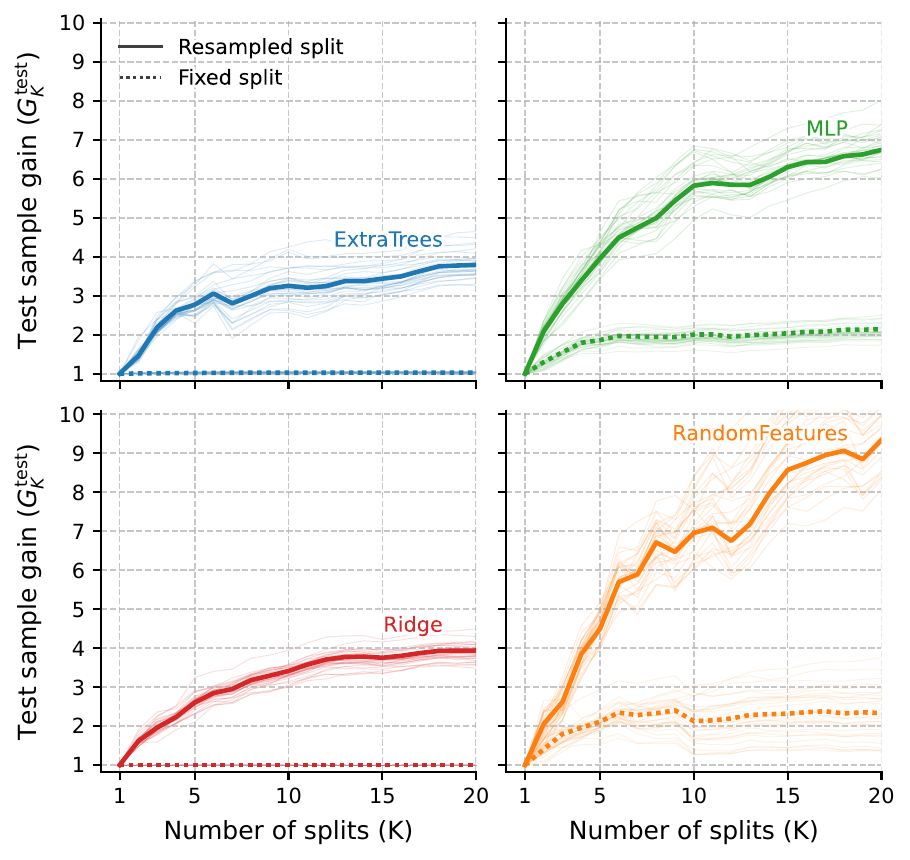}
    \caption{Test sample gain $G^{\text{test}}_K$ on the mean squared error on the simulated linear process at $n_{\text{tr}}=100$, for reshuffled MCCV (solid) and a frozen split repeated $K$ times (dotted). Thick lines: estimate on all seeds. Thin lines: $30$ bootstrap resamples of the seeds, in a lighter shade for the frozen split ($95\%$ intervals in \autoref{tab:ci-figJ1}).}
    \label{fig:fixed-vs-resampled}
\end{figure}

\paragraph{Result} At training size $n_{\text{tr}} = 100$, reshuffling reduces the standard deviation across seeds of the benchmark-adjusted score, averaged over the $20$ fits and measured on the mean squared error, by a factor of $1.7$ to $2.0$ for every algorithm: $2.0$ for \texttt{RandomFeatures}, $1.9$ for \texttt{Ridge} and \texttt{ExtraTrees}, $1.7$ for the \texttt{MLP} (and for \texttt{GradientBoosting}, run with the same design).{} This factor is a \emph{standard-deviation ratio} between two protocols at a fixed number of fits, and should not be confused with the variance-equivalent sample gain $G^{\text{test}}_K$ used elsewhere in the paper. The two are nonetheless linked: the ratio is of the order of $\bigl(G^{\text{test}}_{20,\text{reshuffled}}/G^{\text{test}}_{20,\text{frozen}}\bigr)^{1/2}$. For \texttt{Ridge}, whose $20$ fits on a frozen split are identical, it is the square root of its reshuffled gain ($\sqrt{{3.9}} \approx 2.0$), and with $20$ fits it cannot exceed about $\sqrt{20} \approx 4.5$ unless the split errors are negatively correlated.{} The residual variance of the frozen arm is the shared part $(\mathrm{A}_1)$, the error that any fit makes on that particular partition, which no number of additional fits removes{} (\autoref{fig:fixed-vs-resampled}).

The ratio varies little across algorithms, whether or not their fits vary on a fixed split. In the decomposition \eqref{eq:evaluation-noise-split}, resampling acts on the shared evaluation noise $(\mathrm{A}_1)$, which every algorithm incurs, whereas what repeating a frozen split averages out, the fit-specific part $(\mathrm{A}_2)$, differs from one algorithm to the next. \autoref{fig:fixed-vs-resampled} shows the two effects separately.

The two randomised learners of \autoref{fig:fixed-vs-resampled} make this decomposition visible within a single figure. \texttt{RandomFeatures} draws new random features at every fit. With $50$ features for $100$ training observations and a light penalty, two fits on the same split get different test observations wrong, and the fit-specific part $(\mathrm{A}_2)$ makes up $65\%$ of its evaluation noise, against $55\%$ for the \texttt{MLP}, $1\%$ for \texttt{ExtraTrees}, whose $200$ trees average out their own randomisation within a single fit, and $0$ for the deterministic \texttt{Ridge} (within each seed the $20$ fits share the split, so the spread of their errors estimates $(\mathrm{A}_2)$). Repeating its frozen split therefore already yields $G^{\text{test}}_{20} = {2.3}$, against {$2.1$} for the \texttt{MLP} and $1.0$ for \texttt{Ridge} and \texttt{ExtraTrees}, in line with the gain $K/\bigl(1+(K-1)\,\rho_{\mathrm{fix}}\bigr)$ of a repeated fixed split derived in \autoref{appendix:sample_gain_justification} ($2.6$, $2.1$, $1.0$ and $1.0$). Resampling the split still raises the gain of \texttt{RandomFeatures} to {$9.3$}, because only resampling acts on the shared part $(\mathrm{A}_1)$.

\FloatBarrier
\section{Comparison with repeated $5$-fold}
\label{appendix:repeated-kfold}

\begin{figure}[htbp]
    \centering
    \includegraphics[width=\linewidth]{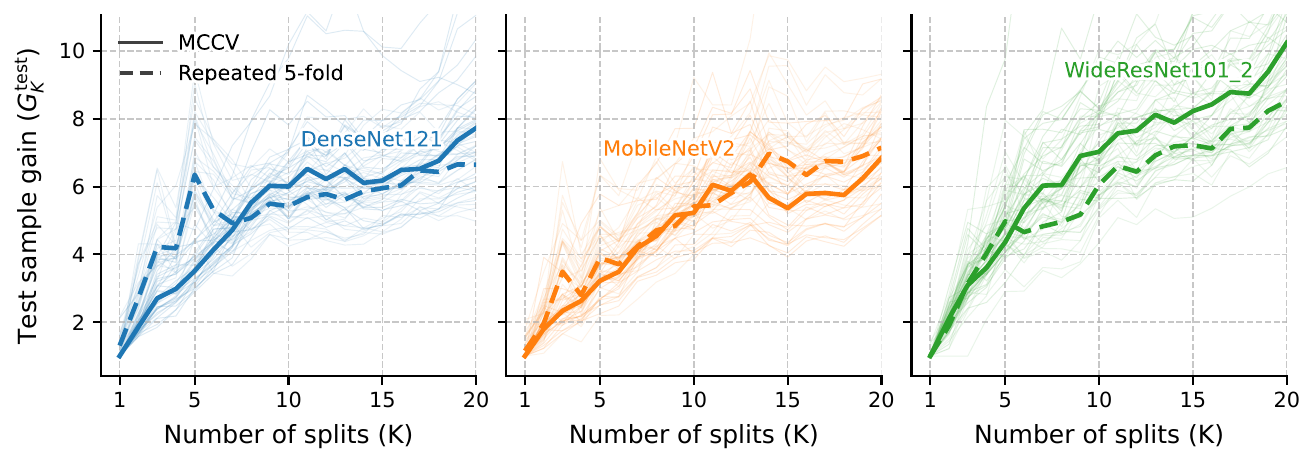}
    \caption{{Sample gains} on \texttt{PCam} {at $n_{\text{tr}}=240$ (study sets of $300$) under accuracy,} for MCCV (solid) and {repeated $5$-fold} (dashed). {Thick lines: estimate on all seeds. Thin lines: $30$ bootstrap resamples of the seeds ($95\%$ intervals in \autoref{tab:ci-figF1}). Both procedures are measured against the single-split variance of the MCCV runs, so at $K=1$ the repeated $5$-fold curves, from separate runs, need not equal $1$ ($1.31$ for \texttt{DenseNet121}, $1.15$ for \texttt{MobileNetV2}).}}
    \label{fig:pcam_mccv_vs_kfold}
\end{figure}

$K$-fold CV uses each sample of the study set exactly once as a test sample during the procedure. This CV procedure is the most commonly used one{, whereas our protocol uses MCCV (\autoref{sec:resampling}). This appendix checks that the sample gains of \autoref{sec:empirical-results} do not hinge on this choice}.

We performed the exact same experiments {on \texttt{PCam}} with {repeated $5$-fold} instead of MCCV. As can be observed on \autoref{fig:pcam_mccv_vs_kfold}, while {repeated $5$-fold gives the larger gain over the first $5$ splits for \texttt{DenseNet121} ({$6.3$ against $3.5$} at $K=5$), the two procedures end within each other's $95\%$ intervals at $K=20$ for every network (\autoref{tab:ci-figF1})}.

While $K$-fold CV does not perform the resampling step described in \autoref{sec:resampling}, MCCV {allows one to estimate dynamically} the redundancy defined in {Appendix}~\ref{sec:study-only-redundancy-statistics}. Its overlap factor $t$ is specific to MCCV: in repeated $K$-fold, splits of the same repeat have disjoint test sets, so the share of a test set that other splits also test differs.

\FloatBarrier
\section{Stratified resampling}
\label{appendix:stratification}

Stratifying the class distribution when splitting is a standard variance-reduction device \citep{kohavi_cv_1995}. In the protocol of \autoref{appendix:experimental_details}, the study set is drawn from the full dataset by stratified sampling, but the subsequent fold-level partition into training, validation and test subsets is not stratified. The reason is that both of our classification datasets, \texttt{PCam} and \texttt{Yelp Review Full}, have balanced classes, which limits what fold-level stratification can achieve{\ for models whose errors are spread evenly over the classes (\texttt{ViT-B/16} below is the exception)}. It is also worth noting that the sample gain is a \emph{within-protocol} ratio: it compares a single split to $K$ splits under the same resampling scheme, so stratifying both sides leaves the ratio well defined and the reported gains are not an artefact of an unstratified baseline.

We nonetheless re-ran the \texttt{PCam} experiments at $n_{\text{tr}}=400$ for the four networks ($100$ seeds){} with stratified splitting. {For the three CNNs, variances and sample gains behave closely to the unstratified case under all three metrics: the ratio of the unstratified to the stratified variance is $0.75$ to $1.06$ for the single-split score and $0.88$ to $1.33$ for the $20$-split average, every $95\%$ interval containing $1$ (\autoref{tab:strat-variance}). For them, the \texttt{PCam} results of \autoref{sec:empirical-results} are therefore robust to this choice of protocol. \texttt{ViT-B/16} is discussed below.} We did not re-run \texttt{Yelp Review Full} with stratified splitting.

\begin{figure}[htbp]
    \centering
    \includegraphics[scale=0.82]{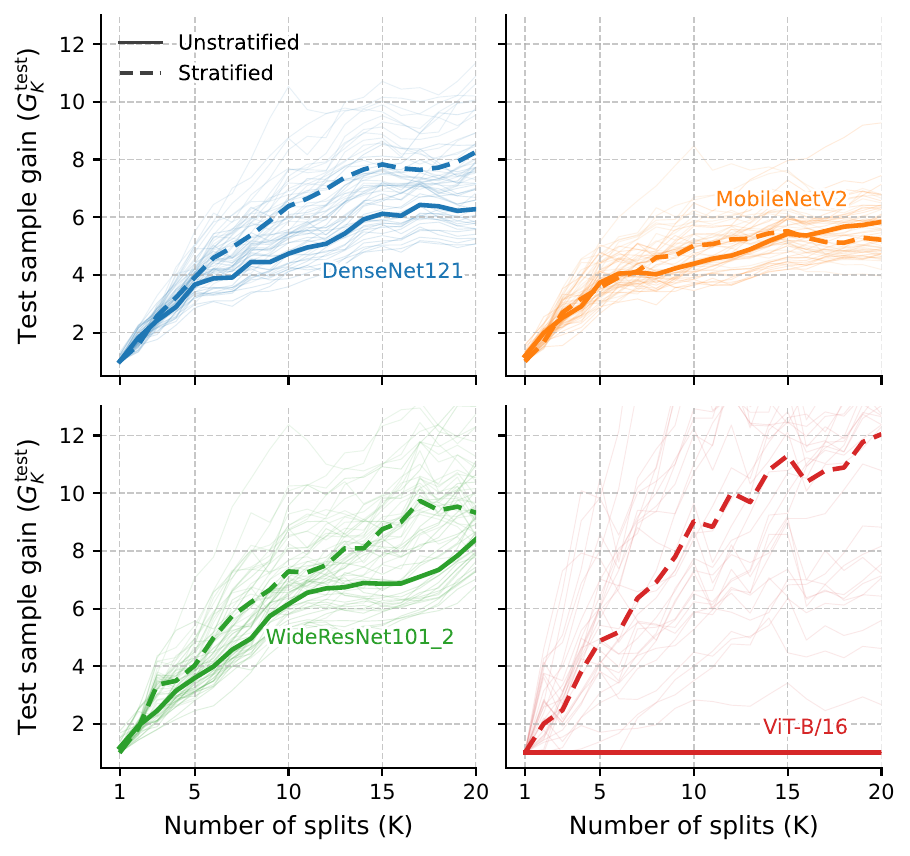}
    \caption{Test sample gain on \texttt{PCam} at {$n_{\text{tr}}=400$} under the negative log-likelihood, for unstratified (solid) and stratified (dashed) resampling. {Both are} measured against the same stratified single-split baseline, {so at $K=1$ the unstratified curves need not equal $1$ ($1.19$ for \texttt{MobileNetV2} and $1.18$ for \texttt{WideResNet101\_2}, within seed noise, \autoref{tab:strat-variance}). Thick lines: estimate on all seeds. Thin lines: $30$ bootstrap resamples of the seeds, in a lighter shade for stratified resampling ($95\%$ intervals in \autoref{tab:ci-figK1})}.}
    \label{fig:stratified-overlay}
\end{figure}

\paragraph{A case worth reporting} {\texttt{ViT-B/16} is the exception, under every metric. In our protocol it often learns little: its mean benchmarking accuracy is $0.58$ at $n_{\text{tr}}=400$, against $0.78$ to $0.82$ for the CNNs, about half of its fits stay below $0.6$, and its median benchmarking negative log-likelihood, $0.692$, is that of a constant prediction of $1/2$. Even when the population classes are balanced, a small random test set fluctuates in its realised class share, with a standard deviation of $0.05$ for a fold of $100$ images. The score of a near-constant predictor follows that share, whereas a model whose errors are spread evenly over the classes is barely affected. Stratification removes this component of the variance: for \texttt{ViT-B/16}, it divides the variance of the single-split score by $2.0$ under accuracy, $35$ under the Brier score and $48$ under the negative log-likelihood, and that of the $20$-split average by $1.5$, $14$ and $22$ (\autoref{tab:strat-variance}). Under the negative log-likelihood, unstratified $20$-split CV even remains noisier than a stratified single split (variance ratio $2.8$, $95\%$ interval $[1.1, 10.5]$), which is why its gain stays at $1$ against the stratified baseline of \autoref{fig:stratified-overlay}.}

\begin{table}[htbp]
    \centering
        \caption{Effect of stratified splitting on \texttt{PCam} ($n_{\text{tr}}=400$, $K=20$, $100$ paired seeds, $99$ for \texttt{MobileNetV2}): ratio of the unstratified to the stratified variance across seeds of the benchmark-adjusted score, for the single split and the $20$-split average, with $95\%$ bootstrap intervals over the seeds ($1{,}000$ draws). Ratios above $1$ mean that stratification lowers the variance.}
    \label{tab:strat-variance}
    \footnotesize
    \begin{tabular}{@{} llcccc @{}}
        \toprule
        \thead{Metric} & \thead{Score} & \texttt{DenseNet121} & \texttt{MobileNetV2} & \texttt{WideResNet101\_2} & \texttt{ViT-B/16} \\
        \midrule
        Accuracy & $1$ split & $1.02$ {\scriptsize $[0.67,\,1.62]$} & $0.80$ {\scriptsize $[0.57,\,1.14]$} & $0.86$ {\scriptsize $[0.55,\,1.34]$} & $1.99$ {\scriptsize $[1.32,\,3.10]$} \\
         & $20$ splits & $1.12$ {\scriptsize $[0.81,\,1.51]$} & $0.89$ {\scriptsize $[0.66,\,1.19]$} & $1.01$ {\scriptsize $[0.75,\,1.35]$} & $1.49$ {\scriptsize $[1.13,\,1.93]$} \\
        \midrule
        Neg.\ log-likelihood & $1$ split & $1.06$ {\scriptsize $[0.70,\,1.63]$} & $0.84$ {\scriptsize $[0.49,\,1.41]$} & $0.83$ {\scriptsize $[0.48,\,1.41]$} & $47.6$ {\scriptsize $[14.2,\,201]$} \\
         & $20$ splits & $1.33$ {\scriptsize $[0.97,\,1.78]$} & $0.88$ {\scriptsize $[0.65,\,1.18]$} & $1.11$ {\scriptsize $[0.84,\,1.42]$} & $22.2$ {\scriptsize $[11.5,\,41.0]$} \\
        \midrule
        Brier score & $1$ split & $0.95$ {\scriptsize $[0.59,\,1.42]$} & $0.75$ {\scriptsize $[0.52,\,1.09]$} & $0.76$ {\scriptsize $[0.50,\,1.13]$} & $34.8$ {\scriptsize $[11.5,\,137]$} \\
         & $20$ splits & $1.21$ {\scriptsize $[0.87,\,1.61]$} & $0.91$ {\scriptsize $[0.68,\,1.21]$} & $1.12$ {\scriptsize $[0.82,\,1.51]$} & $14.3$ {\scriptsize $[7.49,\,27.1]$} \\
        \bottomrule
    \end{tabular}
\end{table}

Under matched protocols the picture is unchanged: stratified CV keeps reducing variance as $K$ grows, with {sample gains at $K=20$ of $8.3$, $5.2$ and $9.3$ for the CNNs and $12.0$ for \texttt{ViT-B/16}} at {$n_{\text{tr}} = 400$} (\autoref{fig:stratified-overlay}{, \autoref{tab:ci-figK1}}).

The two devices are therefore complementary rather than competing. Stratification is essentially free{, did not detectably raise the variance in any of our runs, and protects the evaluation when a model's losses differ between classes. It} should be used whenever the task is classification, and repeated splitting continues to pay on top of it.

\FloatBarrier
\section{Sample gains across evaluation metrics}
\label{appendix:metrics}

{The sample gain can be computed on any test metric, and the redundancy on any metric that decomposes over observations. This appendix checks how much the metric moves the gain, explains why the main text reports gains on the mean squared error rather than $R^2$, and lists the metrics to which the redundancy applies.}

On the simulated regression process at $n_{\text{tr}} = 1{,}000$, the gain at $K=20$ for the \texttt{MLP} of the main experiments is {$6.8$} on the mean squared error and {$8.7$} on the mean absolute error{, and} for \texttt{RandomFeatures} it is {$8.6$ and $6.9$}{: the metric moves the gain, by $28\%$ at most here, but not its order of magnitude}.

$R^2${, the other usual regression score,} is not a suitable target for the redundancy, and the main text reports gains on the mean squared error. A split's $R^2$ divides its squared error by the variance of that split's own test targets, centred on their own mean, so it is not an average of per-sample losses. For a strongly regularised learner this normalisation dominates: {on a separate $1{,}000$-seed run of the same design}, \texttt{Ridge} ($\alpha = 1613$) has $G^{\text{test}}_{20} = {8.2}$ on $R^2$ against {$4.4$} on the mean squared error ($95\%$ intervals {$[7.3, 9.0]$ and $[3.9, 5.0]$}). In a separate $200$-seed run of the same design that records per-sample errors, recomputing each split's $R^2$ with the targets centred on the study mean instead of the split's own mean brings the \texttt{Ridge} gain from $7.5$ down to $3.2$, against $3.4$ on the mean squared error (on these runs, the gain is read as the ratio of the single-split variance, averaged over the splits, to the variance of the $20$-split average), while the gains of the other algorithms barely move. That part of the $R^2$ gain reflects the normalisation of the metric rather than the redundancy of the predictors, which is why the redundancy is computed, and the gain read, on the mean squared error.

\begin{figure}[htbp]
    \centering
    \includegraphics[width=0.5\linewidth]{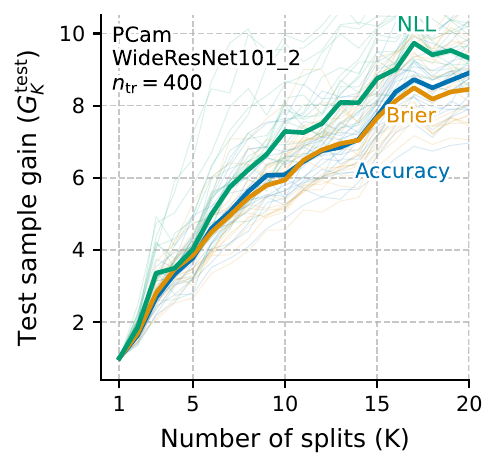}
    \caption{Test sample gain $G^{\text{test}}_K$ for \texttt{WideResNet101\_2} on \texttt{PCam} at {$n_{\text{tr}}=400$} (stratified resampling), under accuracy, the Brier score and the negative log-likelihood. {Thick lines: estimate on all seeds. Thin lines: $20$ bootstrap resamples of the seeds ($95\%$ intervals in \autoref{tab:ci-figL1}).}}
    \label{fig:metrics-gain}
\end{figure}

On \texttt{PCam}, re-running the experiments with accuracy, the Brier score and the negative log-likelihood produces sample-gain curves that are close to one another across the range of $K$ we consider (\autoref{fig:metrics-gain}). These \texttt{PCam} runs stratify the folds, which departs from the unstratified fold-level default of \autoref{appendix:experimental_details} but follows the recommendation of \autoref{appendix:stratification}. That appendix also shows that variances and sample gains stay close between the two protocols ({stratification matters only for \texttt{ViT-B/16}, under every metric}), so the comparison across metrics does not hinge on this choice.

\paragraph{Applicability} {The sample gain is defined for any metric evaluable on a test set, AUC-ROC included, whereas the redundancy needs the per-sample losses of the metric whose gain is targeted (\autoref{sec:empirical-results}). Accuracy, squared error, absolute error, the Brier score and the negative log-likelihood provide them. AUC-ROC, a rank statistic over \emph{pairs} of observations, and $R^2$ do not, so the early-stopping rule does not apply to them in its present form.}

\FloatBarrier
\section{Behaviour at small label budgets}
\label{appendix:label-scarce}

The regime this paper targets is one in which labels are expensive, so the behaviour of both the recommendation and the diagnostic at very small budgets deserves separate treatment.

\paragraph{Protocol} We sweep training budgets from $5$ to $100$ samples on the simulated linear process ($1{,}000$ seeds, $K=20$ splits) {for} \texttt{Ridge}, \texttt{ExtraTrees}, the one-hidden-layer \texttt{MLP} with $32$ units and \texttt{RandomFeatures} of the main experiments, computing at each budget the variance-equivalent test sample gain on the mean squared error and the redundancy on the squared error.

\begin{figure}[t]
    \centering
    \begin{subfigure}{0.49\linewidth}
        \centering
        \includegraphics[width=\linewidth]{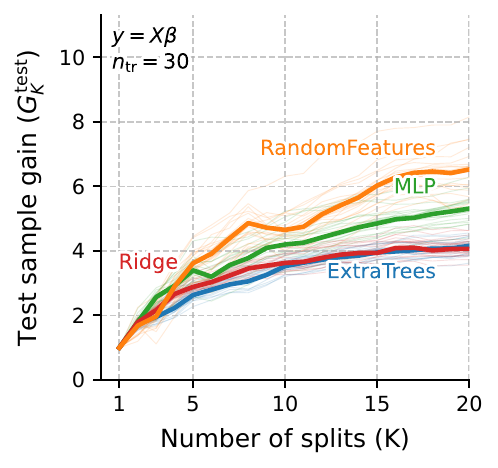}
    \end{subfigure}\hfill
    \begin{subfigure}{0.49\linewidth}
        \centering
        \includegraphics[width=\linewidth]{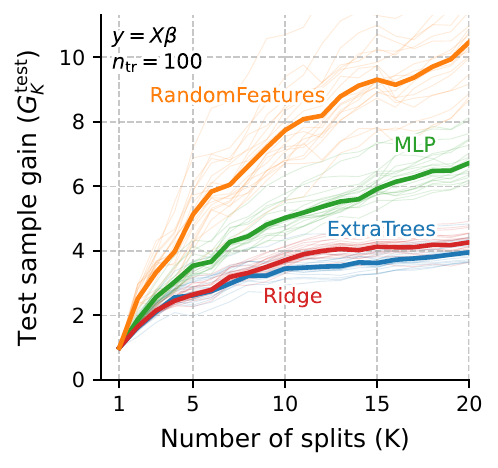}
    \end{subfigure}
    \caption{Test sample gain $G^{\text{test}}_K$ on the mean squared error{} on the simulated linear process, at two small training budgets (shared vertical axis). {Thick lines: estimate on all seeds. Thin lines: $20$ bootstrap resamples of the seeds ($95\%$ intervals in \autoref{tab:ci-figM1}).}}
    \label{fig:small-label-gain}
\end{figure}

\paragraph{Gains persist, but coupling grows as labels become scarcer} At every budget the gain is still rising at $K=20$: between $5$ and $20$ splits it increases by a factor of $1.1$ to $2.0$, and $20$ splits are worth a test set {$2.2$ to $10.5$} times larger than a single hold-out (\autoref{fig:small-label-gain}). The gain does not grow as the budget shrinks, however: it is smallest at $5$ to $10$ labels ($G^{\text{test}}_{20}$ between {$2.2$ and $4.0$}). The redundancy changes little with the budget (after five splits, about $0.05$ to $0.09$ for the \texttt{MLP} and $0.17$ to $0.20$ for \texttt{Ridge}), so the curve $K/\bigl(1+(K-1)\,\omega\bigr)$ of {Appendix}~\ref{sec:study-only-redundancy-statistics} predicts similar gains at every budget, and the measured gains fall further below it as labels become scarcer. Inverting the measured gain, the train/test coupling term averages {$0.15$ to $0.22$} across algorithms at $5$ to $10$ labels, against about {$0.02$ to $0.07$} from $25$ labels on. With so few observations, a sample tested by one split weighs heavily in the fits of the splits that train on it, so the coupling, intrinsic to cross-validation, carries a larger share of the covariance between splits. This is the shortfall at small training sizes already noted in {Appendix}~\ref{sec:study-only-redundancy-statistics}. Repeated resampling therefore remains worthwhile at every budget, but its benefit does not exceed that at larger budgets.

\paragraph{Where the diagnostic applies} The redundancy is estimated from the observations held out by two different splits (here on the squared error), so its reliability is tied to the size of the held-out fold. As a per-study signal it is noisy when only a handful of observations are held out: in our experiments it becomes reliable for clearly separated algorithms above a budget of roughly $20$ labels, corresponding to held-out folds of four to five observations. At $20$ labels, the redundancy is defined (some pair of splits shares at least two held-out observations) in about $60\%$ of single runs after three splits and $95\%$ after five, and a single run orders the two algorithms whose mean redundancies are furthest apart (\texttt{Ridge} and the \texttt{MLP}) as their means do in $91\%$ of runs after three splits and $97\%$ after five. At $10$ labels it is defined in only a third of runs after three splits.{} At the very smallest budgets, where a fold contains one or two observations, the fold statistics are degenerate for cross-validation itself, independently of the redundancy.

\paragraph{On real data} {The only real-data budget in this range is \texttt{PCam} at $n_{\text{tr}}=24$, the smallest size of \autoref{tab:xp_splitting}: study sets of $30$ images, test folds of $6$, the four networks, $50$ seeds and $K=20$, from runs that also record the negative log-likelihood and the Brier score. It behaves like the simulated budgets from $25$ labels on. Under the three metrics, the gain is still rising at $K=20$, by a factor of {$1.5$ to $4.1$} between $K=5$ and $K=20$, and reaches $G^{\text{test}}_{20}$ between {$8.1$ and $20.6$}. Under accuracy it is at least as large as at $n_{\text{tr}}=400$ (\autoref{fig:all_data_test_sample_gains}, middle). The redundancy over the $20$ splits, $0.03$ to $0.09$, gives through $K/\bigl(1+(K-1)\,\omega\bigr)$ gains of $7.6$ to $12.6$, and the implied coupling term lies between {$-0.09$ and $0.04$}. With folds of $6$ observations, the redundancy is defined in $73\%$ of runs after three splits and $99\%$ after five. These runs record the redundancy only over all $20$ splits, so the single-run ordering of the networks cannot be checked. \texttt{ViT-B/16} is at chance at this budget (benchmarking accuracy $0.51$), and with $50$ seeds the $95\%$ intervals are wide (for \texttt{DenseNet121} under accuracy, $G^{\text{test}}_{20}$ lies in {$[8.1, 17.8]$}).}

\paragraph{The two together cover the range} Below about $20$ labels the redundancy is often undefined or unreliable, but the recommendation does not depend on it: every algorithm still gains from additional splits{, by a factor of $1.1$ to $2.0$ between $K=5$ and $K=20$}. From about $20$ labels on, where algorithms begin to separate (the \texttt{MLP} and \texttt{RandomFeatures} reach $G^{\text{test}}_{20}$ between {$5.3$ and $10.5$}, against $3.1$ to {$4.3$} for \texttt{Ridge} and \texttt{ExtraTrees}), the held-out folds are large enough for the redundancy to tell them apart.

\FloatBarrier
\section{Statistical significance of pairwise rankings on real data}
\label{appendix:statistical_significance}

{This appendix details the significance tests of the quasi-oracle score differences used in \autoref{sec:pairwise-ranking}, and the retrieval proportions of \autoref{tab:stat_significance}. Like that section, it bears on comparing learning algorithms, a question complementary to the precision of one algorithm's performance estimate.}

\subsection{Formalization and methodology}
\label{appendix:sig-formalization}
To rigorously compare the performance of two learning algorithms $\mathcal{A}$ and $\mathcal{B}$, we employ a \textit{paired $t$-test} across $S$ independent seeds{{} ($S=50$)}. This approach is necessary because for each seed, both algorithms are trained and evaluated on identical data splits, meaning the resulting scores are not independent.

Let $R_s^{(\mathcal{A})}$ and $R_s^{(\mathcal{B})}$ represent the benchmarking scores of the two algorithms for seed $s \in \{1, \dots, S\}$. We define the performance difference as:
\begin{equation}
D_s = R_s^{(\mathcal{A})} - R_s^{(\mathcal{B})}
\end{equation}

Our null hypothesis ($H_0$) and alternative hypothesis ($H_1$) are:
\begin{itemize}
    \item $H_0: \mu_D = 0$ (There is no significant difference in performance).
    \item $H_1: \mu_D \neq 0$ (The performance difference is statistically significant).
\end{itemize}

\begin{table}[t!]
\centering
\caption{Paired t-test results on real datasets comparing mean benchmarking differences over {$S=50$} paired seeds. Significant results ($p < 0.05$) are highlighted. The last two columns report the proportion of paired runs retrieving the displayed oracle direction with a single split and with $K=20$ {MCCV}. {$n_{\text{tr}}$ counts the training and validation subsets together. For both datasets, every column uses the $50$ seeds of the $20$-split MCCV runs, whose first split is the single split, and the $t$-test compares per-seed means of the benchmarking-set scores over the $20$ splits.}}
\label{tab:stat_significance}
\scriptsize
\setlength{\tabcolsep}{2.2pt}
\renewcommand{\arraystretch}{0.92}
\makebox[\textwidth][c]{%
\resizebox{0.98\textwidth}{!}{%
\begin{tabular}{@{}llccccccc@{}}
\toprule
\thead{Highest\\mean} & \thead{Lowest\\mean} & \thead{Training\\size ($n_{\text{tr}}$)} & \thead{Mean diff\\($\bar{D}$)} & \thead{Std dev\\($s_D$)} & \thead{$t$-stat} & \thead{$p$-value} & \thead{Single-split\\retrieval} & \thead{$K=20$ CV\\retrieval} \\
\midrule
\multicolumn{9}{c}{\textbf{PatchCamelyon}}\\
\midrule
{DenseNet121} & {MobileNetV2} & {24} & {0.02569} & {0.03083} & {5.891} & {\textbf{$<$ 0.0001}} & {38\%} & {56\%} \\
{DenseNet121} & {WideResNet101\_2} & {24} & {0.04415} & {0.03033} & {10.291} & {\textbf{$<$ 0.0001}} & {44\%} & {74\%} \\
{MobileNetV2} & {WideResNet101\_2} & {24} & {0.01846} & {0.03046} & {4.285} & {\textbf{$<$ 0.0001}} & {50\%} & {56\%} \\
\midrule
{DenseNet121} & {MobileNetV2} & {80} & {0.00807} & {0.02337} & {2.441} & {\textbf{0.0183}} & {56\%} & {60\%} \\
{DenseNet121} & {WideResNet101\_2} & {80} & {0.06697} & {0.01842} & {25.701} & {\textbf{$<$ 0.0001}} & {70\%} & {100\%} \\
{MobileNetV2} & {WideResNet101\_2} & {80} & {0.05890} & {0.02227} & {18.698} & {\textbf{$<$ 0.0001}} & {64\%} & {96\%} \\
\midrule
{MobileNetV2} & {DenseNet121} & {240} & {0.00904} & {0.00715} & {8.942} & {\textbf{$<$ 0.0001}} & {46\%} & {82\%} \\
{DenseNet121} & {WideResNet101\_2} & {240} & {0.02522} & {0.00878} & {20.308} & {\textbf{$<$ 0.0001}} & {62\%} & {96\%} \\
{MobileNetV2} & {WideResNet101\_2} & {240} & {0.03425} & {0.00777} & {31.160} & {\textbf{$<$ 0.0001}} & {72\%} & {100\%} \\
\midrule
{MobileNetV2} & {DenseNet121} & {400} & {0.00340} & {0.00604} & {3.988} & {\textbf{0.0002}} & {46\%} & {70\%} \\
{DenseNet121} & {WideResNet101\_2} & {400} & {0.03570} & {0.00828} & {30.476} & {\textbf{$<$ 0.0001}} & {72\%} & {100\%} \\
{MobileNetV2} & {WideResNet101\_2} & {400} & {0.03911} & {0.00667} & {41.451} & {\textbf{$<$ 0.0001}} & {78\%} & {100\%} \\
\midrule
{MobileNetV2} & {DenseNet121} & {640} & {0.00062} & {0.00398} & {1.111} & {0.2721} & {46\%} & {54\%} \\
{DenseNet121} & {WideResNet101\_2} & {640} & {0.03950} & {0.00730} & {38.280} & {\textbf{$<$ 0.0001}} & {76\%} & {100\%} \\
{MobileNetV2} & {WideResNet101\_2} & {640} & {0.04012} & {0.00690} & {41.088} & {\textbf{$<$ 0.0001}} & {74\%} & {100\%} \\
\bottomrule
\multicolumn{9}{c}{\textbf{Yelp Review Full}}\\
\toprule
BERT & XLM-RoBERTa & 1,000 & {0.02888} & {0.03195} & {6.392} & {\textbf{$<$ 0.0001}} & {68\%} & {78\%} \\
\midrule
BERT & XLM-RoBERTa & 3,000 & {0.00561} & {0.01850} & {2.146} & {\textbf{0.0369}} & {24\%} & {56\%} \\
\bottomrule
\end{tabular}%
}%
}
\end{table}

\subsection{Test statistic}
\label{appendix:sig-estimators}
The mean difference $\bar{D}$ and the sample standard deviation $s_D$ are computed from the $S$ experimental runs:
\begin{equation}
\bar{D} = \frac{1}{S} \sum_{s=1}^S D_s, \quad s_D = \sqrt{\frac{1}{S-1} \sum_{s=1}^S (D_s - \bar{D})^2}
\end{equation}

Under the Central Limit Theorem, the test statistic $t$ follows a Student's t-distribution with $df = S-1$ degrees of freedom:
\begin{equation}
t = \frac{\bar{D}}{s_D / \sqrt{S}}
\end{equation}

We evaluate significance at the $0.05$ level. For $df=99$ {(\texttt{PCam})}, the two-tailed critical value is $t_{crit}=1.9842$. A result is considered statistically significant if $|t| > t_{crit}$, or equivalently, if the $p$-value $< 0.05$.

\subsection{Empirical results}
\label{appendix:sig-results}

\subsubsection{Medical imaging}

\autoref{tab:stat_significance} reveals a stable separation between \texttt{WideResNet101\_2} and the two other CNN architectures: {both \texttt{DenseNet121} and \texttt{MobileNetV2} significantly outperform it at every training size. At the two smallest sizes, $n_{\text{tr}}=24$ and $80$, \texttt{DenseNet121} significantly outperforms \texttt{MobileNetV2}. From $n_{\text{tr}}=240$ onward, \texttt{MobileNetV2} has the highest mean benchmarking score. Its advantage over \texttt{DenseNet121} is significant at $n_{\text{tr}}=240$ and $400$, but no longer at $n_{\text{tr}}=640$.} This suggests that \texttt{DenseNet121} is preferable in the lowest-data regime, whereas \texttt{MobileNetV2} and \texttt{DenseNet121} become statistically close at larger training sizes.

It is important to note the role of the benchmarking set size ({$n_{\mathrm{bench}}$} $= 244{,}912$). In a binomial classification task, the variance of the accuracy estimator attributed to the test set size is bounded by $\frac{1}{4 n_{\mathrm{bench}}}$. Given our large {$n_{\mathrm{bench}}$}, this bound of the variance is $\approx 1.02 \times 10^{-6}$, corresponding to a standard deviation of at most $\approx 0.00101$, which is {about four times smaller than the smallest observed sample standard deviation $s_D$ ($0.0040$; the largest is $0.0308$)}.

Consequently, the benchmarking size {$n_{\mathrm{bench}}$} ensures that the standard deviation $s_D$ reported in \autoref{tab:stat_significance} is mostly measuring the algorithms' sensitivity to the training data composition and split seed, rather than noise from the finite benchmarking set.

\subsubsection{NLP fine-tuning}

Here, we apply the paired $t$-test methodology to evaluate the performance differences between \texttt{BERT} and \texttt{XLM-RoBERTa} architectures. For this dataset, models were evaluated on a benchmarking set of {$600{,}000$ reviews. Its} extremely large size ensures that the standard error of the evaluation metric is negligible, meaning the observed variance $s_D$ is a direct reflection of model instability across training subsets. {Every split of the $50$ seeds is scored on the benchmarking set: the per-seed score of a network is the mean over its $20$ models, and its quasi-oracle score the mean over all its $1{,}000$ models.}
\autoref{tab:stat_significance} summarizes the findings for {two} training scales. Notably, as the training size $n_{\text{tr}}$ increases, the mean difference $\bar{D}$ {shrinks much faster than the standard deviation $s_D$}, suggesting that both models exhibit significant sensitivity to the training seed in this specific task.

The analysis indicates that \texttt{BERT} is significantly superior to \texttt{XLM-RoBERTa} at the smallest experimented training scale ($n_{\text{tr}}=1{,}000$, $p < 0.0001$). {This advantage shrinks as more data is provided: at $n_{\text{tr}}=3{,}000$, the mean difference is five times smaller, although still significant ($p=0.0369$). This suggests that \texttt{XLM-RoBERTa}'s performance catches up with \texttt{BERT}'s as the sample size increases. Its failed runs (\autoref{fig:nlp_ranking_violin}) keep its mean below \texttt{BERT}'s, although it wins most individual splits.}

{The retrieval proportions of \autoref{fig:nlp_ranking_retrieval} are shares of $50$ seeds, given with {pointwise} $95\%$ Wilson score intervals{, a tie (equal test accuracy) counting as a failure}. Different split counts are compared on the same seeds, whose $K$-split runs are nested and start with the single split, so the comparison is paired: an exact McNemar test on the seeds where only one of the two split counts retrieves the quasi-oracle ranking gives, at $n_{\text{tr}}=3{,}000$, $p=0.0025$ between $K=20$ and the single split ($56\%$ against $24\%$, {with} $21$ such seeds {favouring} $K=20$ and $5$ the single split) and $p=0.019$ between $K=20$ and $K=5$ ($56\%$ against $30\%$, {with} $20$ against $7$), although the intervals of $K=5$ and $K=20$ overlap. At $n_{\text{tr}}=1{,}000$, where both shares are high, $K=20$ and the single split do not differ significantly ($78\%$ against $68\%$, $p=0.38$).}

\FloatBarrier
 \section{Significance tests {and racing} within a single cross-validation run}
\label{appendix:significance-power}

This paper targets the precision with which cross-validation estimates the performance of \emph{one} learning algorithm. Deciding whether one algorithm outperforms another is a complementary question, secondary in this paper (\autoref{sec:pairwise-ranking}). We follow \citet{bouckaert_evaluating_2004}, who showed that the number of repetitions entering a performance estimate affects the behaviour of significance tests comparing learning algorithms. \autoref{appendix:statistical_significance} compares algorithms across independent seeds against the benchmarking set. This appendix asks how a decision taken from a single CV run behaves as the number of splits $K$ grows{, and closes with racing, which takes such decisions while cross-validation proceeds}.

\paragraph{Protocol} For each pair of the three \texttt{PCam} CNNs we perform the corrected resampled $t$-test of \citet{nadeau_inference_2003}, whose variance result already underlies {Appendix}~\ref{appendix:cv-variance}, on the $K$ paired fold scores of a single run. The test is two-sided, at level $0.05$, with $K-1$ degrees of freedom and variance factor $1/K + n_{\text{te}}/n_{\text{tr}}$, where $n_{\text{te}}/n_{\text{tr}} = 0.25$. The $K$-split test uses the first $K$ splits of a $20$-split run, for $K = 2, \dots, 20$. The reference decision is {the paired $t$-test of \autoref{tab:stat_significance}:} two-sided, at the same level, across $50$ seeds, on the benchmarking-set scores of the two algorithms, each averaged over the {$20$} folds of a CV run. We report the proportion of runs in which the $K$-split test rejects in the direction of the reference, that is its empirical power against the estimated oracle. Where the reference is not significant, we report the plain rejection rate. This is related to, but distinct from, the replicability measure of \citet{bouckaert_evaluating_2004}, which concerns agreement between two independent runs of the same test rather than agreement with a reference direction. As a calibration check, we also apply the test to each algorithm against itself, comparing splits $1, \dots, K$ with splits $K+1, \dots, 2K$ of the same run, for $K \le 10$.

\begin{table}[htbp]
    \centering
    \caption{Rejection rate of the corrected resampled $t$-test within a single CV run on \texttt{PCam}, over $50$ seeds, at $K = 2$, $5$, $10$ and $20$ splits. D, M and W denote \texttt{DenseNet121}, \texttt{MobileNetV2} and \texttt{WideResNet101\_2}. $n_{\text{tr}}$ is the per-fold training size (training and validation subsets together), and the study set has $1.25\,n_{\text{tr}}$ observations. {The five study sets, of $30$ to $800$ observations, are those of \autoref{tab:stat_significance}.} \emph{Reference} is the benchmarking-set decision. Where it is significant, entries count rejections in its direction (power). Where it is not (n.s.), they are plain rejection rates. $^{\ast}$Runs that reject in the opposite direction are not counted. With $50$ seeds, a rate near $0.3$ carries a binomial $95\%$ interval of about $\pm 0.13$.}
    \label{tab:sig-power-grid}
    \small
    \begin{tabular}{@{} l r c c c c c @{}}
        \toprule
        Pair & $n_{\text{tr}}$ & Reference & $K=2$ & $K=5$ & $K=10$ & $K=20$ \\
        \midrule
        \texttt{DenseNet121} vs \texttt{MobileNetV2} & 24  & D $>$ M & 0.02\rlap{$^{\ast}$} & 0.02 & 0.00 & 0.00 \\
                                                     & 80  & D $>$ M & 0.06\rlap{$^{\ast}$} & 0.02 & 0.00 & 0.00 \\
                                                     & 240 & M $>$ D & 0.00 & 0.02 & 0.00 & 0.00 \\
                                                     & 400 & M $>$ D & 0.02 & 0.00 & 0.00 & 0.00 \\
                                                     & 640 & n.s.    & 0.00 & 0.02 & 0.00 & 0.00 \\
        \midrule
        \texttt{DenseNet121} vs \texttt{WideResNet101\_2} & 24  & D $>$ W & 0.08 & 0.04 & 0.00 & 0.00 \\
                                                          & 80  & D $>$ W & 0.00 & 0.04 & 0.04 & 0.06 \\
                                                          & 240 & D $>$ W & 0.02 & 0.02 & 0.00 & 0.00 \\
                                                          & 400 & D $>$ W & 0.02 & 0.08 & 0.04 & 0.10 \\
                                                          & 640 & D $>$ W & 0.00 & 0.16 & 0.24 & 0.32 \\
        \midrule
        \texttt{MobileNetV2} vs \texttt{WideResNet101\_2} & 24  & M $>$ W & 0.00\rlap{$^{\ast}$} & 0.00 & 0.00 & 0.00 \\
                                                          & 80  & M $>$ W & 0.00 & 0.00 & 0.04 & 0.00 \\
                                                          & 240 & M $>$ W & 0.00 & 0.04 & 0.06 & 0.00 \\
                                                          & 400 & M $>$ W & 0.00 & 0.10 & 0.10 & 0.10 \\
                                                          & 640 & M $>$ W & 0.04 & 0.28 & 0.24 & 0.26 \\
        \bottomrule
    \end{tabular}
\end{table}

\paragraph{A weak instrument at these sizes} \autoref{tab:sig-power-grid} reports the whole grid. Of the fourteen configurations in which the reference is significant, nine have power $0$ at $K=20$, and eleven never exceed $0.10$ at any $K$ between $2$ and $20${\ (\texttt{MobileNetV2} versus \texttt{WideResNet101\_2} at $n_{\text{tr}}=400$ reaches $0.14$ at $K=9$, a split count the table does not show)}. Only the two comparisons against \texttt{WideResNet101\_2} at $n_{\text{tr}}=640$ reach about $0.3$. Where power rises, it does so up to $K$ of about $5$ to $10$ and then fluctuates within seed noise: for \texttt{DenseNet121} versus \texttt{WideResNet101\_2} at $n_{\text{tr}}=640$, it reads $0.28$ at $K=8$, $0.24$ at $K=10$, $0.36$ at $K=19$ and $0.32$ at $K=20$. Since the tests at different $K$ share their first splits, these values are correlated across $K$, and differences between neighbouring values should not be read as trends. None of this contradicts \autoref{sec:pairwise-ranking}: at the two largest sizes, $20$-split CV places \texttt{WideResNet101\_2} below both other networks in every run (\autoref{tab:stat_significance}). A single CV run gets the direction right far more often than its own test can certify it.

\paragraph{Calibrated but conservative} The low power is not a calibration failure. In the self-comparison, the test rejects in at most $8\%$ of runs at $K=2$ and at most $6\%$ for $3 \le K \le 10$, with averages over algorithms and sizes below $3\%$, consistent with the nominal level of $5\%$. The power is lost to the correction itself: its variance factor $1/K + n_{\text{te}}/n_{\text{tr}}$ never falls below $n_{\text{te}}/n_{\text{tr}}$, whatever $K$. As $K$ grows, the test statistic therefore tends to $\bar d / (s_d \sqrt{n_{\text{te}}/n_{\text{tr}}})$, where $\bar d$ and $s_d$ are the mean and standard deviation of the fold-score differences, instead of growing like $\sqrt{K}$. Power levels off at a value set by the study set, not by the number of splits.

More splits do not rescue significance testing in this small-sample regime. This bears on comparisons only, and leaves the paper's main object, the precision of one algorithm's performance estimate, unaffected.

\paragraph{{Racing: comparisons that stop cross-validation early}} {The tests above compare two algorithms once, after a fixed number of splits. Racing \citep{maron_hoeffding_1994, moore_efficient_1994} instead compares candidates while cross-validation proceeds, and stops evaluating a candidate as soon as it is almost certainly worse than a rival. Like these tests, it bears on comparisons, not on the precision of one algorithm's performance estimate. It shares with the redundancy (\autoref{sec:empirical-results} and {Appendix}~\ref{sec:study-only-redundancy-statistics})} the objective of stopping cross-validation early, but not the trigger: it stops when a \emph{comparison between candidates} is statistically settled, and therefore needs at least two candidates, whereas {the redundancy} $\widehat{\omega}^{\mathrm{study}}_k$ is defined for a single algorithm and asks whether the \emph{estimator itself} has stopped improving. The statistical regimes differ too: racing accumulates leave-one-out errors on fresh observations, treated as independent, while our setting is governed by the dependence induced by reusing the same study observations across splits, the very correlation that gives \eqref{eq:delta_cvvar} its strictly positive floor.

\FloatBarrier
\section{Varying training size: ranking-equivalent sample gain}
\label{appendix:ranking-sample-gain}

The ranking-equivalent sample gain compares the study sizes at which single-split and $K$-split CV recover the oracle ranking, so it needs an oracle ranking that is the same at every training size. Whether it is the same depends on the learning algorithms compared, through their learning curves, the oracle risk $\mathcal{R}^*_{n_{\text{tr}}}(F_\lambda)$ as a function of the training size $n_{\text{tr}}$. The learning curves of different algorithms can cross: on \texttt{PCam}, \texttt{DenseNet121} has a significantly higher benchmarking score than \texttt{MobileNetV2} at $n_{\text{tr}} = {24}$ and a significantly lower one at $n_{\text{tr}} = {240}$ (\autoref{tab:stat_significance}), which is why \autoref{sec:pairwise-ranking} evaluates rankings separately at each training size. In this appendix, we compare algorithms that differ only in a hyper-parameter that orders their oracle risk at every training size, such as \texttt{ExtraTrees} with different numbers of trees, whose oracle risk decreases as trees are added. Their oracle ranking is therefore the same at every training size we consider.{} This makes it possible to ask a complementary question: how much can increasing the number of partitions reduce the study size needed to recover the oracle ranking with high probability?

\paragraph{From score variance to ranking reliability} Consider a family of $M$ learning algorithms with distinct oracle {risks} at training size $n_{\text{tr}}$, indexed without loss of generality in oracle order $F_1 \succ F_2 \succ \dots \succ F_M$. For a study set of size $N {= n_{\text{tr}} + n_{\text{te}}}$ and a CV procedure with $K$ splits, let $\widehat{R}_K(F_{j}; N)$ be the MCCV estimator of $F_{j}$, all computed on the same splits (paired). We say the \emph{full ranking} is recovered when the empirical order of the whole family matches the oracle order, and define the ranking-retrieval probability
\begin{equation}
\pi_K(N) \;=\; \mathbb{P}\!\left[\, \widehat{R}_K(F_1; N) \succ \widehat{R}_K(F_2; N) \succ \dots \succ \widehat{R}_K(F_M; N) \,\right],
\label{eq:ranking_retrieval_prob}
\end{equation}
where $\succ$ denotes the same ordering convention as the oracle ranking (loss vs.\ utility). This probability is monotone non-decreasing in both $N$ and $K$ when score variance dominates the comparison, and tends to $1$ as {$N$ grows. As $K$ grows at fixed $N$, it rises only towards the retrieval probability of the $K \to \infty$ estimate on a study set of that size, which stays below $1$}.

The reasoning of \autoref{sec:sample_gain} can then be transposed from score variance to ranking reliability: rather than asking how much extra test data a single hold-out would need to match the variance of a $K$-split estimator, we ask how much smaller a study set $K$-split CV needs to reach a given full-ranking-retrieval level relative to a single hold-out.

\begin{definition}[Threshold-Crossing Study Size]
For a family of algorithms with a fixed oracle ranking, a CV procedure with $K$ splits, and a confidence level $\alpha \in (\tfrac{1}{2}, 1)$, the \emph{threshold-crossing study size} is the smallest study size at which the full oracle ranking is recovered with probability at least $\alpha$:
\begin{equation}
N^{\star}_{K,\alpha} \;=\; \min\Big\{ N \geq 1 \;\Big|\; \pi_K(N) \;\geq\; \alpha \Big\}.
\label{eq:N_star_rank}
\end{equation}
\end{definition}

\begin{figure}[h!]
    \centering
    \begin{subfigure}[t]{0.48\linewidth}
        \centering
        \includegraphics[width=\linewidth]{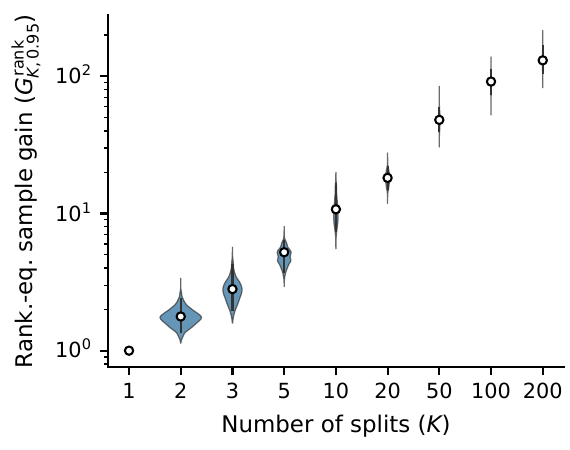}
        \caption{Ranking-equivalent sample gain $G^{\mathrm{rank}}_{K,\alpha=0.95}$ for the \texttt{ExtraTrees} family. {White dots: estimate on all seeds. Violins: bootstrap distribution, with its $95\%$ interval (vertical lines).}}
        \label{fig:violin_extratreesvsridge}
    \end{subfigure}\hfill
    \begin{subfigure}[t]{0.48\linewidth}
        \centering
        \includegraphics[width=\linewidth]{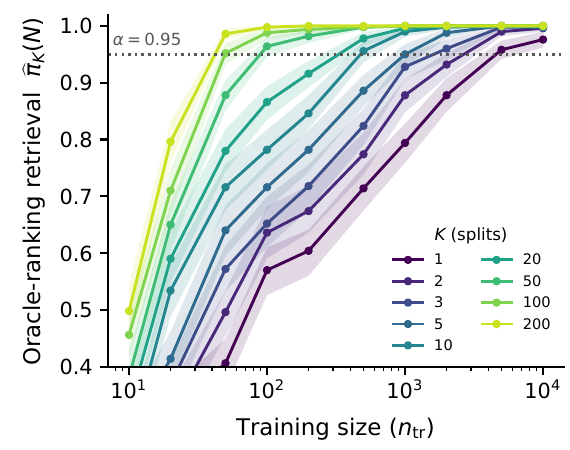}
        \caption{Empirical full-ranking retrieval-probability curves $\widehat{\pi}_K(N)$ for the \texttt{ExtraTrees} family at several split counts $K$. {Dotted} line: $\alpha = 0.95$ threshold.}
        \label{fig:extratreesrank}
    \end{subfigure}
    \caption{Ranking-equivalent sample gain on the simulated regression task for the \texttt{ExtraTrees} family, whose oracle order ($n_{\text{estimators}}\!=\!200 \succ 20 \succ 2$) is fixed by hyperparameter monotonicity. \textbf{(a)}~{Gain estimated on all $500$ seeds. The bootstrap over seeds recomputes the numerator and the denominator on the same resample. At $K=1$ the gain is $1$ in every resample, so only its dot is drawn.} \textbf{(b)}~The {study size at which the isotonic fit of each curve crosses} the $\alpha = 0.95$ threshold is the empirical analogue of $N^{\star}_{K,\alpha}$ used to compute the gain in (a) (the $K=1$ curve providing the numerator). {The curves are drawn on $10$ of the $25$ study sizes, while $N^{\star}_{K,\alpha}$ is read on all $25$. The horizontal axis gives the training size $n_{\text{tr}}$. The study sets, of size $N$, are about $1.25$ times larger, a constant factor that cancels in the gain.}}
    \label{fig:ranking_sample_gain}
\end{figure}

By the monotonicity of $\pi_K$ in $N$, this size is well-defined as soon as $\alpha$ is reachable. In practice it is read off an empirical retrieval-probability curve $\widehat{\pi}_K(N)$ at the level $\alpha$ (typically $\alpha = 0.95$), as illustrated in \autoref{fig:extratreesrank}. {The empirical proportions are noisy and need not be monotone in $N$. We therefore read the crossing off their isotonic fit (the non-decreasing least-squares fit over the study sizes), interpolating log-linearly in $N$ between the two sizes that bracket $\alpha$.}

\begin{definition}[Ranking-Equivalent Sample Gain]
The \textbf{ranking-equivalent sample gain} of a $K$-split CV procedure is the ratio of the single-split threshold-crossing size to the $K$-split threshold-crossing size:
\begin{equation}
G^{\mathrm{rank}}_{K,\alpha} \;=\; \frac{N^{\star}_{1,\alpha}}{N^{\star}_{K,\alpha}}.
\label{eq:rank_sample_gain}
\end{equation}
\end{definition}

A value $G^{\mathrm{rank}}_{K,\alpha} = 10$ indicates that, for the family at hand, $K$-split CV recovers the full oracle ranking at the $\alpha$ retrieval level with a study set ten times smaller than the one a single hold-out would require. By construction $G^{\mathrm{rank}}_{1,\alpha} = 1$, and the gain is non-decreasing in $K$ whenever $\pi_K$ is non-decreasing in $K$. In practice, $\pi_K$ is estimated by the empirical full-ranking retrieval proportion across independent seeds. {\autoref{fig:violin_extratreesvsridge} reports $G^{\mathrm{rank}}_{K,\alpha}$ estimated on all $500$ seeds, with a $95\%$ percentile interval from a non-parametric bootstrap over seeds ($10{,}000$ resamples). Each resample draws the seeds once, with replacement, and recomputes from them the numerator $N^{\star}_{1,\alpha}$ as well as every $N^{\star}_{K,\alpha}$, each read off the isotonic fit of its resampled curve. The interval thus includes the estimation noise of the numerator, and $G^{\mathrm{rank}}_{1,\alpha}=1$ in every resample. We checked its calibration by splitting the seeds into disjoint groups of $100$ ($60$ partitions, $300$ groups). The intervals computed on one group contain the $500$-seed estimate in $2{,}359$ of $2{,}400$ cases ($98.3\%$, $K=2$ to $200$). Each group makes up a fifth of the reference, so exactly calibrated intervals would contain it in about $97\%$ of the cases.}

\paragraph{A family with a known oracle ranking} Computing $G^{\mathrm{rank}}_{K,\alpha}$ requires knowing the full oracle ranking of the family. To guarantee this without relying on benchmarking-set estimates, we construct families in which the ordering is fixed by design, by varying a single hyperparameter that monotonically orders the oracle risk: for \texttt{Ridge}, the regularization strength (under the simulated linear data process of \autoref{appendix:experimental_details}, {a stronger regularization} induces more bias and monotonically degrades the oracle risk), and for \texttt{ExtraTrees}, the number of estimators $n_{\text{estimators}}$ (more estimators monotonically reduce the oracle risk through ensemble averaging). We illustrate with the \texttt{ExtraTrees} family ($n_{\text{estimators}} \in \{2, 20, 200\}$, oracle order $n_{\text{estimators}}\!=\!200 \succ 20 \succ 2$).

\autoref{fig:violin_extratreesvsridge} shows that recovering the full oracle ranking benefits markedly from additional splits: even $K=10$ partitions already multiply the effective study size roughly tenfold {($10.7$, $95\%$ interval $7.3$ to $16.8$)}, and the gain keeps climbing steeply with $K$, {reaching about two orders of magnitude by $K=100$ ($92$, $73$ to $113$) and exceeding it at $K=200$ ($131$, $104$ to $169$)}. The effect continues well beyond the first few {splits}, again indicating that the ``each observation has been held out once'' heuristic is too conservative for ranking-oriented benchmarks. The retrieval-probability curves in \autoref{fig:extratreesrank} make the construction concrete: each $K$-curve crosses the $\alpha=0.95$ threshold at its own $N^{\star}_{K,\alpha}$, and the gain is the ratio of these threshold-crossing study sizes relative to the $K=1$ curve. {Up to $K=100$ the gain grows roughly in proportion to $K$ ($G^{\mathrm{rank}}_{K,0.95}/K$ between $0.89$ and $1.07$), and it falls behind only at $K=200$ ($0.65\,K$).} Thus, for ranking-oriented benchmarks, the number of splits should be chosen as a statistical--computational trade-off rather than fixed by the traditional single-pass intuition.

This quantity is complementary to $G^{\text{test}}_K$: it does not measure the variance of a score estimator directly, but the study size needed to make the ranking decision reliable. {The two gains express} the same variance-reduction mechanism in different units: additional test samples for score estimation, or additional study samples for ranking recovery. Both views point to the same operational recommendation: use as many splits as the computational budget permits when the benchmark data are scarce and the target comparison is close.

\FloatBarrier
\newpage
\phantomsection
\label{appendix:checklist}
\section*{NeurIPS Paper Checklist}

\begin{enumerate}

\item {\bf Claims}
    \item[] Question: Do the main claims made in the abstract and introduction accurately reflect the paper's contributions and scope?
    \item[] Answer: \answerYes{} %
    \item[] Justification: {\NEWon The abstract and introduction state the three contributions: the sample gain (\autoref{sec:sample_gain}), diminishing returns setting in later than expected (\autoref{sec:empirical-results}), and early-stopped cross-validation (\autoref{sec:empirical-results}, {\NEWon Appendix}~\ref{sec:study-only-redundancy-statistics}).}

\item {\bf Limitations}
    \item[] Question: Does the paper discuss the limitations of the work performed by the authors?
    \item[] Answer: \answerYes{} %
    \item[] Justification: We discuss limitations in the conclusion.

\item {\bf Theory assumptions and proofs}
    \item[] Question: For each theoretical result, does the paper provide the full set of assumptions and a complete (and correct) proof?
    \item[] Answer: \answerYes{} %
    \item[] Justification: {\NEWon The paper states no theorems. Its definitions (Sections~2 and~3, \autoref{appendix:ranking-sample-gain}) and derivations (Appendices~\ref{appendix:benchmarking-derivations} to~\ref{sec:sample-gain-diagnostics}) state their assumptions: i.i.d.\ observations and exchangeable splits, independent seeds and finite fourth moments for the covariance estimators (\autoref{sec:inter-fold-covariance-estimation}), and the {\NEWon inverse-test-size} scaling of the single-split variance for the variance-ratio form of the gain ({\NEWon Appendix}~\ref{sec:sample-gain-estimation}). The gain-redundancy link ({\NEWon Appendix}~\ref{sec:study-only-redundancy-statistics}) builds on the evaluation-noise approximation of Section~3 and includes a coupling term that the study set cannot estimate: it is an approximation, checked empirically, not proved.}

    \item {\bf Experimental result reproducibility}
    \item[] Question: Does the paper fully disclose all the information needed to reproduce the main experimental results of the paper to the extent that it affects the main claims and/or conclusions of the paper (regardless of whether the code and data are provided or not)?
    \item[] Answer: \answerYes{} %
    \item[] Justification: {\NEWon \autoref{sec:experimental-evaluation} and \autoref{appendix:experimental_details} describe the simulated processes and redundancy designs ({\NEWon Appendix}~\ref{appendix:simulated-process}, {\NEWon Appendix}~\ref{appendix:redundancy-runs}), the real-data splits, networks, training loops and hyper-parameters ({\NEWon Appendix}~\ref{appendix:real-training-loops}), and the training sizes (\autoref{tab:xp_splitting}), seeds, numbers of splits and compute (\autoref{tab:xp_time_long}). {\NEWon Appendix}~\ref{sec:sample-gain-estimation} and \autoref{appendix:gain-intervals} give the gain estimator and its intervals.}

\item {\bf Open access to data and code}
    \item[] Question: Does the paper provide open access to the data and code, with sufficient instructions to faithfully reproduce the main experimental results, as described in supplemental material?
    \item[] Answer: \answerYes{} %
    \item[] Justification: {\NEWon The code that reruns all experiments and draws all figures is at \codeurl, and both real datasets are public (\autoref{tab:assets}).}

\item {\bf Experimental setting/details}
    \item[] Question: Does the paper specify all the training and test details (e.g., data splits, hyperparameters, how they were chosen, type of optimizer) necessary to understand the results?
    \item[] Answer: \answerYes{} %
    \item[] Justification: This is detailed in {\NEWon\autoref{appendix:experimental_details}}.

\item {\bf Experiment statistical significance}
    \item[] Question: Does the paper report error bars suitably and correctly defined or other appropriate information about the statistical significance of the experiments?
    \item[] Answer: \answerYes{} %
    \item[] Justification: {\NEWon The factor of variability is the seed, which redraws the data or study set, the splits and the training randomness. The gain figures show the estimate on all seeds (thick lines) and bootstrap resamples of the seeds (thin lines), and \autoref{appendix:gain-intervals} gives nonparametric percentile intervals at each $K$ from $1{,}000$ such resamples, nominally $95\%$, whose coverage on the simulated runs is $90.0\%$ at $50$ seeds (the real-data size) and $93.3\%$ at $200$. {\NEWon Appendix}~\ref{sec:study-only-redundancy-statistics} reports the $10$--$90\%$ range of single-run redundancies, and pairwise rankings use paired $t$-tests (\autoref{appendix:statistical_significance}) and $95\%$ Wilson intervals on retrieval rates (\autoref{fig:nlp_ranking}).}

\item {\bf Experiments compute resources}
    \item[] Question: For each experiment, does the paper provide sufficient information on the computer resources (type of compute workers, memory, time of execution) needed to reproduce the experiments?
    \item[] Answer: \answerYes{} %
    \item[] Justification: This is detailed in {\NEWon\autoref{tab:xp_time_long}}.
    
\item {\bf Code of ethics}
    \item[] Question: Does the research conducted in the paper conform, in every respect, with the NeurIPS Code of Ethics \url{https://neurips.cc/public/EthicsGuidelines}?
    \item[] Answer: \answerYes{} %
    \item[] Justification: This research {\NEWon conforms} with the NeurIPS Code of Ethics. Possible societal impacts of this work are discussed in the conclusion.

\item {\bf Broader impacts}
    \item[] Question: Does the paper discuss both potential positive societal impacts and negative societal impacts of the work performed?
    \item[] Answer: \answerYes{} %
    \item[] Justification: Possible societal impacts of this work are discussed in the conclusion.
    
\item {\bf Safeguards}
    \item[] Question: Does the paper describe safeguards that have been put in place for responsible release of data or models that have a high risk for misuse (e.g., pre-trained language models, image generators, or scraped datasets)?
    \item[] Answer: \answerNA{} %
    \item[] Justification: The paper poses no such risks.

\Needspace{5\baselineskip}
\item {\bf Licenses for existing assets}
    \item[] Question: Are the creators or original owners of assets (e.g., code, data, models), used in the paper, properly credited and are the license and terms of use explicitly mentioned and properly respected?
    \item[] Answer: \answerYes{} %
    \item[] Justification: {\NEWon \autoref{tab:assets} and the paragraph \emph{Assets and licenses} ({\NEWon Appendix}~\ref{appendix:real-training-loops}) cite the creators of the datasets, pre-trained weights and libraries used, with their versions and licenses or terms of use. All uses are non-commercial research.}

\item {\bf New assets}
    \item[] Question: Are new assets introduced in the paper well documented and is the documentation provided alongside the assets?
    \item[] Answer: \answerYes{} %
    \item[] Justification: {\NEWon The released code (\codeurl) is documented by a README at its root and in each of its folders. No new dataset is released.}

\item {\bf Crowdsourcing and research with human subjects}
    \item[] Question: For crowdsourcing experiments and research with human subjects, does the paper include the full text of instructions given to participants and screenshots, if applicable, as well as details about compensation (if any)? 
    \item[] Answer: \answerNA{} %
    \item[] Justification: {\NEWon The paper involves neither crowdsourcing nor research with human subjects.}

\item {\bf Institutional review board (IRB) approvals or equivalent for research with human subjects}
    \item[] Question: Does the paper describe potential risks incurred by study participants, whether such risks were disclosed to the subjects, and whether Institutional Review Board (IRB) approvals (or an equivalent approval/review based on the requirements of your country or institution) were obtained?
    \item[] Answer: \answerNA{} %
    \item[] Justification: {\NEWon The paper involves neither crowdsourcing nor research with human subjects.}

\item {\bf Declaration of LLM usage}
    \item[] Question: Does the paper describe the usage of LLMs if it is an important, original, or non-standard component of the core methods in this research? Note that if the LLM is used only for writing, editing, or formatting purposes and does \emph{not} impact the core methodology, scientific rigor, or originality of the research, declaration is not required.
    \item[] Answer: \answerNA{} %
    \item[] Justification: The core method development in this research does not involve LLMs as any important, original, or non-standard components.

\end{enumerate}

\end{document}